%% file: main.tex
\documentclass[10pt,twocolumn,letterpaper]{article}
\usepackage[T1]{fontenc}

\usepackage[pagenumbers]{cvpr} 

\usepackage{needspace}
\usepackage{placeins}  
\usepackage{lmodern}

\usepackage{amsmath,amssymb,booktabs,graphicx,multirow,xcolor}
\usepackage{tikz}
\usepackage{booktabs,tabularx,array}
\usetikzlibrary{arrows.meta,positioning}
\definecolor{cvprblue}{rgb}{0.21,0.49,0.74}
\usepackage[pagebackref,breaklinks,colorlinks,allcolors=cvprblue]{hyperref}
\usepackage[capitalize,noabbrev]{cleveref}
\usepackage{bibunits}  

\input{appendix_refs}
\makeatletter
\long\def\appx@pagebackref#1#2#3{\hyperlink{page.S#1}{#1}}
\AtBeginEnvironment{bibunit}{%
  \edef\@extra@b@citeb{@bu\the\numexpr\@bibunitauxcnt+1\relax}%
  \let\@extra@binfo\@extra@b@citeb
  \ifx\backrefxxx\hyper@page@backref\let\backrefxxx\appx@pagebackref\fi}
\AtBeginDocument{%
  \let\appx@orig@backout\Hy@backout
  \def\Hy@backout#1{{\let\@auxout\@mainaux\appx@orig@backout{#1}}}}
\makeatother

\def\confName{CVPR}
\def\confYear{2027}

\newcommand{\fitcolumn}[1]{\resizebox{\ifdim\width>\columnwidth\columnwidth\else\width\fi}{!}{#1}}
\newcommand{\sg}{\operatorname{sg}}
\newcommand{\TV}{\operatorname{TV}}

\newif\ifdraft\drafttrue

\title{AtomEx: Atomic Explanations of Vision Transformers with Integer Scorecards}

\author{
\textbf{Magdalena Tr\k{e}dowicz}$^{1,2}$ \\
\texttt{\small magdalena.tredowicz@doctoral.uj.edu.pl} \\
\and
\textbf{\L{}ukasz Struski}$^{1}$ \\
\texttt{\small lukasz.struski@uj.edu.pl} \\
\and
\textbf{Arkadiusz Lewicki}$^{4}$ \\
\texttt{\small alewicki@wsiz.edu.pl}
\and
\textbf{Jacek Tabor}$^{1,3}$ \\
\texttt{\small jacek.tabor@uj.edu.pl} \\
\\[-0.5em]
\parbox{\linewidth}{\small\raggedright
$\blacktriangleright\,^{1}$\,Faculty of Mathematics and Computer Science, Jagiellonian University, Poland; \quad
$\blacktriangleright\,^{2}$\,Doctoral School of Exact and Natural Sciences, Jagiellonian University, Poland; \quad
$\blacktriangleright\,^{3}$\,Centre for Credible Artificial Intelligence, Warsaw University of Technology, Poland; \quad
$\blacktriangleright\,^{4}$\,University of Information Technology and Management in Rzeszów, Poland; \quad
}
}

\begin{document}
\twocolumn[{%
\renewcommand\twocolumn[1][]{#1}%
\maketitle
\begin{center}
\centering
\includegraphics[width=\textwidth]{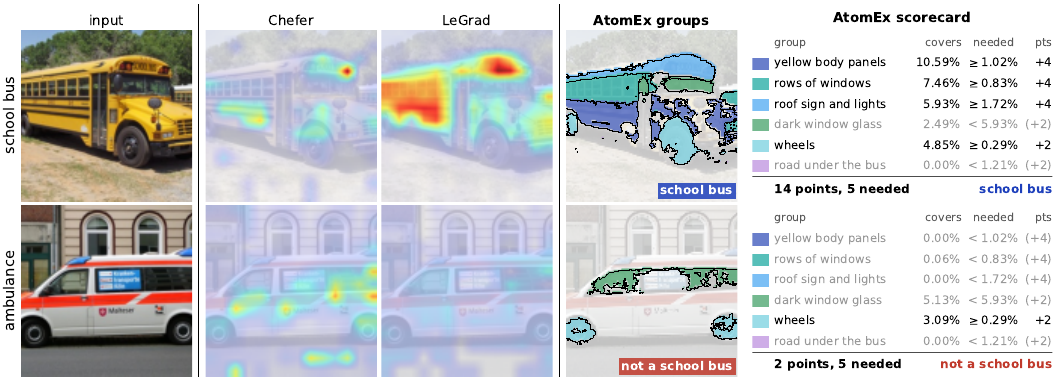}
\captionof{figure}{\textbf{The same atoms and rule explain acceptance and rejection.} Left to right: input, Chefer and LeGrad maps, AtomEx groups, and the school-bus scorecard (seed $101$). Each attribution method is normalized jointly over the two images, preserving its weaker response to the ambulance. Only the six groups retained by the card are colored. A group earns points when it covers at least the ``needed'' share of the native $56\times56$ lattice; gray entries with parenthesized points are inactive. Acceptance requires $5$ points. The bus earns $14$; the ambulance earns $2$ for wheels, while dark glass falls below its threshold and the other groups do not count. Names are assigned by inspection (\cref{fig:collage}). Group displays use denser maps, while the card uses native-lattice areas (\cref{app:finer,app:card}).}
\label{fig:teaser}
\end{center}
\vspace{2mm}
}]
\begin{abstract}
A visual explanation should make its components understandable and give a simple rule for combining them. Heat maps show where the relevant evidence lies, but not how it combines into a decision. We introduce AtomEx, a method that explains the class decision of a frozen vision transformer using recurring spatial components and an integer scorecard. A component earns points when its area exceeds its threshold, and the image is accepted when the total reaches the required score. AtomEx learns the components from the classifier's own tokens and decisions, without a segmentation model or part annotations. To learn the components, we use a monotone surrogate that reads only their maps, together with total variation on a 56 × 56 lattice, which favors smaller and less fragmented components. The scorecard is then fitted on the frozen components. Across twenty randomly sampled ImageNet classes, these positive-evidence cards achieve $98.4\%$ balanced fidelity to the teacher, versus $98.9\%$ for the surrogate; $58$ of $60$ cards use four to six components. Two-sided cards can additionally represent evidence against the class. The components often correspond to recognizable parts and localize evidence competitively with most attribution methods, although LeGrad is more precise. AtomEx thus turns a visual explanation into an explicit procedure: identify the components, compare their areas with the thresholds, and add up the points. Project page: \url{https://github.com/gmum/AtomEx}.
\end{abstract}
\vspace{-0.2cm}
\section{Introduction}
\label{sec:intro}
An explanation should make a decision understandable by identifying its elementary factors and then showing how they combine. We take this as a design principle: these factors, which we call atoms, should be recognizable across cases, and their combination should be simple enough for a person to understand, learn and apply to a new case. The principle joins decomposition as a route to understanding~\cite{descartes1637discourse,simon1962architecture,marr1982vision} with scoring systems in which a few conditions contribute small integer points~\cite{ustun2016slim,ustun2019riskslim}.

In explanations of visual classifiers, these two requirements are rarely met together. A heat map may highlight a school bus, but it does not provide a reusable vocabulary of components or a rule for combining them. In \cref{fig:teaser}, attribution maps also highlight regions of an ambulance that the network rejects as a school bus. Their magnitudes differ, but the maps leave open which combination of features is sufficient for acceptance.

AtomEx makes this combination explicit. For one class of a frozen classifier, the \emph{teacher}, it learns disjoint spatial components, which we call \emph{groups}. Each group is defined by one prototype shared across images. The vocabulary is derived directly from the teacher’s tokens and decisions, without relying on an external segmentation model or part annotations. A scorecard turns this vocabulary into a rule: a group earns integer points when its area exceeds a threshold, and the total must reach a fixed score. The school bus in \cref{fig:teaser} earns $14$ points, with $5$ required. The ambulance earns only $2$ for bus-like wheels; its dark-glass group falls below its threshold. The same atoms and the same scoring rule account for both acceptance and rejection.

Our primary scorecard counts only \emph{positive evidence}. Rejection simply means that the recognized components do not provide enough support. This gives a concise account of what makes an image resemble the target class. A two-sided variant also subtracts points for negative groups, making opposing evidence explicit. We distinguish these outputs throughout: the main results report positive-evidence cards, and a separate selection rule chooses the two-sided variant when its selection fidelity improves by more than $0.5$ percentage points (\cref{sec:card,app:stress}).

Our central methodological choice is to learn the groups first and to simplify their combination afterwards. Training a discrete card directly can yield accurate decisions but poorer components: as a sigmoid approaches a step, the threshold gradients vanish and the groups grow in order to cross the thresholds (\cref{app:card-routes}). AtomEx instead learns the partition through a flexible monotone CNN that sees only group maps. Signed coefficients define each group's direction of influence in this surrogate. Discard channels absorb regions outside the explanatory vocabulary, while total variation (TV) on a shifted-view lattice reduces the extent and fragmentation of used groups. We then freeze the groups and fit the scorecard to the teacher's decisions. Across the three main tasks, five or six positive groups with points $1$--$7$ reach $99.0\%$ balanced fidelity, close to the surrogate's $99.3\%$.

Our contributions are:
\begin{itemize}\itemsep0pt
\item \textbf{Visual atoms with an explicit composition rule.} AtomEx combines a persistent, disjoint spatial vocabulary with an integer scorecard, learned from a frozen classifier's own representations and decisions. To our knowledge, it is the first explainer to provide this combination without external segmentation or part supervision (\cref{sec:model,sec:card}).
\item \textbf{From a flexible model to a simple rule.} A monotone surrogate with discard channels and TV learns the components. A subsequent fit compresses their combination into a short rule. Positive-evidence cards achieve $98.4\%$ fidelity across twenty randomly sampled classes, compared with $98.9\%$ for the surrogate (\cref{sec:lattice,sec:card-results}).
\item \textbf{Evaluation of the components and the resulting rules.} We test recurrence, part alignment, localization against nine attribution methods, hard-negative discrimination and responses to image edits. The surrogate study covers $33$ ImageNet classes and the card study $23$; further experiments examine a second ViT, chest X-rays and video transformers (\cref{sec:experiments,sec:checks,app:cxr,app:echo}).
\end{itemize}

The components often correspond to recognizable parts, such as bus windows or a beagle's ears. What matters is that they recur across images and serve as inputs to an explicit rule. Our results support faithful prediction on the evaluated distribution and meaningful positive components. They do not show that we recover the teacher’s causal mechanism, nor that people can learn the rule unaided. The second question is the intended test of the approach.

\begin{figure*}[t]
\centering
\begin{tikzpicture}[font=\scriptsize,>=Stealth,
box/.style={draw=black!55,rounded corners=2pt,align=center,inner sep=2.5pt},
arr/.style={->,semithick,black!70},lab/.style={align=center,text=black!85,inner sep=1pt},img/.style={inner sep=0pt,draw=black!40,line width=.3pt},
slab/.style={draw=green!40!black,fill=green!12,line width=.35pt,minimum width=.3cm,minimum height=1.1cm,inner sep=0pt}]
\node[img] (in) at (0.6,0) {\includegraphics[width=1.15cm]{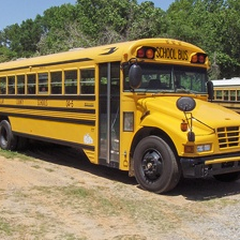}};
\node[lab,above=2pt of in] {image};
\node[img] at (2.0,0.12) {\includegraphics[width=.95cm]{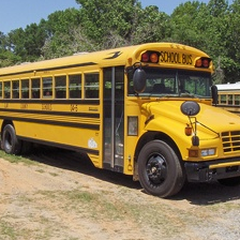}};
\node[img] at (2.12,0) {\includegraphics[width=.95cm]{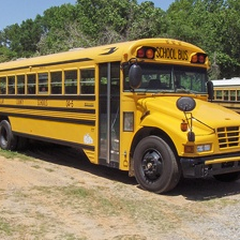}};
\node[img] at (2.24,-0.12) {\includegraphics[width=.95cm]{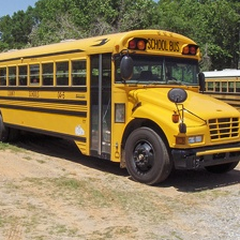}};
\node[lab] at (2.12,0.84) {16 shifted views};
\node[box,fill=black!5,text width=2.25cm] (vit) at (4.3,0) {\textbf{frozen ViT-B/16}\\blocks 1--11 + norm\\of block 12: tokens\\$x_p\in\mathbb R^{768}$};
\node[box,fill=cvprblue!9,text width=2.0cm] (soft) at (8.0,0) {\textbf{cosine softmax}\\$K{=}20$ prototypes:\\one group per site};
\node[img] (grp) at (9.95,0) {\includegraphics[width=1.15cm]{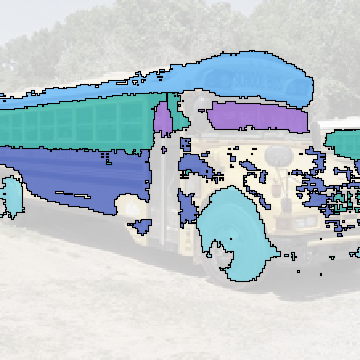}};
\node[lab,above=2pt of grp] {groups $H$};
\node[img] at (11.8,0.14) {\includegraphics[width=.8cm]{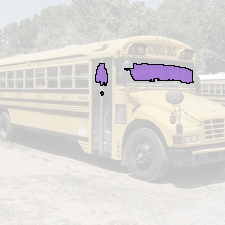}};
\node[img] at (11.67,0) {\includegraphics[width=.8cm]{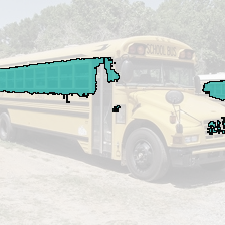}};
\node[img] (h0) at (11.54,-0.14) {\includegraphics[width=.8cm]{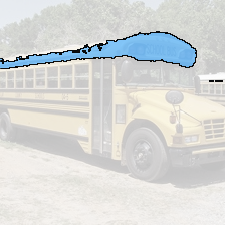}};
\node[lab] at (11.67,0.84) {$\gamma_k H_k$};
\foreach \i/\x in {1/12.85,2/13.55,3/14.25} {
  \node[slab] (sb\i) at (\x+0.1,-0.07) {};\node[slab] at (\x+0.05,-0.035) {};\node[slab] (s\i) at (\x,0) {};}
\node[lab] at (13.6,0.84) {\textbf{positive CNN}};
\node[lab] at (13.75,-0.98) {$3{\times}3$ convs, kernels $\mathrm{softplus}(\Theta){\ge}0$,\\nondecreasing activations $\phi$};
\node[box,fill=green!6] (pool) at (15.3,0) {mean\\pool};
\node[circle,draw=black!60,fill=orange!10,inner sep=1.2pt] (F) at (16.75,0) {$F$};
\draw[arr] (in.east)--(1.51,0);
\draw[arr] (2.73,0)--(vit.west);
\draw[arr] (vit.east)--node[lab,above]{interleave}node[lab,below]{$56{\times}56$}(soft.west);
\draw[arr] (soft.east)--(grp.west);
\draw[arr] (grp.east)--node[lab,above]{used}(11.13,0);
\draw[arr] (12.21,0)--(s1.west);
\draw[arr] (13.1,0)--(s2.west);\draw[arr] (13.8,0)--(s3.west);
\draw[arr] (14.48,0)--(pool.west);
\draw[arr] (pool.east)--node[lab,above]{${\ge}0$}(F.west);
\node[box,fill=black!3,text width=3.4cm] (teach) at (4.3,-1.95) {teacher decision $t$: is the top-1 class\\of the full frozen ViT the target?};
\draw[arr] (vit.south)--(teach.north);
\node[img] (dis) at (9.95,-1.95) {\includegraphics[width=.65cm]{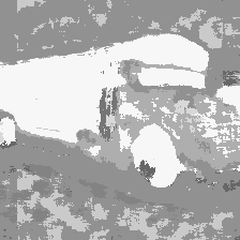}};
\node[lab,anchor=east] at (9.55,-1.95) {discard channels 10--19:\\no readout};
\draw[arr,dashed] (grp.south)--(dis.north);
\node[box,fill=orange!8] (loss) at (15.85,-1.95) {$\mathrm{BCE}(F,t)+\rho\,\TV_{\rm ch}(A)$};
\draw[arr] (F.south)--(F.south|-loss.north);
\draw[arr] (teach.south)--++(0,-0.32)-|(loss.south);
\draw[arr,dashed] (h0.south)|-node[lab,pos=.72,above]{TV, dropout $p$}(loss.west);
\end{tikzpicture}
\caption{\textbf{Learn the atoms through a flexible surrogate.} Sixteen shifted views of the frozen ViT-B/16 supply a $56\times56$ descriptor lattice. A cosine softmax over twenty prototypes assigns each site to one group. Ten used maps, multiplied by signed coefficients $\gamma_k$, enter a monotone CNN; ten discard channels are excluded from the readout. Balanced BCE fits the teacher's decision, and TV regularizes soft memberships (\cref{sec:training}). After training, the prototypes are frozen and a scorecard replaces the CNN (\cref{sec:card}). Thumbnails show a school-bus model.}
\label{fig:flat}
\end{figure*}

\section{Related Work}
\label{sec:related}
\noindent\textbf{Attribution maps.}
Attribution methods assign relevance to locations within an image: CAM and Grad-CAM~\cite{zhou2016cam,selvaraju2017gradcam}, path and relevance methods~\cite{sundararajan2017integrated,bach2015lrp}, and perturbation or mask methods~\cite{petsiuk2018rise,fong2017meaningful,fong2019extremal,dabkowski2017realtime}. ViT methods include attention rollout and flow~\cite{abnar2020attention}, attention-gradient relevance~\cite{chefer2021transformer,chefer2021generic}, and more recent approaches~\cite{achtibat2024attnlrp,bousselham2025legrad,wu2024tokentm,xie2023vitcx,wrobel2026dave}. TV-regularized masks~\cite{fong2017meaningful} and amortized Shapley estimators~\cite{covert2023vitshapley,jethani2022fastshap} likewise produce attribution maps. AtomEx instead provides a recurring spatial vocabulary together with a decision rule over its components.

\noindent\textbf{Concepts.}
Concept methods expose recurring structure in representations, including TCAV~\cite{kim2018tcav}, ACE, ICE, CRAFT and related factorizations~\cite{ghorbani2019ace,zhang2021ice,fel2023craft,fel2023holistic,vielhaben2023mcd}, sparse-autoencoder dictionaries~\cite{fel2025archetypal,pach2025sparse,stevens2025interpretable}, concept relevance propagation~\cite{achtibat2023crp} and concept bottlenecks~\cite{koh2020concept,yuksekgonul2023posthoc,oikarinen2023label}. AtomEx connects a learned spatial vocabulary to a compact integer rule for one frozen class decision.

\noindent\textbf{Prototypes, parts and post-hoc variants.}
ProtoPNet and successors~\cite{chen2019looks,nauta2021neural,rymarczyk2022interpretable,nauta2023pipnet,donnelly2022deformable,xue2024protopformer,ma2024protovit} learn classifiers through prototype evidence; prototype heads on frozen ViTs~\cite{turbe2024protosvit,mannix2024comfe} use cosine similarity to frozen tokens. Particularly close are PDiscoNet~\cite{vanderklis2023pdisconet} and PDiscoFormer~\cite{aniraj2024pdiscoformer}, which learn persistent spatial parts, the latter using DINOv2 features and TV. Their classifiers read features pooled within parts, so predictive information need not be contained in the partition itself. AtomEx's readout sees only group maps, requiring the partition to carry the decision (released PDiscoFormer comparison: \cref{app:pdisco}). Signed evidence also appears in classification-by-components~\cite{saralajew2019cbc} and SCOUTER~\cite{li2020scouter}; other post-hoc methods~\cite{tan2024posthoc,borycki2026epic,struski2024infodisent} produce continuous, overlapping maps.

\noindent\textbf{Scorecards, monotone readouts and the lattice.}
Scoring systems combine binary conditions through small integer points~\cite{ustun2016slim,ustun2019riskslim}; additive models likewise expose how inputs contribute~\cite{lou2012intelligible,agarwal2021nam}. AtomEx learns visual inputs for such a rule through a signed monotone surrogate~\cite{sill1997monotonic,gupta2016monotonic,you2017deep,runje2023constrained}. Our shifted-view lattice follows a preprint applying multiple patch-grid phases to frozen DINOv2 features~\cite{ercan2026phase}. Further related work appears in \cref{app:related}.

\section{Model}
\label{sec:model}
AtomEx treats the representation used in the explanation and the rule that combines it as two separate parts. The representation has four properties: disjointness, since each lattice site has one group assignment; persistence, since shared prototypes define group identities across images; signed influence in the surrogate, relative to discard channels (\cref{eq:sign}); and \emph{closure}, meaning that the surrogate reads only group maps. Here an atom is an elementary unit of the explanation. We claim neither statistical nor causal independence, nor semantic indivisibility. During training, the surrogate may combine atoms nonlinearly, whereas the final scorecard restricts their combination to threshold tests and addition.

The teacher is the frozen torchvision ImageNet ViT-B/16~\cite{dosovitskiy2021vit}, used in inference mode. We extract patch descriptors after the layer norm preceding the last block's attention: these are the patch representations available to the class token in that block. We drop the class token and truncate the backbone before the last block, whose output patch tokens do not reach the classifier head. The surrogate therefore has no access to the class token's residual path. Each view provides $196$ descriptors $z_p\in\mathbb R^{768}$, normalized as $x_p=z_p/\|z_p\|_2$.

Let $c_0,\dots,c_{K-1}\in\mathbb R^{d}$ be trainable prototypes, $\bar c_k=c_k/\|c_k\|_2$, with $d=768$ and $K=n+m$: $n=10$ used and $m=10$ discard channels. The soft and hard memberships of a token are
\begin{equation}
\begin{aligned}
A_{p,k}&=\frac{\exp(\sqrt d\,\bar c_k^\top x_p/\tau)}{\sum_{k'}\exp(\sqrt d\,\bar c_{k'}^\top x_p/\tau)},\\
H_{p,k}&=\mathbf 1\!\left[k=\arg\max_{k'}\bar c_{k'}^\top x_p\right].
\end{aligned}
\label{eq:softmax}
\end{equation}
Each token has exactly one hard group, and the channels sum to one at every site, so groups are disjoint. The hard assignment does not depend on $\tau$.

\paragraph{Used channels, discard channels and signed coefficients.}
After the interleaving of \cref{sec:lattice}, $A_k$ and $H_k$ become $56\times56$ maps. Only the first $n$ channels reach the readout, each multiplied by a trainable real scalar,
\begin{equation}
\tilde A_k=\gamma_k A_k,\qquad k=0,\dots,n-1.
\label{eq:coef}
\end{equation}
Here $A_k$ denotes the map supplied to the readout: soft memberships early in training, a straight-through hard map in the final quarter, and the hard map $H_k$ at evaluation (\cref{sec:training}). Each $\gamma_k$ is one trainable scalar shared across images. Its final sign defines the group's role; a value near zero suppresses its contribution. The remaining $m$ discard channels have no coefficient and never reach the readout. Assigning a site to a discard channel therefore removes it from the used maps, allowing the explanation to leave irrelevant regions uncolored.

The signed maps $\tilde A$ enter a monotone CNN: three $3\times3$ convolutions with nonnegative kernels, nondecreasing activations, mean pooling and nonnegative output weights produce logit $F$ (\cref{app:readout}). Increasing one used map while holding the others fixed can only raise $F$ if $\gamma_k>0$ and lower it if $\gamma_k<0$ (\cref{eq:sign}). These signs describe the surrogate. The primary card retains positive groups; the two-sided variant can use both signs. The CNN receives no image, descriptors, class token or teacher score: predictive information must pass through the partition.

\begin{figure}[t]
\centering
\begin{tikzpicture}[font=\scriptsize,>=Stealth,
arr/.style={->,semithick,black!70},lab/.style={align=center,text=black!85,inner sep=1pt},
img/.style={inner sep=0pt,draw=black!40,line width=.3pt}]
\def\cw{0.24}\def\ch{0.2}
\node[img] (g) at (0.62,-0.05) {\includegraphics[width=1.2cm]{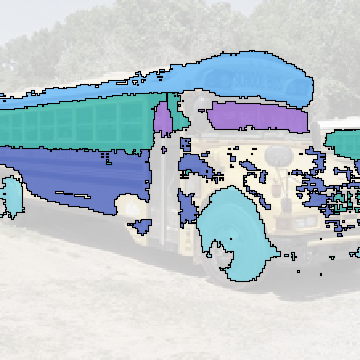}};
\node[lab] at (0.62,0.9) {groups of the\\frozen surrogate};
\begin{scope}[shift={(1.72,0.4)}]
\foreach \r/\vals in {0/{55,40,30,15,25,10,0,0,0,0},1/{0,0,0,10,20,5,35,5,10,0},2/{0,0,0,0,0,0,0,10,0,5},3/{40,50,25,0,30,15,0,0,5,0}} {
  \foreach \v [count=\c from 0] in \vals {
    \ifnum\c<6 \def\hue{cvprblue}\else\def\hue{red!75!black}\fi
    \fill[\hue!\v!white] (\c*\cw,-\r*\ch-\ch) rectangle ++(\cw,\ch);
  }
}
\draw[black!30,line width=.2pt] (0,-4*\ch) grid[xstep=\cw,ystep=\ch] (10*\cw,0);
\foreach \c in {0,...,5} \fill[cvprblue] (\c*\cw+0.04,0.05) rectangle ++(\cw-0.08,0.12);
\foreach \c in {6,...,9} \fill[red!75!black] (\c*\cw+0.04,0.05) rectangle ++(\cw-0.08,0.12);
\node at (5*\cw,-4*\ch-0.1) {$\vdots$};
\foreach \r/\t in {0/1,1/0,2/0,3/1} \node at (10*\cw+0.3,-\r*\ch-\ch/2) {$\t$};
\node[lab] at (5*\cw,0.5) {areas $a_{ik}$ of the\\ten groups, fit images};
\node[lab] at (10*\cw+0.32,0.5) {teacher\\$t_i$};
\end{scope}
\node[draw=black!55,rounded corners=2pt,fill=orange!8,align=left,inner sep=2.5pt] (fit) at (6.75,-0.05) {\textbf{fit to the teacher}\\groups with $\gamma_k>0$\\$m_k$: coordinate descent\\$p_k\ge0$: convex\\integer points};
\draw[arr] (g.east)--(1.66,-0.05);
\draw[arr] (4.72,-0.05)--(fit.west|-0,-0.05);
\node[draw=black!55,rounded corners=2pt,inner sep=2.5pt] (card) at (4.1,-2.25) {\begin{tabular}{@{}l@{\;}r@{\qquad}l@{\;}r@{}}
yellow panels $\ge1.02\%$ & $+4$ & dark glass $\ge5.93\%$ & $+2$\\
windows $\ge0.83\%$ & $+4$ & wheels $\ge0.29\%$ & $+2$\\
roof sign $\ge1.72\%$ & $+4$ & road $\ge1.21\%$ & $+2$\\[1pt]
\multicolumn{4}{@{}c@{}}{school bus if the points of the counted groups reach $5$}
\end{tabular}};
\draw[arr] (fit.south)--(fit.south|-card.north);
\node[lab,anchor=south west] at (card.north west) {integer scorecard (school bus, seed $101$)};
\end{tikzpicture}
\caption{\textbf{Freeze the atoms, simplify their composition.} Group areas and teacher decisions form a fitting table (schematic values; blue: positive surrogate coefficients, red: negative). The primary card uses only positive groups. For fixed area thresholds, fitting nonnegative points is convex logistic regression; thresholds are searched by coordinate descent. The final points and required total are integers (\cref{app:card}). The illustrated school-bus card retains all six positive groups. The full search, selection and rounding procedure is not jointly convex.}
\label{fig:fit}
\vspace{-0.3cm}
\end{figure}

\vspace{-0.3cm}
\paragraph{From the surrogate to a scorecard.}
\label{sec:card}
The CNN serves only to learn the components; the scorecard is the final explanation. After training, we freeze the prototypes and replace the CNN with a rule over group areas. Let $a_k$ be the fraction of lattice sites assigned to used group $k$. We fit the teacher's decisions with
\begin{equation}
\begin{gathered}
\text{class}\iff \sum_{k:\,\gamma_k>0} p_k\,\mathbf 1[a_k>m_k]\ge T ,\\
p_k\in\{0,1,2,\dots\}.
\end{gathered}
\label{eq:card}
\end{equation}
The \emph{positive-evidence card} awards $p_k$ points when $a_k>m_k$ and accepts when the total reaches $T$. At most eight groups may contribute; the fitted ImageNet cards use two to six. Every fitted positive-evidence card has $T\ge1$, so an image with no active group is rejected. This is an observed property of the fitted cards, not a constraint imposed by the optimizer. Applying a card thus takes three steps: identify the groups, compare their areas with the thresholds, and add up the points.

For fixed area thresholds, we fit nonnegative weights by logistic regression, a convex problem. We search area thresholds by coordinate descent, compare candidate fits on selection images, round the points and intercept, and then choose the decision cutoff on the selection set (\cref{app:card}). The resulting rule is expressed as an integer total $T$; the full procedure is not jointly convex. Areas are counts of native-lattice sites. Thus $a_k>m_k$ is exactly equivalent to requiring at least a particular number of sites; figures display that attainable share with $\ge$ (\cref{app:card}).

A \emph{two-sided card} uses all used groups through $b+\sum_k p_k\mathbf 1[a_k>m_k]>\theta$, with $p_k$ constrained to the sign of $\gamma_k$. A separate selection rule replaces the positive-evidence card with this variant when selection fidelity improves by more than $0.5$ percentage points (\cref{app:sides,app:stress}). Unless specified otherwise, reported card results refer to the positive-evidence card of \cref{eq:card}. In the main figures, only the groups retained on this card are colored; the appendix figures of the surrogate show all ten used channels.

\section{Lattice, Total Variation and Training}
\label{sec:lattice}
\noindent\textbf{Lattice.}
To resolve components below the original $16$-pixel patch spacing, we run the frozen ViT on sixteen views shifted by multiples of $4$ pixels. Interleaving their tokens yields a $56\times56$ lattice with $4$-pixel spacing (\cref{fig:flat}). Because adjacent sites usually come from different forward passes, TV also discourages disagreement between shifted views. Prototype assignment depends only on the descriptor at a site, allowing the same groups to be displayed more densely than the lattice used by the surrogate (\cref{fig:res,app:finer}).

\noindent\textbf{Total variation.}
We regularize the boundaries of the used groups through their soft memberships:
\begin{equation}
\TV_{\rm ch}=n\,\TV_\Delta(A_{0:n-1}),
\label{eq:tv-ch}
\end{equation}
where $\TV_\Delta$ is anisotropic total variation normalized per $16$-pixel patch (\cref{eq:tv} in \cref{app:finer}). Evaluation uses hard maps (\cref{sec:protocol}). Boundaries between discard channels are free; a used--discard boundary is charged only to the used channel. This distinction matters. Without discard channels, every site would have to belong to a used group, so a single dominant group would be a cheap way to reduce TV. Discard channels let the used groups contract and unnecessary groups disappear.

\paragraph{Training.}
\label{sec:training}
We initialize $K=20$ prototypes by spherical $k$-means (30 iterations) on unshifted tokens from up to $1{,}000$ fitting positives and $1{,}500$ fitting negatives. The $n=10$ clusters with the largest absolute $t$-statistic for presence differences between teacher-positive and teacher-negative fitting images become used channels; $\gamma_k$ is initialized to the sign of that statistic.

The target is the frozen teacher's top-1 decision, $t_i=1$ when image $i$ is assigned to the target class. Batches contain four positives and eight negatives (\cref{sec:protocol}); weights $w_i=1.5$ and $0.75$, respectively, balance the two classes. We minimize
\begin{equation}
\mathcal L=\frac1B\sum_iw_i\Big[\operatorname{BCE}(F_i,t_i)+\rho\,\TV_{\rm ch}(A_i)\Big],
\label{eq:loss}
\end{equation}
with a constant weight $\rho=0.05$ from the first epoch. TV always uses soft memberships.
The readout is trained on soft maps first and on hard maps at the end. In the first $30$ of $40$ epochs it receives the soft memberships $A$ of \cref{eq:softmax}, with a temperature that falls geometrically from $0.5$ to $0.05$, $\tau(e)=0.5\cdot0.1^{\min(1,e/30)}$, so that $A$ approaches a one-hot map. In the last quarter (epochs $\ge30$) it receives the hard assignments $H$ in the forward pass and the gradient of the soft ones in the backward pass, the straight-through construction~\cite{jang2017gumbel},
\begin{equation}
A^{\rm ST}=\sg(H-A)+A,
\end{equation}
and \cref{eq:coef} is applied to $A^{\rm ST}$. TV is computed on the soft memberships in all epochs, also when the readout sees $H$.

Before the readout, each used channel is independently zeroed per image with probability $p=0.25$, without rescaling. TV uses undropped maps, and evaluation uses all channels. Dropout preserves monotonicity; without it, one channel can dominate and TV removes the others (\cref{app:sweep}). After each epoch, hard-map selection fidelity determines the logit threshold and checkpoint; lower hard TV breaks ties. The selected threshold is fixed for testing.

\section{Fidelity, Scorecard and Groups}
\label{sec:experiments}
\paragraph{Protocol.}
\label{sec:protocol}
The main tasks are ImageNet beagle, school bus and valley, each with seeds $101$, $211$, $307$. We compare the default setting with $\rho=0$ (same dropout): $18$ runs. Thirty further classes, twenty randomly sampled, follow in \cref{sec:further}. Main-text results use $56\times56$ models; figures show seed $101$ as dense maps (\cref{fig:res}). Descriptors are cached from $64$ shifted views per image; sixteen supply the training lattice (\cref{app:store}). Fitting uses class training images and a random pool of ten training images per ImageNet class (about $10{,}000$), with disjoint selection and test splits. Test data comprise $4{,}000$ random validation images and all teacher-positive validation images of the three tasks: $64$ for beagle and $56$ for school bus. One positive changes balanced fidelity by $0.78$/$0.89$ percentage points. Random negatives are mostly easy: fidelity is $96.1\%$/$92.8\%$ against sibling classes and $67\%$/$85\%$ against the five nearest classes in the teacher's head (\cref{app:hard}).

Evaluation and figures use hard assignments $H$. We balance teacher-positive and teacher-negative images when reporting fidelity (balanced accuracy against teacher decisions), hard TV (presence maps aggregated by sign), used area by sign, and TV/area (hard TV divided by half the used area). For localization, let $P$ be the positive-group region and $B$ the union of class boxes mapped to the crop. We report precision $|P\cap B|/|P|$, recall $|P\cap B|/|B|$ and lift (precision divided by $|B|/224^2$; random-mask reference $1$). Boxes are a coarse proxy. Of the $64$/$56$ teacher-positive beagle/school-bus images, $43$/$49$ have the corresponding class annotation; the remaining positives are the teacher's predictions on other classes.

\begin{figure}[t]
\centering
\includegraphics[width=\columnwidth]{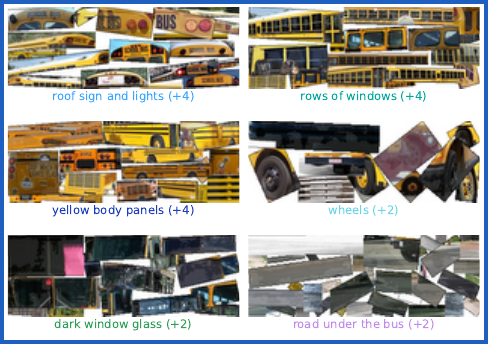}
\caption{\textbf{A visual vocabulary shared across images.} Each sheet collects one of the six school-bus card groups and gives its points. We paste the largest group fragment from each of the first $20$ test school-bus images; the road group uses up to six fragments per image (\cref{app:collage}). Names are assigned by inspection. Three negative surrogate groups, excluded from this card, appear in \cref{fig:collage-both}.}
\label{fig:collage}
\vspace{-0.3cm}
\end{figure}

\begin{figure*}[t]
\centering
\includegraphics[width=\textwidth]{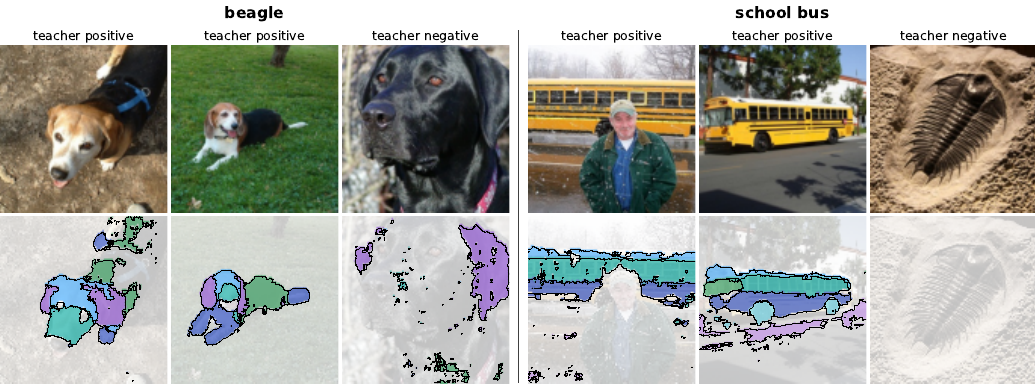}
\caption{The groups of AtomEx ($\rho=0.05$, dropout $0.25$) on beagle (left) and school bus (right): the $56\times56$ models of the tables (seed $101$) as dense maps (\cref{fig:res}), drawn as in \cref{fig:teaser} (colors do not correspond between classes); only the groups on the card are colored (\cref{fig:teaser,fig:card-beagle}). The images were fixed before the maps were drawn: the first two test images of each class (ImageNet label) that the teacher calls the class (on school bus the image of \cref{fig:teaser} is skipped and the second image is the next one that shows a whole bus) and one negative. With and without TV: \cref{fig:gallery-56}.}
\label{fig:gallery}
\vspace{-0.2cm}
\end{figure*}

\Cref{fig:gallery} shows the groups of the cards on images fixed in advance: they cover the beagles ($20\%$ and $14\%$ of the image) and parts of the buses ($23\%$ and $25\%$); on the black dog only the ears and the back occur ($13\%$, $4$ of the $6$ points needed), and the fossil has none.

\paragraph{What the groups are.}
\label{sec:groups}
A reusable atom should have a recurring visual meaning. In \cref{fig:collage}, school-bus groups collect roof signs and lights, window rows, yellow panels, wheels and dark glass. The sixth card group captures road beneath the bus: the vocabulary includes context as well as object parts. Four negative surrogate groups collect other vehicles' sides, red vehicles, lettering and trim; the positive-evidence card excludes them (\cref{fig:collage-both}). These descriptions are assigned after inspection, without part supervision.

\begin{figure}[t]
\centering
\includegraphics[width=\columnwidth]{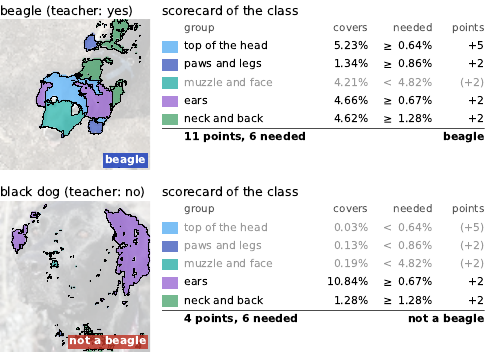}
\caption{The scorecard of the beagle surrogate of \cref{fig:gallery} (seed $101$), drawn as the card in \cref{fig:teaser}. On the beagle the head, the paws, the ears and the back count, $11$ points where $6$ are needed; on the black dog, which the teacher does not call a beagle, only the ears and the back count, $4$ points. The muzzle group counts only above $4.8\%$ of the image, on close-ups. The names are ours and only descriptive.}
\label{fig:card-beagle}
\vspace{-0.5cm}
\end{figure}

The card distinguishes shared features from sufficient evidence. Wheels and dark-glass groups occur on $78\%$ and $63\%$ of negative road-vehicle images, versus under $1\%$ of random negatives. Roof-sign and yellow-panel groups occur on only $9\%$ and $11\%$ of negative road vehicles (\cref{app:examples}). The illustrated card assigns $4$ points to these more bus-specific groups and $2$ to the shared groups: wheels and dark glass together do not reach the required $5$. Beagle groups similarly track recurring dog parts. On the rejected black dog, ears occupy $10.8\%$ of the image, but ears and back provide only $4$ of the required $6$ points (\cref{fig:card-beagle}).

We complement these examples with quantitative checks (\cref{app:seeds,app:parts}). Across three seeds, positive regions overlap substantially (IoU $0.55$ for beagle and $0.75$ for school bus, versus null values $0.16$ and $0.17$); the school-bus partition is also stable at the group level. Against PartImageNet++ annotations, groups with TV attain NMI $0.12$ and $0.10$, respectively, $11$--$17$ times the shifted-mask and other-image nulls, and $0.11$ on images excluded from the gradient fit. Three of the five positive beagle groups place $85$--$92\%$ of their pixels in the head. This supports recurring parts, not a one-to-one match to human labels.

\begin{table}[t]
\centering
\small
\caption{AtomEx per task and macro mean over the three tasks, each the mean over three seeds (fidelity: $\pm$ standard deviation over seeds), best-fidelity checkpoint, $p=0.25$. AUROC: surrogate score against the teacher decision. Hard TV per $16$-pixel patch (\cref{eq:tv}); TV/area: hard TV divided by half the used area.}
\label{tab:tasks}
\fitcolumn{\begin{tabular}{@{}llccccc@{}}
\toprule
task & arm & bal.\ fid.\ (\%) & AUROC & hard TV & area (\%) & TV/area\\
\midrule
beagle & $\rho=0$ & $98.71\pm0.22$ & 0.999 & 0.775 & 44.1 & 3.52\\
 & $\rho=0.05$ & $99.24\pm0.38$ & 0.999 & 0.438 & 22.1 & 4.00\\
school bus & $\rho=0$ & $99.49\pm0.44$ & 1.000 & 0.807 & 45.5 & 3.56\\
 & $\rho=0.05$ & $98.89\pm1.48$ & 0.999 & 0.346 & 18.2 & 3.78\\
valley & $\rho=0$ & $99.35\pm0.43$ & 1.000 & 0.931 & 67.3 & 2.77\\
 & $\rho=0.05$ & $99.67\pm0.17$ & 1.000 & 0.483 & 20.0 & 5.29\\
\midrule
macro mean & $\rho=0$ & 99.18 & 0.999 & 0.838 & 52.3 & 3.28\\
 & $\rho=0.05$ & 99.27 & 0.999 & 0.422 & 20.1 & 4.36\\
\bottomrule
\end{tabular}}
\vspace{-0.5cm}
\end{table}

\paragraph{Fidelity and the effect of TV.} TV substantially simplifies the components while preserving predictive fidelity (\cref{tab:tasks}). Across the six task--setting combinations, balanced fidelity is $98.7$--$99.7\%$. Differences between TV and no TV are generally comparable to the effect of one test positive, except for one school-bus seed. Meanwhile, used area falls from $44$--$67\%$ to $18$--$22\%$, hard TV falls from $0.78$--$0.93$ to $0.35$--$0.48$, and a group has $11$--$19$ connected pieces instead of $32$--$53$. An area penalty alone leaves more pieces and longer boundaries at matched area (\cref{app:sweep}), indicating that TV does more than shrink the masks. TV also shrinks groups below a $1\%$ reporting threshold: all ten used groups exceed $1\%$ balanced area without TV, against seven for school bus, eight or nine for beagle and five to nine for valley with TV.

\paragraph{The scorecard.}
\label{sec:card-results}
\begin{table}[t]
\centering
\small
\caption{Balanced fidelity to the teacher (\%, test images, mean over three seeds) of the surrogates of \cref{tab:tasks} and of the integer scorecards of \cref{eq:card} read from their groups with $\gamma_k>0$ (card; groups: number of groups on a card; points: range of the integer points; 4 groups: at most four, points not rounded). Agree: share of the test images on which the card decides like the surrogate. Logistic: a logistic regression on all ten areas, not a card, as a reference. 20 random: mean over the classes of \cref{app:random}. Two-sided and negative cards: \cref{app:sides}.}
\label{tab:card}
\fitcolumn{\begin{tabular}{@{}llccccccc@{}}
\toprule
task & arm & surrogate & card & groups & points & 4 groups & agree & logistic\\
\midrule
beagle & $\rho=0.05$ & 99.24 & 98.60 & 5 & 1--5 & 98.77 & 97.7 & 99.20\\
 & $\rho=0$ & 98.71 & 98.33 & 5 & 1--5 & 98.47 & 97.8 & 98.92\\
school bus & $\rho=0.05$ & 98.89 & 99.57 & 5--6 & 1--5 & 99.47 & 99.3 & 99.43\\
 & $\rho=0$ & 99.49 & 99.38 & 5--6 & 1--5 & 99.43 & 99.0 & 99.42\\
valley & $\rho=0.05$ & 99.67 & 98.67 & 5--6 & 1--7 & 97.74 & 98.1 & 99.57\\
 & $\rho=0$ & 99.35 & 98.83 & 5--6 & 2--5 & 98.65 & 98.6 & 99.33\\
\midrule
mean & $\rho=0.05$ & 99.27 & 98.95 & & & 98.66 & 98.4 & 99.40\\
 & $\rho=0$ & 99.18 & 98.85 & & & 98.85 & 98.5 & 99.22\\
\midrule
20 random & $\rho=0.05$ & 98.92 & 98.42 & 2--6 & 1--11 & 96.54 & 98.6 & 99.14\\
\bottomrule
\end{tabular}}
\vspace{-0.5cm}
\end{table}

A short positive-evidence rule preserves most of the flexible surrogate's fidelity (\cref{tab:card}). On beagle, school bus and valley, cards reach $98.60\%$, $99.57\%$ and $98.67\%$: a macro mean of $98.95\%$, versus $99.27\%$ for the CNN. Each card uses five or six groups, point values $1$--$7$ and a required total of $3$--$7$. Four-group rules reach $98.7\%$ on average, with unrounded points. Across twenty random classes, integer cards reach $98.42\%$ versus $98.92\%$ for the surrogate; $58$ of $60$ use four to six groups. The paired fidelity gap is $-0.50$ percentage points, with bootstrap interval $[-0.88,-0.13]$ (\cref{app:stress}). Simplicity therefore has a small but detectable cost on these classes. Without TV, main-task cards reach a similar $98.85\%$, but the surrogate groups cover more than twice the area (\cref{tab:tasks}).

The examples make the rule transparent. The bus earns $14$ points against a requirement of $5$, and the ambulance earns $2$. The beagle earns $11$ against a requirement of $6$, and the black dog earns $4$ (\cref{fig:card-beagle}). Rejection means insufficient activation of the learned components, not the literal absence of anatomical parts: the black dog has a head, but the group described as ``head'' does not count on it. Group names summarize recurring appearances; they are not exhaustive semantic definitions.

The limitation of positive-only scoring is clearest on similar classes. Across the twenty random classes, the card rejects $62\%$ of teacher-negative images from the five nearest classes in the teacher's head, versus $65\%$ for the surrogate; on beagle the figures fall to $16\%$ and $34\%$. Two-sided scoring recovers much of the beagle gap ($31\%$). The selection rule chooses it for $5$ of $9$ main-task models and $11$ of $60$ random-class models, reaching $99.49\%$ and $98.61\%$ test fidelity, respectively (\cref{app:stress}). These selected-card results are distinct from the positive-evidence results in \cref{tab:card}.

On the evaluated teacher-positive images, card decisions remain unchanged under the tested natural perturbations whenever the teacher's decision remains unchanged. Under deletion, however, the card rarely follows the teacher's decision flips, much like the surrogate (\cref{app:stress}). Training the step directly keeps larger, less specific groups; one-sigmoid routes give comparable positive cards but catch-all negative groups (\cref{app:card-routes}). Learning useful atoms and fitting an accurate composition rule are distinct requirements.

\section{Signs, Localization and Further Checks}
\label{sec:checks}
\paragraph{Does the teacher confirm the signs?}
\label{sec:sign}
We test whether surrogate signs predict the teacher's response to identical edits (\cref{app:sign}). Deletion replaces a group's pixels with those of another image; insertion pastes them into another image. We measure changes in the teacher's class margin, $z_{\rm class}-\max_{c\ne\rm class}z_c$, against a control applying the same shape elsewhere. On class images, deleting positive groups lowers the margin about twice as much as the control on average. For beagle and school bus, respectively, $66\%$ and $63\%$ of groups beat their control, versus $84\%$ and $85\%$ for signed AttnLRP relevance. Surrogate and teacher move in the same direction on $72\%$ and $77\%$ of positive-group deletions, but the surrogate follows only $2$--$4\%$ of teacher decision flips. Negative groups have weaker support: $51$--$61\%$ beat the deletion control on sibling images, and teacher--surrogate directional agreement is $60$--$62\%$. Positive signs thus receive aggregate support, not group-by-group verification; negative signs remain properties of the surrogate. Neither variant's fidelity on natural images establishes causal fidelity under edits.

\paragraph{Localization against object boxes.}
\label{sec:loc}
On teacher-positive images, TV raises positive-group box precision to $95.0\%$ for beagle and $94.7\%$ for school bus, from $91.1\%$ and $81.1\%$ without TV (\cref{app:loc}). Positive groups cover $24\%$ and $26\%$ of the image. A centered disk of matched area attains $95.7\%$ and $81.6\%$: AtomEx exceeds this strong spatial prior by $13$ points on school bus, while the beagle boxes do not distinguish it from the disk.
\paragraph{Comparison with attribution methods.}
\label{sec:compare}
We re-run nine attribution methods and one unsupervised-group baseline on the same frozen teacher and on the teacher-positive test images with boxes: Grad-CAM on the input of the last block~\cite{selvaraju2017gradcam}, attention rollout~\cite{abnar2020attention}, gradient-weighted attention relevance~\cite{chefer2021generic}, integrated gradients~\cite{sundararajan2017integrated}, RISE~\cite{petsiuk2018rise}, DAVE~\cite{wrobel2026dave}, LeGrad~\cite{bousselham2025legrad}, ViT-CX~\cite{xie2023vitcx} and AttnLRP~\cite{achtibat2024attnlrp}, and spherical $k$-means groups in the spirit of CRAFT~\cite{fel2023craft} (\cref{app:compare}). Every continuous map is binarized to the pixels with the highest value up to the area of our positive groups on the same image, so all are compared at an equal area set by our model. Besides the box precision we report a replacement test that needs no ground truth: the region is replaced by the same pixels of a negative image, and we report how often the teacher's top-1 decision changes, minus the same for the same shape elsewhere.

The components retain useful localization despite the additional vocabulary and scorecard constraints (\cref{tab:compare}). AtomEx reaches $95.0\%$ and $94.7\%$ box precision, competitive with most baselines; LeGrad is best at $98.3\%$ and $96.5\%$. Replacing AtomEx's positive region flips the teacher's decision about $70$ percentage points more often than the shape-matched control on both tasks. Corresponding gains are $72$/$74$ for LeGrad, $88$/$83$ for AttnLRP and $24$/$6$ for a centered disk. With only $43$ and $49$ images, small differences should be interpreted cautiously; on valley, AtomEx's net flip rate is comparable to the disk (\cref{app:compare}). AtomEx's distinctive contribution is the recurring components and their explicit composition rule, alongside this localization performance.

\paragraph{Further checks.}
\label{sec:further}
The surrogate's constraints have no detectable fidelity penalty in our comparisons: neither an unconstrained readout nor a linear probe on mean descriptors is more faithful (\cref{tab:sweep}). An additive model explains $99.97\%$ of CNN logit variance over subsets of present groups, helping explain its compressibility. A readout linear in group areas is faithful but yields larger regions ($37\%$/$31\%$ versus $26\%$/$16\%$; \cref{tab:readout}). Conventional sparse scoring systems on the same frozen areas achieve similar card fidelity at matched numbers of conditions (\cref{tab:complexity}). The contribution is the learned visual vocabulary and the resulting form of explanation, rather than superiority of a new tabular rule learner.

The broader component study supports the same pattern. Across twenty random classes and three seeds, TV retains $98.9\%$ surrogate fidelity while reducing used area from $47\%$ to $18\%$ and groups above the area-reporting threshold from ten to $6.1$. Box precision is $79.1\%$, versus $72.1\%$ for a matched-area centered disk, exceeding the disk on $16$ classes (\cref{app:random}); ten manually selected classes show a similar pattern (\cref{app:classes}). DeiT-III-B/16 gives similar surrogate results ($99.1\%$ fidelity, $20$--$26\%$ used area; \cref{app:deit}). On Swin-B, fidelity and area reduction persist, but groups on $32$-pixel cells fail to follow the object (\cref{app:swin}).

\begin{figure}[t]
\centering
\includegraphics[width=\columnwidth]{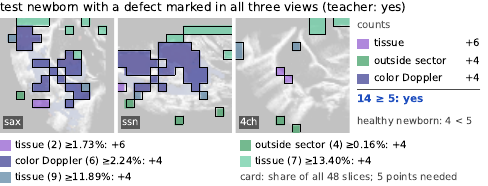}
\caption{\textbf{An echocardiography card}: EchoJEPA-B probe, ``a defect in all three views''; test newborns fixed in advance (\cref{fig:echo-card}).}
\label{fig:echo-main}
\end{figure}

\noindent\textbf{Echocardiography videos.} AtomEx also explains echocardiography video transformers that read three views of a newborn (EchoJEPA-B, EchoFM; \cref{app:echo}). Surrogates agree with the teachers on $87$--$94\%$ of test samples, two-sided cards on $83$--$92\%$; color-Doppler groups carry the sign of a defect, and removing the color shifts the teachers' decisions beyond a control. A pediatric cardiologist judged that the groups in \cref{fig:echo-main} cover most of the defect.
\section{User study}
We ran an online user study to see whether people can learn the decision rule of an image classifier from its integer scorecard. It covered three tasks: prairie chicken, beagle, school bus. For each task, participants studied the scorecard
with six solved examples and then predicted the model's YES/NO answer for eight new photos. They agreed with the
model on $60.4\%$ of the 24 photos (95\% CI $[56.1,64.7]$; $t(29)=4.71$, $p<0.001$), above chance. Agreement was highest
for photos of the target class ($79.3\%$) and lowest where the scorecard itself disagrees with the model ($30.0\%$),
where participants followed the scorecard. Full design, tables and figures are in App.~\ref{app:userstudy}.

\section{Conclusions and Limitations}
\label{sec:scope}
AtomEx makes a concrete proposal for visual explanation: learn recognizable atoms and express their composition through a rule simple enough to inspect and reuse. It derives a persistent spatial vocabulary from a frozen classifier without external segmentation or part labels, then turns that vocabulary into an integer scorecard. Positive-evidence cards reach $99.0\%$ balanced fidelity on the three main tasks and $98.4\%$ on twenty random classes. Their value is an explicit procedure shared across images: identify the components, check their thresholds, and sum their points.

The choice of composition matters. Positive-only cards explain rejection through insufficient support and are weaker on similar classes. Two-sided cards add opposing evidence; a selection-based choice reaches $99.5\%$ and $98.6\%$ fidelity on the main and random-class tasks. In the chest X-ray experiments, we report two-sided or complementary one-sided rules to inspect evidence in both directions (\cref{app:cxr}). These are surrogate explanations, not verified causal accounts: positive signs have aggregate intervention support, negative signs remain unconfirmed, and both CNN and card rarely follow teacher decision flips. Random negatives dominate the main test distribution, so hard-negative results are essential to interpreting fidelity (\cref{app:stress}).

Each class requires a separate fit and has its own vocabulary. LeGrad localizes more precisely, $28\%$ of used area lies in sub-token specks, and discard channels are uninterpreted. The surrogate study covers $33$ ImageNet classes and the integer-card study $23$; object-aligned groups do not transfer successfully to Swin-B. Finally, an inspectable rule is a design step toward human learning, not evidence of it. Testing whether people can recognize the atoms, retain the rule and apply it to unseen cases remains central to evaluating the proposed paradigm.

\section*{LLM Usage Statement}
We used Claude (Anthropic), a large language model used as a coding assistant, for parts of the analysis code of selected experiments, for drafting and editing text, and for checking the consistency and layout of the manuscript. The authors designed the method and the experiments, checked every generated result against the data, and take responsibility for the content.

{\small
\bibliographystyle{ieeenat_fullname}
\bibliography{refs}
}

\clearpage
\setcounter{page}{1}
\maketitlesupplementary

\appendix
\appendixrefs  
\begin{bibunit}[ieeenat_fullname]  

\renewcommand{\floatpagefraction}{.85}
\renewcommand{\dblfloatpagefraction}{.85}
\noindent The appendix follows the order of the paper: implementation (\cref{app:store,app:readout}) and further related work (\cref{app:related}); the lattices, the dense maps and the $56\times56$ models of the tables (\cref{app:finer}) and their stability across seeds (\cref{app:seeds}); what the groups are, with more images (\cref{app:collage,app:examples,app:parts,app:pdisco,app:discard}); the sign test (\cref{app:sign}); localization and the comparison with attribution methods (\cref{app:loc,app:compare}); the further checks of \cref{sec:further} (\cref{app:sweep,app:random,app:classes,app:deit,app:swin,app:hard}); a chest X-ray classifier (\cref{app:cxr}); a scorecard read from the groups, two-sided and negative cards, and the routes that learn a card more directly (\cref{app:card,app:sides,app:card-routes}); and echocardiography videos (\cref{app:echo}). References to the main paper are marked ``\maintag'' (e.g., \cref{tab:tasks}); references without a marker point to this appendix.

\section{Descriptor store and compute}
\label{app:store}
\paragraph{Sets.} For each task the fitting images are four fifths of the $1{,}300$ training images of the class plus a pool of ten random training images per class, without the images of the task classes ($9{,}970$ images for the three tasks of the main text); one class image in five and the last $1{,}500$ pool images form the selection set ($1{,}760$ images), and everything else is used for fitting ($9{,}510$ images, of which $982$ are teacher-positive for beagle and $1{,}032$ for school bus). The negatives of a step are drawn without repetition from the shuffled pool, and $5\%$ of them are replaced by class images that the teacher does not call positive. The test set holds $4{,}000$ random validation images and all teacher-positive validation images of the three tasks ($4{,}158$ images); for beagle $64$ of them are teacher-positive and the rest negative, for school bus $56$, so the positives are enriched against natural prevalence, as in a class-balanced comparison, while the negatives cover the whole validation set. A shifted view is a $224\times224$ window of the reflection-padded image; at most the outermost $8$ pixels of a shifted view are mirrored (half of a border token), and the unshifted view is exact. Training uses AdamW with learning rate $10^{-3}$ for the readout and $\gamma$ and $10^{-4}$ for the prototypes, gradients clipped at norm $5$, and epochs of $150$ steps of $12$ images.

\paragraph{Store.} The backbone is frozen, so the sixty-four shifted views of every image are processed once and the fp16 descriptors ($64\times196\times768$, about $19$ MB per image) are written to disk, one array per view and data split; the store of the three tasks (about $18{,}000$ images, $354$ GB) was built in $22$ minutes on one H100 node. A lattice of $R\times R$ views reads $R^2$ of the stored views (sixteen for $56\times56$), and training reads an epoch as a block from the memory-mapped arrays, so the surrogate trains without touching the backbone. The truncated trunk followed by the last block reproduces the torchvision forward pass exactly (maximum logit difference $0$), and the top-1 accuracy of the backbone on the full validation set is $80.8\%$ ($81.1\%$ in the torchvision table; our images come from a copy of ImageNet that is already resized to $256$ px, which may explain the difference).

\paragraph{Compute.}\label{app:impl} One fit of $40$ epochs on the $56\times56$ lattice takes about ten to twenty-five minutes on an A100 shared by two processes, from the stored descriptors. The training step is compiled with \texttt{torch.compile}, which makes it $2$--$2.7\times$ faster on the GPUs we timed (RTX 5060 laptop, RTX 4090, V100 and GH200), more where kernel launches dominate; if compilation fails in the first calls, the code prints a warning and falls back to eager execution, and the report records the mode that ran. Compiling the frozen backbone does not pay: it is $3$--$26\%$ slower. TF32 would speed the backbone up ($1.5$--$3.7\times$) but changes the cached descriptors slightly, so the defaults are a compiled training step and an eager FP32 backbone.

\paragraph{The three costs.} Preprocessing, training and explanation cost different amounts. The store is built once for a set of classes: the $64$ views of the $18{,}000$ images of the three tasks take $22$ minutes on one H100 and $354$ GB of disk; the $56\times56$ lattice reads $16$ of the $64$ views, a quarter of that. A fit takes ten to twenty-five minutes from the store, per class and seed. Explaining a new image runs the frozen trunk on its $16$ shifted views and the surrogate on the lattice, $46$ ms per image on a V100, against $7$ ms for the teacher's forward pass, $20$ ms for one gradient, $62$ ms for AttnLRP and $660$ ms for integrated gradients with $32$ steps (\texttt{fig1\_attnlrp\_cost.py}). The dense maps of the figures need $256$ views, about $16$ times as much, and are made for display only (\cref{app:finer}).

\section{The monotone readout}
\label{app:readout}
The surrogate of \cref{sec:model} reads the used maps through the following CNN; its monotonicity fixes the signs of the groups, and the card of \cref{eq:card} keeps the groups with $\gamma_k>0$.

The map $\tilde A\in\mathbb R^{n\times56\times56}$ goes through three $3\times3$ convolutions of width $48$ with kernels $\operatorname{softplus}(\Theta_r)\ge0$ (biases unconstrained). Each is followed by a per-channel nondecreasing activation
\begin{equation}
\begin{aligned}
\phi(t)={}&a_1\operatorname{softplus}(b_1(t-m_1))\\
&-a_2\operatorname{softplus}(b_2(m_2-t))+a_3\sigma(b_3(t-m_3)),
\end{aligned}
\label{eq:act}
\end{equation}
with $a_j\ge0$, $b_j>0$ and $\sigma$ the logistic function. Global mean pooling follows, then a positive affine map gives the logit $F=\sum_c\operatorname{softplus}(\theta_c)\,\overline{h}_c+\beta$. The readout sees only group maps and their arrangement: no image, no continuous descriptor, no class token, no teacher score.

This construction is what gives the signs their meaning. Every term of $\phi'$ is nonnegative, and kernels, pooling and the output weights are nonnegative. Hence $F$ is nondecreasing in every coordinate of $\tilde A$, and with \cref{eq:coef}, wherever derivatives exist,
\begin{equation}
\gamma_k>0\Rightarrow\frac{\partial F}{\partial A_k(u,v)}\ge0,\quad 
\gamma_k<0\Rightarrow\frac{\partial F}{\partial A_k(u,v)}\le0.
\label{eq:sign}
\end{equation}
For hard maps, with all other sites fixed, this says that moving one site from a discard channel to a used channel $k$ cannot lower the score if $\gamma_k>0$ and cannot raise it if $\gamma_k<0$ (the converse move reverses the inequality). Moving a site between two used channels changes two coordinates, and \cref{eq:sign} does not fix the sign of that total effect. The sign of $\gamma_k$ may change during training, so the property refers to the final coefficients of the selected checkpoint.

\section{Further related work}
\label{app:related}
\paragraph{Prototype models on ImageNet.}
We do not show the maps of ProtoPNet~\cite{chen2019looks}, PIP-Net~\cite{nauta2023pipnet}, ProtoViT~\cite{ma2024protovit} or ProtoPFormer~\cite{xue2024protopformer} next to ours, because we found no public ImageNet-1k checkpoint of any of them: the published checkpoints of PIP-Net (four) and ProtoViT (three) are trained on CUB-200 and Stanford Cars, and ProtoPNet and ProtoPFormer release none. Running them on beagle and school bus would mean training new models on ImageNet, that is, explaining a different classifier built for this purpose, not the frozen ViT whose decision we explain. The post-hoc methods that report ImageNet-1k, Tan et al.~\cite{tan2024posthoc}, EPIC~\cite{borycki2026epic}, InfoDisent~\cite{struski2024infodisent} and SIDE~\cite{dubovik2025side}, are closer in spirit, but we found no code for Tan et al. and SIDE, the code of EPIC and InfoDisent is public without trained weights, EPIC applies to networks with a pooling and a single linear head, not to a ViT with a class token, and InfoDisent replaces the head by a trained one and lowers the ImageNet accuracy (from $81.1\%$ to $79.2\%$ for ViT-B/16). A model with parts of this kind whose released checkpoints cover one of our classes is PDiscoFormer~\cite{aniraj2024pdiscoformer} (school bus; none covers beagle), which we ran as released (\cref{app:pdisco}). Our comparison of explanations of the same decision is with attribution methods that run on the same frozen teacher (\cref{sec:compare}).

\paragraph{Concept decompositions of the teacher's logit.}
CRAFT~\cite{fel2023craft}, MCD~\cite{vielhaben2023mcd} and DRSA~\cite{chormai2024drsa} decompose the logit of the explained network itself into concept contributions with local maps, which is the faithful alternative to a surrogate. Their released code is written for CNN activations: CRAFT factorizes nonnegative activations by NMF, the completeness of MCD needs a linear head on pooled features, and DRSA uses LRP rules for convolutional networks. Running them on a ViT would mean changing the methods, so we compare instead with baselines built from the same components: k-means clusters of the same descriptors ranked by the teacher's response to their replacement (\cref{app:compare}), and the prototypes frozen at the k-means start with an additive readout (\cref{tab:readout}).

\paragraph{Interpretable by design and surrogates.}
B-cos networks~\cite{bohle2022bcos} make the explanation a property of the architecture. LIME and Anchors~\cite{ribeiro2016lime,ribeiro2018anchors} fit local surrogates on superpixels of one image, whereas our surrogate is global for a class and is trained on cached tokens.

\paragraph{Resolution of ViT features.}
Shift-and-stitch~\cite{sermanet2014overfeat,long2015fully} obtains dense outputs from a strided network by shifting its input. Learned feature upsamplers~\cite{fu2024featup,huang2025loftup,couairon2025jafar,wimmer2026anyup} (FeatUp trains with jittered views and a TV term) and dense ViT descriptors~\cite{amir2022deep,caron2021emerging,oquab2024dinov2} raise the usable resolution of ViT features. Tokens of high norm are an artifact with known fixes~\cite{darcet2024registers,yang2024denoising}. Shift equivariance and aliasing are studied for CNNs and ViTs~\cite{zhang2019making,rojasgomez2024making}. We have not compared our lattice with learned upsamplers on standard probes.

\paragraph{Evaluating explanations.}
Common tests are the pointing game, insertion and deletion and their refinements~\cite{zhang2018top,petsiuk2018rise,rong2022road,wu2024faithfulness} and retraining-based removal and sanity checks~\cite{hooker2019benchmark,adebayo2018sanity}; they rank continuous maps and our groups are regions, so we compare at equal area (\cref{sec:compare}); ground-truth masks come from ImageNet-S and PartImageNet~\cite{guillaumin2014imagenet,gao2022large,he2022partimagenet}.

\paragraph{Token grouping and part discovery.}
Unsupervised grouping of frozen ViT tokens, including object discovery and segmentation~\cite{simeoni2021lost,wang2022tokencut,melaskyriazi2022deep,hamilton2022stego,wang2023cutler}, is class-agnostic and follows feature affinity. Our groups follow a given per-class decision of a frozen classifier.

\section{Lattices, dense maps and the models of the tables}
\label{app:finer}
\paragraph{The lattice and the TV term.} The frozen ViT (stride $16$) is run on views shifted by offsets $(a,b)\in\{-4,0,4,8\}^2$ pixels with reflection padding $(4,8,4,8)$ (every view is a $224\times224$ window, the native input size), which gives sixteen $14\times14$ grids. The token $(i,l)$ of view $(a,b)$ is placed at site $(4i+a',4l+b')$ of a $56\times56$ lattice, where $a',b'\in\{0,1,2,3\}$ index the offsets; site $u$ is centered at pixel $3.5+4u$. The classifier, the TV term and the checkpoint selection all use this lattice; all $64$ stored views give a $112\times112$ lattice with $2$-pixel spacing. Because the group of a token depends on that token alone, the trained prototypes also apply to views that the surrogate never saw; TV, which ties neighboring sites of different views, makes this slightly more consistent ($79\%$ of the pixels at unseen shifts keep the label of the nearest trained site, $76$--$77\%$ without TV). For a stack of maps $M$ with lattice spacing $\Delta$ pixels ($\Delta=4$ on the $56\times56$ lattice) the TV per original $16$-pixel patch is
\begin{equation}
\TV_\Delta(M)=\frac{16}{\Delta}\left(\operatorname{mean}|D_uM|+\operatorname{mean}|D_vM|\right),
\label{eq:tv}
\end{equation}
where the means run over channels and sites and $D_u$, $D_v$ are the differences between neighboring sites.

\begin{figure}[tp]
\centering
\includegraphics[width=\columnwidth]{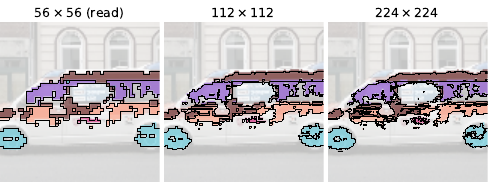}
\caption{One surrogate, three renderings (ambulance of \cref{fig:teaser}): the $56\times56$ lattice that the readout reads, and the same prototypes on $64$ and $256$ shifted views, shifts never seen in training.}
\label{fig:res}
\end{figure}

\begin{figure*}[tp]
\centering
\includegraphics[width=\textwidth]{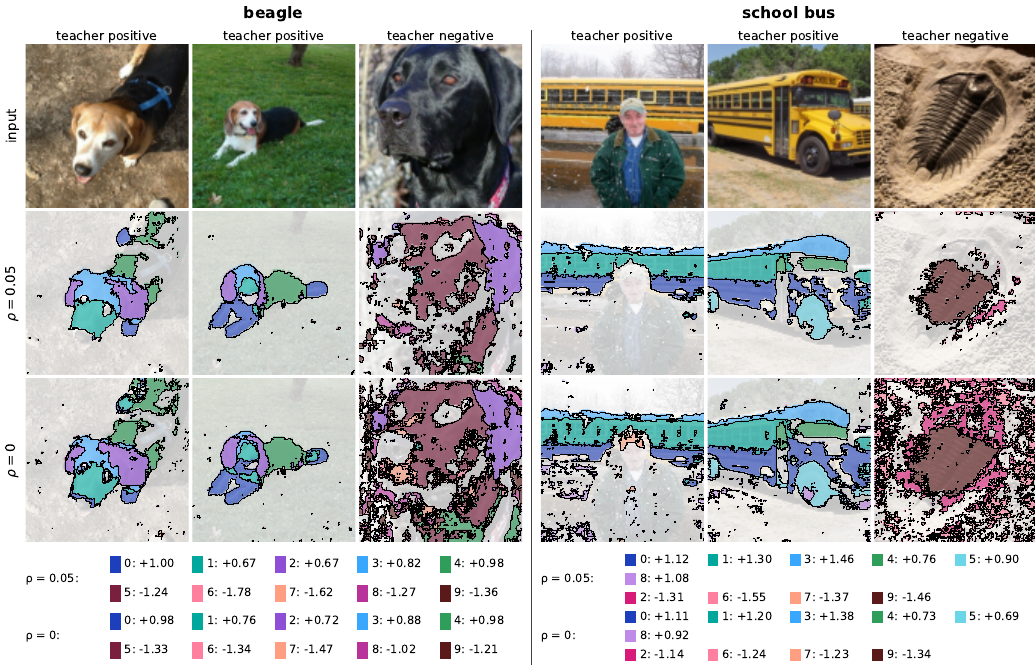}
\caption{The $56\times56$ models of the tables (seed $101$, $p=0.25$, best-fidelity checkpoint) with TV ($\rho=0.05$, middle row) and without it ($\rho=0$, bottom row) on the images of \cref{fig:gallery} (second school-bus image: the bus of \cref{fig:teaser}), as dense maps: every used group in its own color (cool: positive coefficient, warm: negative; the key gives channel and coefficient), discard channels uncolored, a thin black line around every used group. A color names a channel of one model only. The images were fixed before any map was drawn.}
\label{fig:gallery-56}
\end{figure*}

\paragraph{The $112\times112$ lattice.}
The store holds all sixty-four views (offsets $-6,-4,\dots,8$ pixels per axis, reflection padding $(6,8,6,8)$), so the $112\times112$ lattice (\cref{sec:lattice}, sites $2$ pixels apart) can be trained from it in the same protocol. We ran it for beagle and school bus with one seed ($101$), $\rho=0.05$ and $\rho=0$, and compare it in \cref{tab:finer} with the $56\times56$ model of the same seed. The tables of the main text use the $56\times56$ lattice, because all of their numbers have three seeds; the appendix figures of \cref{app:collage,app:examples,app:pdisco,app:discard} show the $112\times112$ models of seed $101$, which are close to the $56\times56$ models in fidelity, area and box precision (\cref{tab:finer}). On beagle seed $101$ is more precise in the boxes than the mean of the three seeds ($99.4\%$ against $95.0\%$ with TV and $99.3\%$ against $91.1\%$ without, \cref{tab:loc}), so the pictures of the main text show a favorable seed for localization on beagle, and the box numbers of \cref{tab:finer} should not be compared with the three-seed numbers of \cref{sec:loc}; on school bus seed $101$ is typical ($94.7\%$ against $94.7\%$ with TV, $82.5\%$ against $81.1\%$ without).

\begin{table}[tp]
\centering
\small
\caption{The $112\times112$ lattice against the $56\times56$ lattice, same seed ($101$), best-fidelity checkpoint, $p=0.25$; single runs. Prec., recall and lift: positive groups against the ImageNet object boxes (\cref{sec:loc}).}
\label{tab:finer}
\fitcolumn{\begin{tabular}{@{}llccccccccc@{}}
\toprule
task & $\rho$ & lattice & fid.\ (\%) & TV & area (\%) & TV/area & alive & prec.\ (\%) & recall (\%) & lift\\
\midrule
beagle & 0.05 & $56$ & 99.39 & 0.412 & 23.6 & 3.48 & 9 & 99.4 & 39.7 & 1.39\\
 & & $112$ & 99.56 & 0.526 & 25.4 & 4.14 & 8 & 99.4 & 41.9 & 1.38\\
 & 0 & $56$ & 98.84 & 0.710 & 44.2 & 3.21 & 10 & 99.3 & 44.5 & 1.38\\
 & & $112$ & 98.85 & 1.001 & 43.7 & 4.58 & 10 & 99.4 & 44.6 & 1.38\\
school bus & 0.05 & $56$ & 99.83 & 0.305 & 17.4 & 3.52 & 7 & 94.7 & 48.0 & 1.58\\
 & & $112$ & 99.82 & 0.334 & 14.8 & 4.50 & 6 & 94.1 & 44.6 & 1.57\\
 & 0 & $56$ & 99.74 & 0.779 & 43.4 & 3.59 & 10 & 82.5 & 60.7 & 1.38\\
 & & $112$ & 99.57 & 1.109 & 42.9 & 5.17 & 10 & 82.4 & 60.4 & 1.37\\
\bottomrule
\end{tabular}}
\end{table}

Fidelity at $112\times112$ is within one test positive of the $56\times56$ model in every row (one positive is $0.78$ points on beagle and $0.89$ on school bus), the used area and the number of alive channels are close, and the box measures do not improve: precision is $99.4\%$ against $99.4\%$ on beagle and $94.1\%$ against $94.7\%$ on school bus with TV, and the lift differs by at most $0.01$ in every pair. TV and TV per area are higher at $112\times112$, where TV is measured at a finer resolution; we suspect boundaries that are ragged at the scale of two pixels, which the coarser grid cannot see, but we did not test this. The finer lattice thus samples the descriptor field more densely without tightening the groups to the object. With one seed per setting these are single runs and we describe only the pattern.

\paragraph{Dense maps.} The group maps of \cref{fig:teaser,fig:gallery,fig:gallery-56,fig:deit,fig:discard} are dense $224\times224$ maps (\cref{fig:negatives} shows the native lattice), computed from the trained models without retraining, outside the store. The group of a token is the hard argmax of \cref{eq:softmax}, so it can be measured at any shift of the frozen backbone: $16\times16=256$ views, shifted by $-7,\dots,8$ pixels per axis with reflection padding $(7,8,7,8)$, give every pixel one label, the token of the view whose center is that pixel. The readout is not applied, so the decisions and the TV values quoted with these figures are those of the trained lattice; the used area of a dense map differs from that of the lattice by less than two points on the images shown. At the sites of the trained lattice the dense labels equal the lattice labels (they differ at no more than $0.1\%$ of the sites of an image), and elsewhere they agree with the label of the nearest lattice site for $82\%$ (beagle) and $84\%$ (school bus) of the pixels ($112\times112$ models, $50$ test images each). For the $56\times56$ models of the main text, whose lattice is coarser, the agreement is $79\%$ on both tasks with TV and $77\%$ and $76\%$ without (the $93$ beagle and $99$ school-bus images of the collages, \texttt{shift\_consistency.py}), and the positive region of the dense map overlaps the one of the lattice with an IoU of $0.71$ and $0.77$.

\paragraph{The $56\times56$ models of the tables.}\label{app:notv} The figures of the main text show the $56\times56$ TV models of the tables as dense maps; \cref{fig:gallery-56} shows the same models with TV ($\rho=0.05$) and without it ($\rho=0$, same dropout) on the images of \cref{fig:gallery}, but with the bus of \cref{fig:teaser}, which \cref{fig:gallery} skips, as the second school-bus image. Without TV the used area of the lattice is larger on every image shown (first beagle image $24.0\%$ against $20.1\%$, black dog $64.3\%$ against $42.7\%$; first school-bus image $28.3\%$ against $23.0\%$, fossil $52.3\%$ against $14.5\%$), and the hard TV is higher on five of the six (black dog $1.18$ against $0.79$, fossil $0.99$ against $0.27$; the exception is the second beagle image, $0.13$ against $0.14$). The positive groups lie alike in the two models; without TV the negative groups take more of the background, and small specks of all groups appear around the object.

\section{Do three seeds give the same groups?}
\label{app:seeds}
\begin{figure}[tp]
\centering
\includegraphics[width=\columnwidth]{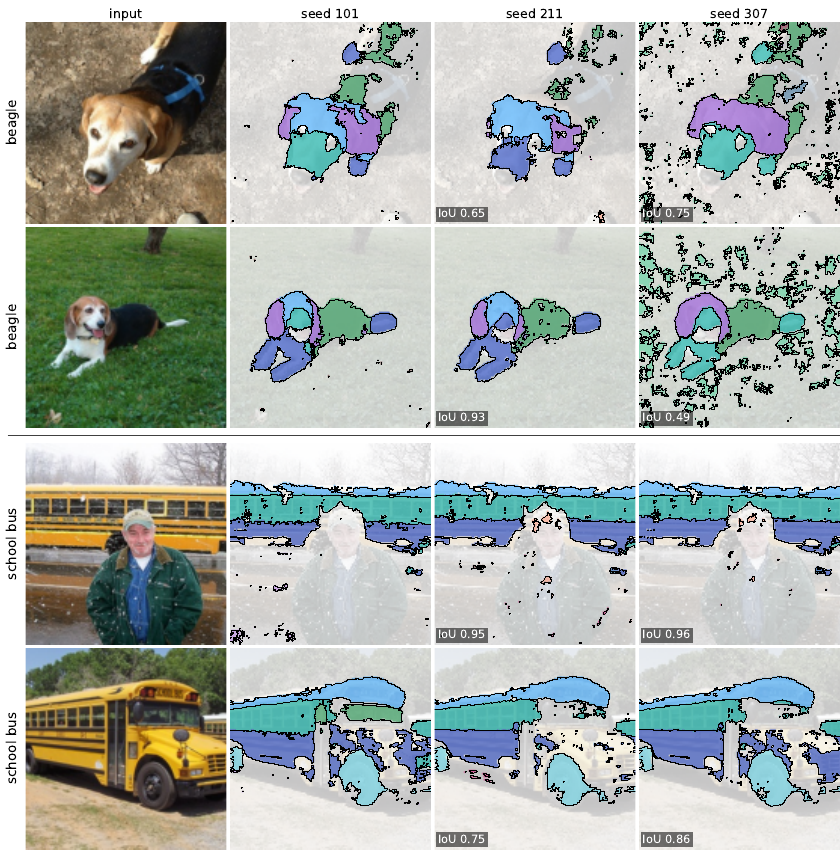}
\caption{The $56\times56$ TV models of the three seeds of the tables (best-fidelity checkpoint) on two teacher-positive test images per class, dense maps drawn as in \cref{fig:gallery-56}. The groups of seeds $211$ and $307$ have the color of their counterpart in seed $101$ (\cref{app:seeds}); positive groups without a counterpart have colors that seed $101$ does not use. Corner: IoU of the positive region with that of seed $101$ on the same image.}
\label{fig:seeds}
\end{figure}

The three seeds of the tables imitate the teacher about equally well (balanced fidelity $99.39$, $99.52$ and $98.82\%$ on beagle, $99.83$, $99.65$ and $97.19\%$ on school bus), but a surrogate could reach that with different groups in every run. We therefore compare the $56\times56$ TV models of the three seeds on the lattice labels of all teacher-positive test images ($64$ beagle and $56$ school-bus images), for every pair of seeds, against a null in which the second seed is taken on another positive image ($200$ random pairings); the measures were fixed before we looked at the result. The positive region of two seeds (the union of their positive groups) overlaps with an IoU of $0.55$ on beagle and $0.75$ on school bus (null $0.16$ and $0.17$). Matching the used groups of two seeds one to one by the pixels they share (Hungarian assignment), $73\%$ (beagle) and $87\%$ (school bus) of the pixels that are positive in both seeds lie in matched groups (null $29\%$ and $35\%$), and the NMI of the two partitions on these pixels is $0.67$ and $0.83$ (null $0.02$ and $0.08$). These two measures leave out the pixels that only one seed calls positive, where the seeds disagree most; the IoU of the positive regions above counts them. In every pair of seeds all but at most two positive groups have a counterpart with a pooled Jaccard index of $0.33$ to $0.85$ (null about $0.05$), always of the same sign; the others have none (Jaccard below $0.15$). \Cref{fig:seeds} shows the three seeds on two images per class. On school bus the decompositions are nearly the same (roof, rows of windows, lower body, wheels). On beagle the positive region is much the same in seeds $101$ and $211$, but the cut into parts differs: the muzzle group of seed $101$ has no counterpart in seed $211$, and seed $307$, the one with the lowest fidelity, has a positive group of its own that also takes specks of grass. The seeds thus agree on where the evidence lies and, on school bus, on how it is split; on beagle the split into parts is only partly reproducible, in line with the weak positional identity of the beagle groups measured below. That identity is measured on all test images that show the class ($43$ beagle and $49$ school-bus images): every lattice site of a positive group inside the object frame (the bounding box of the ImageNet boxes) gets coordinates normalized to the frame, and $R^2$ is the share of their variance explained by the group identity (one-way analysis of variance with the groups as classes; null: identities permuted within each image); with TV $0.09$ (beagle) and $0.29$ (school bus), without TV $0.08$ and $0.25$, null $0.01$ (means over three seeds).

\section{The groups as collages}
\label{app:collage}
\begin{figure*}[tp]
\centering
\includegraphics[width=.96\textwidth]{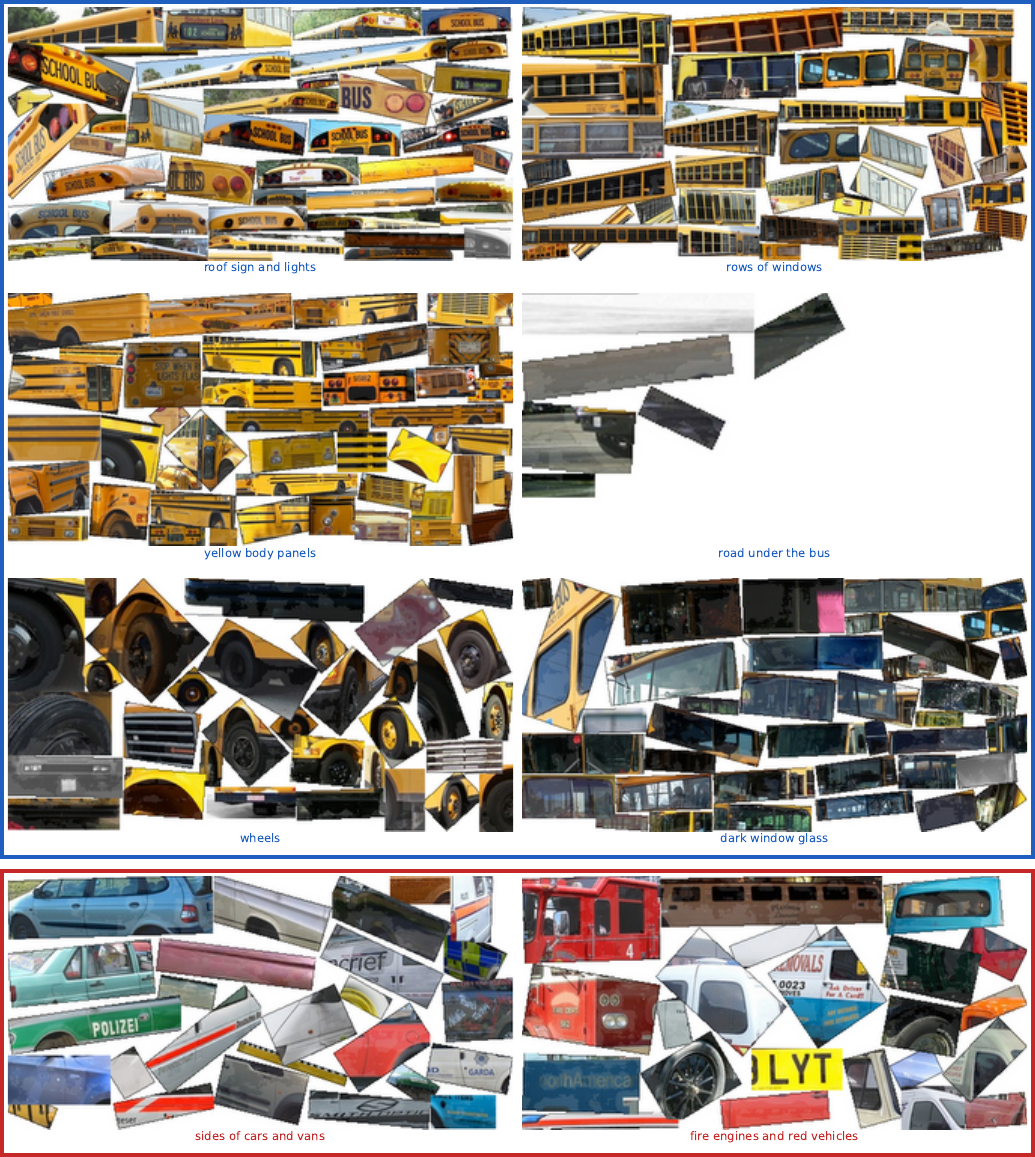}
\caption{The groups of the school-bus model as collages of all test images (the first images by index): positive groups (blue frame) from the $49$ teacher-positive images of the class, negative groups (red frame) from $50$ negative images of sibling classes; the label of a sheet is our name of the group.}
\label{fig:collage-all}
\end{figure*}

\Cref{fig:collage-all} shows all used groups of the surrogate (\cref{fig:collage,fig:collage-both}) with all test images: the $49$ teacher-positive images of the class for the positive groups and $50$ negative images of sibling classes (road vehicles) for the negative groups, the first ones by index, chosen before any map was looked at. A group appears in an image if its largest piece covers at least $1\%$ of the crop. A piece is the smallest rotated rectangle around the largest piece of the group, cut from the image in its orientation and pasted without rotation; the pixels of the rectangle outside the group are washed out by $12\%$, the scale is common to all pieces of a sheet and chosen as large as possible with at most $3\%$ of the piece area overlapping. The model is the $56\times56$ school-bus model of the tables (seed $101$), drawn as a dense map (\cref{fig:res}). Two negative groups (channels $2$ and $6$) have a piece of $1\%$ in only one of the $50$ sibling-class images and are left out; the road group has only $6$ pieces, because it needs a visible road under the bus.

\begin{figure*}[tp]
\centering
\includegraphics[width=.96\textwidth]{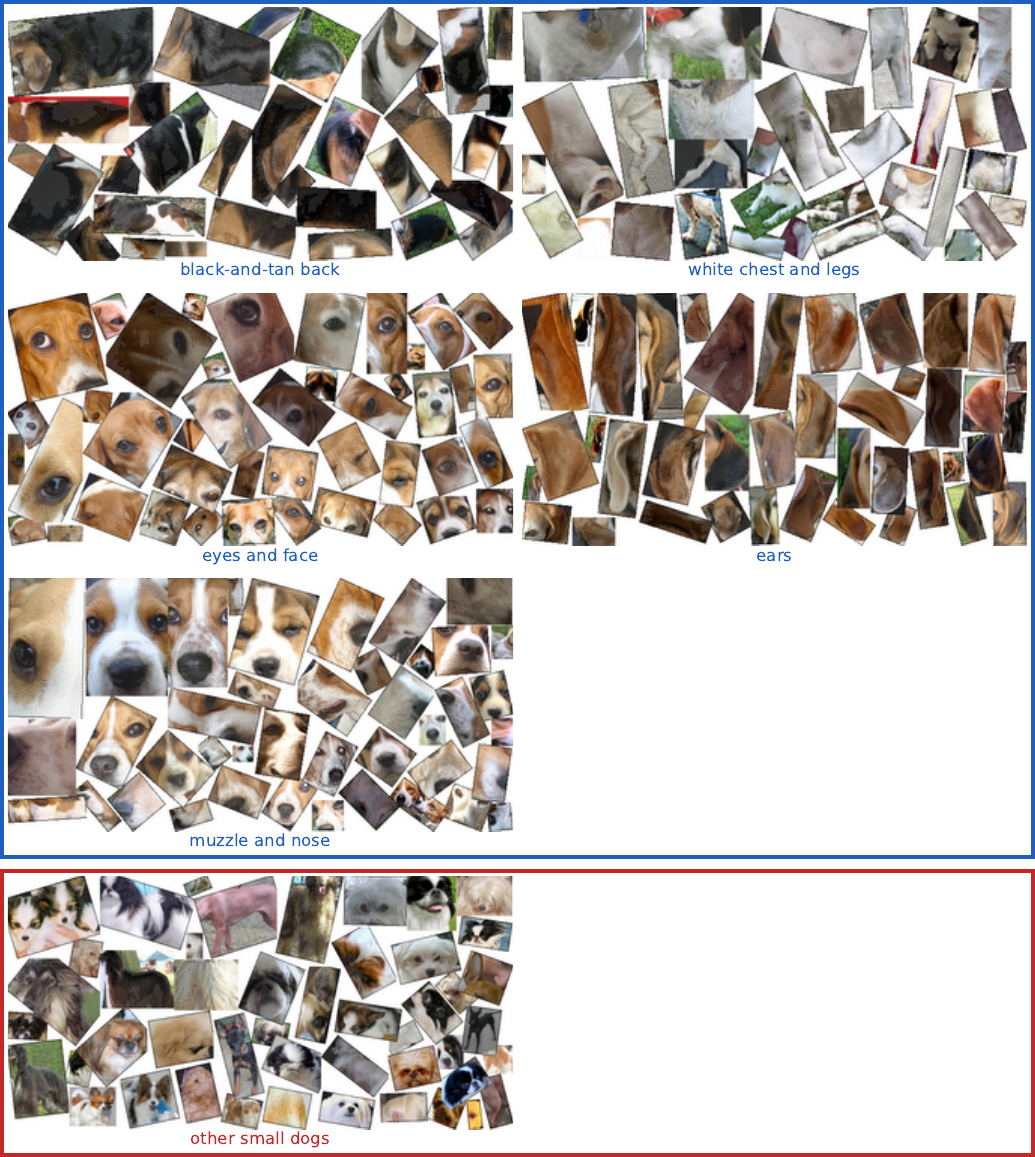}
\caption{The groups of the beagle model as collages of all test images of the class (positive groups, blue frame) and of $50$ negative images of other dog breeds (negative group, red frame), as in \cref{fig:collage-all}; the names are ours.}
\label{fig:collage-beagle}
\end{figure*}

\Cref{fig:collage-beagle} gives the same for the $112\times112$ beagle model of seed $101$ (\cref{app:finer}): the $43$ teacher-positive images of the class for the positive groups and the first $50$ negative images of other dog breeds (sibling classes) for the negative ones. The groups are, by their pieces, the black-and-tan back, the white chest and legs, the eyes and the face, the ears and the muzzle with the nose; the strongest negative group collects other, mostly small and fluffy dogs. The other four negative groups (channels $6$ to $9$) have a piece in at most three of the sibling-class images and are not shown.

\section{More images}
\label{app:examples}
\begin{figure*}[tp]
\centering
\includegraphics[width=\textwidth]{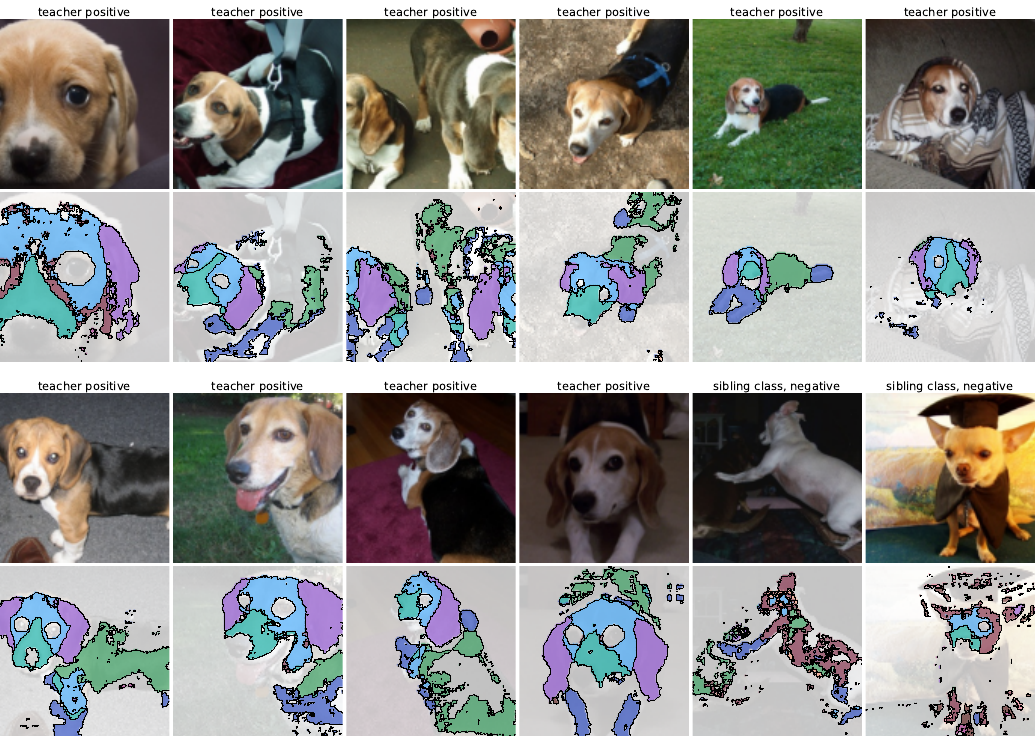}
\caption{Beagle: the $112\times112$ TV model of seed $101$ (\cref{app:finer}) on the first ten teacher-positive test images by index and the first two sibling-class negatives (other dog breeds), drawn as in \cref{fig:gallery-56}; images chosen before any map was drawn.}
\label{fig:more-beagle}
\end{figure*}

\begin{figure*}[tp]
\centering
\includegraphics[width=\textwidth]{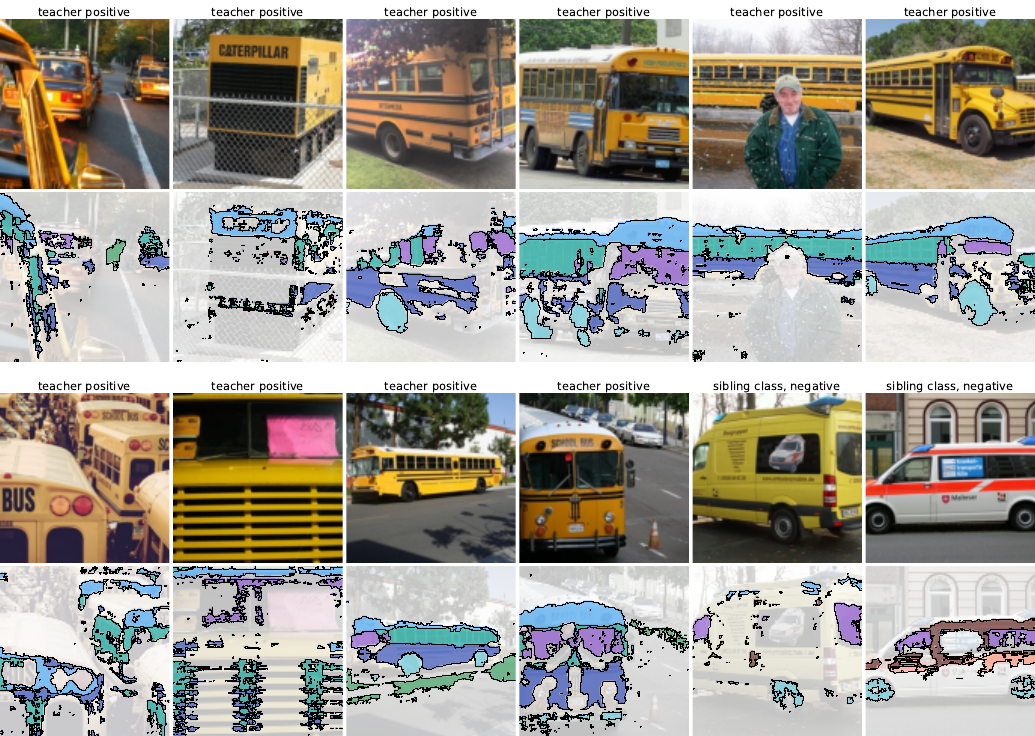}
\caption{School bus, as \cref{fig:more-beagle}; the sibling-class negatives are road vehicles (here two ambulances).}
\label{fig:more-bus}
\end{figure*}

\begin{figure*}[tp]
\centering
\includegraphics[width=\textwidth]{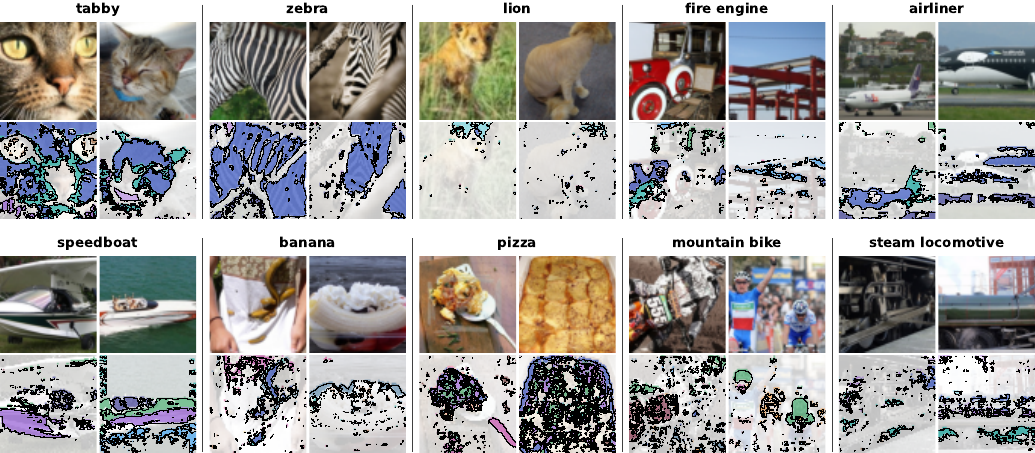}
\caption{The ten further classes of \cref{app:classes} ($56\times56$ TV models, seed $101$) on their first two teacher-positive test images by index, chosen before any map was drawn; dense maps drawn as in \cref{fig:gallery-56}.}
\label{fig:more-ten}
\end{figure*}

\Cref{fig:more-beagle,fig:more-bus} show the $112\times112$ models of seed $101$ (\cref{app:finer}) on the first ten teacher-positive test images of the class by official index and on the first two teacher-negative test images of a sibling class, chosen by this rule before any map was drawn (the images of \cref{fig:gallery} are among them). On beagle the same five positive groups recur on every dog of the class: the eyes and the face, the ears, the muzzle with the nose, the black-and-tan back and the white chest and legs (named as in \cref{app:collage}). On the two sibling-class negatives most of the colored area is negative, but the chihuahua also takes groups of the head. On school bus the first two teacher-positive images show a taxi and a generator, which the teacher also calls a school bus (\cref{sec:protocol}); on the buses the roof, the rows of windows, the lower body and the wheels recur, and on the two ambulances, sibling-class negatives, the dark-glass and the wheel groups recur. Over all test images this is the rule for the parts that buses share with other vehicles ($56\times56$ models of the tables, seed $101$; a group occurs when it covers at least $1\%$ of the crop; \texttt{positives\_on\_negatives.py}): the wheel and dark-glass groups occur on $78\%$ and $63\%$ of the $82$ sibling-class negatives and on under $1\%$ of the other negatives, while the bus-specific roof sign and yellow panels occur on $9\%$ and $11\%$; on the sibling-class negatives positive and negative groups cover about the same area ($8.0\%$ and $6.1\%$ of the crop). On beagle each positive group occurs on $22$ to $32\%$ of the $470$ other dog breeds (on at most $1\%$ of the other negatives), and there the negative groups cover more ($17.0\%$ against $5.0\%$). The surrogate thus finds the shared parts on the negatives as well, and its readout weighs them against the negative groups. \Cref{fig:more-ten} shows the ten further classes of \cref{app:classes} on their first two teacher-positive test images: on some the groups are compact parts of the object (zebra, tabby cat, airliner), on others they are fragmented (pizza, banana, mountain bike) or nearly empty (lion), in line with the spread of \cref{tab:classes}.

\section{Do the groups coincide with annotated parts?}
\label{app:parts}
\Cref{sec:groups} reads the groups as parts of the object. PartImageNet++~\cite{li2024partimagenetpp} annotates parts of ImageNet-1k training images with polygons: beagle head, body, tail and paw, and school bus wheel, stop sign, door, driver cabin, students cabin and label. For the first $100$ annotated images of each class we paint the hard groups of the $56\times56$ surrogates of the tables (three seeds, with and without TV) to the pixels of the standard crop and compare them with the part mask of the same crop (\texttt{part\_purity.py}). We report the normalized mutual information (NMI) between the group and the part label on the pixels that belong to an annotated part, the same after moving every part mask by a random torus shift (null), the purity of the positive groups (the share of the part pixels of a group that lie in its most frequent part, weighted by area) and their precision (the share of their pixels that lie in some part; the base rate is the share of part pixels in the crop). The images are training images of ImageNet-1k; four in five of them were among the images the surrogates were fitted on, and the rest only took part in choosing the checkpoint (below); the part masks were never used.

\begin{table}[!htbp]
\centering
\small
\caption{Groups against annotated parts (PartImageNet++, first $100$ images per class), mean over three seeds. NMI between group and part on part pixels; the same on the held-out images only ($22$ beagle and $25$ school-bus images of the selection split); with shifted part masks (null); and with the groups of another image of the class on the parts of each image (position null). Purity and precision of the positive groups (base rate of part pixels in the crop: $.43$ and $.56$). The NMI of single seeds ranges over $.086$--$.167$ (beagle, TV) and $.088$--$.109$ (school bus, TV).}
\label{tab:parts}
\fitcolumn{\begin{tabular}{llcccccc}
\toprule
class & & NMI & held out & null & position null & purity & precision\\
\midrule
beagle & TV & $.116$ & $.105$ & $.007$ & $.007$ & $.79$ & $.90$\\
 & no TV & $.109$ & $.098$ & $.007$ & $.007$ & $.73$ & $.82$\\
school bus & TV & $.096$ & $.105$ & $.008$ & $.009$ & $.55$ & $.95$\\
 & no TV & $.092$ & $.102$ & $.008$ & $.010$ & $.51$ & $.82$\\
\bottomrule
\end{tabular}}
\end{table}

\Cref{tab:parts} shows that the groups follow the annotated parts far above chance: the NMI is $12$ to $17$ times its null, and the positive groups lie in part pixels $90$ to $95\%$ of the time with TV, where parts cover $43$ and $56\%$ of the crop. The absolute NMI is low, because ten groups are compared with four or six parts of a different granularity, and TV changes it less than the spread over seeds; it raises purity and precision a little. The table of groups against parts (seed $101$, TV) shows how the two cut the object. On beagle three positive groups lie $85$ to $92\%$ in the head, one lies $83\%$ in the body and one splits into the body ($47\%$) and the paws ($39\%$); the tail is not claimed by any group. On school bus the annotation is coarse: the students cabin and the driver cabin are large regions of the body, and the groups that name the windows, the roof lights and the yellow panels lie $83$ to $91\%$ in these two together, so the purity is lower; the wheels group lies $49\%$ in the wheels, and the road under the bus lies $98\%$ outside every annotated part, as the annotation has no label for the road. The groups are thus mostly finer than the school-bus annotation and cross-cut it, and on beagle they follow it.

\emph{Held-out images, position and redundancy} (\texttt{part\_labels.py}, \texttt{part\_analysis.py}). Every fifth image of the sorted class folder belongs to the selection split, on which no surrogate was fitted (it only chose the checkpoint, by fidelity): $22$ of the $100$ beagle and $25$ of the $100$ school-bus images. On them the NMI with TV is $.105$ on both tasks (fitted images: $.122$ and $.096$), against a null of $.018$ and $.020$ (higher with fewer images), and with a bootstrap over the held-out images the NMI minus the null is above zero for every seed (smallest lower bound $.031$). The shifted masks move the parts away from where they lie; the position null keeps where groups and parts tend to lie in the crop and breaks only their correspondence in an image, pairing the groups of one image with the parts of another (a random derangement of the images, $20$ draws). It gives $.007$ and $.009$, no more than the shifted masks, so the agreement with the parts is not explained by position alone. The groups split the larger parts: with TV the positive groups cover $37$ to $48\%$ of each beagle part and $41$ to $58\%$ of each bus part except the stop sign ($1\%$); the covered pixels of the beagle head are shared by an effective $2.9$ groups (the exponential of the entropy of their shares; three groups have the head as their dominant part), those of the two cabins of the bus by $3.5$ and $3.8$, while the wheels belong to one group ($1.0$). On beagle, then, several groups name the head, and the semantic identity of a group is not one annotated part.

\section{PDiscoFormer}
\label{app:pdisco}
PDiscoFormer~\cite{aniraj2024pdiscoformer}, the successor of PDiscoNet~\cite{vanderklis2023pdisconet}, is the closest published model with code and checkpoints for ImageNet classes. It is a ViT (DINOv2 ViT-B/14 with four registers) with $K$ learned part prototypes. Every patch is assigned softly (a Gumbel softmax) to one of the $K$ parts or to a background channel, the feature of a part is the average of the patch features under its assignment, and a linear classifier reads the $K$ part features: the logit of a class is the mean of its part logits. The model is trained end to end for classification with losses that make the parts compact, orthogonal and consistent, so it is interpretable by design: it explains its own decision, not the one of a frozen ViT-B/16.

\paragraph{How it was run.} We took the checkpoints that the authors released for PartImageNet-OOD ($K=8$, $25$ and $50$ parts, $109$ ImageNet classes, school bus among them; no checkpoint covers beagle) and the code of their repository, unmodified. The repository's \texttt{\_from\_pretrained} asks for two arguments that the current \texttt{huggingface\_hub} no longer passes, so we fetch the two files of a checkpoint and call it on the local folder with those arguments set to \texttt{None}; all tensors of the file are used (\texttt{pdisco\_bus\_eval.py}). The input is the standard $224$ crop of the first $100$ PartImageNet++ images of school bus used in \cref{app:parts}, normalized as in the repository's test transform; as a check, the model names school bus as its top class on $94$ to $95\%$ of them. The forward pass is stochastic (Gumbel noise also at test time), so we average three draws; they differ by at most $0.002$ in NMI. The part of a pixel is the argmax of the upsampled soft assignment, the background channel included, as in the repository's own evaluation of part NMI. Our metrics are those of \cref{app:parts}, computed from these labels only.

\begin{table}[!htbp]
\centering
\small
\caption{School bus against the part annotation of PartImageNet++ (first $100$ images): AtomEx (mean over three seeds; purity and precision of the positive groups, area of the positive groups of seed $101$) and the released PDiscoFormer (PDisco; mean over three noise draws, all non-background parts). Labels: all groups or parts, including discard channels or background. Base rate of part pixels: $.56$.}
\label{tab:pdisco}
\setlength{\tabcolsep}{3.5pt}
\begin{tabular}{lcccccc}
\toprule
 & labels & NMI & null & purity & prec. & area\\
\midrule
AtomEx, TV & $20$ & $.096$ & $.008$ & $.55$ & $.95$ & $.31$\\
PDisco $K{=}8$ & $9$ & $.088$ & $.009$ & $.61$ & $.92$ & $.24$\\
PDisco $K{=}25$ & $26$ & $.171$ & $.015$ & $.56$ & $.88$ & $.61$\\
PDisco $K{=}50$ & $51$ & $.170$ & $.011$ & $.51$ & $.82$ & $.66$\\
\bottomrule
\end{tabular}
\end{table}

\begin{figure*}[tp]
\centering
\includegraphics[width=\textwidth]{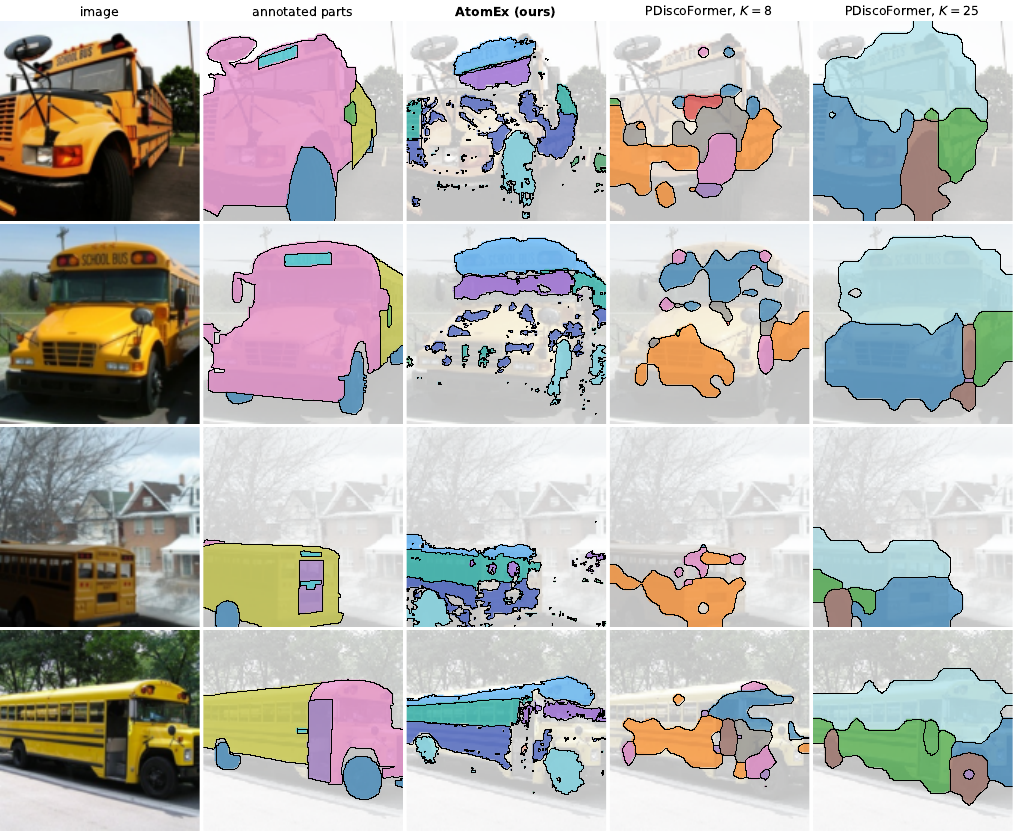}
\caption{AtomEx and the released PDiscoFormer on the first four school-bus images of \cref{tab:pdisco}: the image, the annotated parts of PartImageNet++ (wheel blue, driver cabin pink, students cabin olive, door violet, stop sign green, label cyan), the groups of the AtomEx $112\times112$ TV model of seed $101$ (\cref{app:finer}) on the same crops (drawn as in \cref{fig:gallery-56}), and the parts of PDiscoFormer with $K=8$ and $K=25$ (argmax of its soft assignment, background channel uncolored). Colors are not comparable between columns.}
\label{fig:pdisco}
\end{figure*}

\paragraph{Result.} \Cref{tab:pdisco} gives the same measures for both. With $8$ parts PDiscoFormer is level with AtomEx in NMI ($.088$ against $.096$, ten and twelve times the null), its parts are a little purer ($.61$ against $.55$) and, at a smaller area, slightly less precise ($.92$ against $.95$); with $25$ and $50$ parts its NMI is clearly higher ($.17$), but the number of labels raises both NMI and its null, the parts fill $61$ and $66\%$ of the crop, and purity and precision fall. \Cref{fig:pdisco} puts the two side by side on the same images. With $8$ parts PDiscoFormer separates the driver cabin, the students cabin and the wheels (its largest part, $13\%$ of the crop, lies $55\%$ in the students cabin and $27\%$ in the driver cabin, another part $42\%$ in the wheels), while our groups cut the bus into the roof sign, the rows of windows, the lower body and the wheels; with $25$ (and $50$) parts PDiscoFormer also covers the houses behind the bus, because its parts are shared by all $109$ classes and do not depend on the decision, whereas our groups leave them uncolored. This is a favorable setting for PDiscoFormer: school bus is one of its training classes, its images are named like ImageNet training images and may include the ones we evaluate on, and finding parts is its objective, while our groups come from the decision of a classifier. Where the two differ is the sign: for a part of PDiscoFormer we can read how much it supports school bus above the average class (a part logit), and the share of such (image, part) pairs with positive support is $98\%$, $86\%$ and $69\%$ for $K=8$, $25$ and $50$; the parts that cover pixels of these images support the class, so restricting the table to the parts with positive support changes neither area nor purity. There is no counterpart of our negative groups, which stand for evidence against the class in other vehicles, and the sign is a property of a trained multi-class head, not of a monotone readout checked by interventions. We do not run the intervention tests of \cref{sec:sign} on it, because they would test its own classifier and not the frozen ViT-B/16 whose decision we explain; a comparison of the decisions would need PDiscoFormer to be trained on the labels of that teacher, which we did not do, to keep it as released.

\section{What the discard channels learn}
\label{app:discard}
\begin{figure*}[tp]
\centering
\includegraphics[width=.62\textwidth]{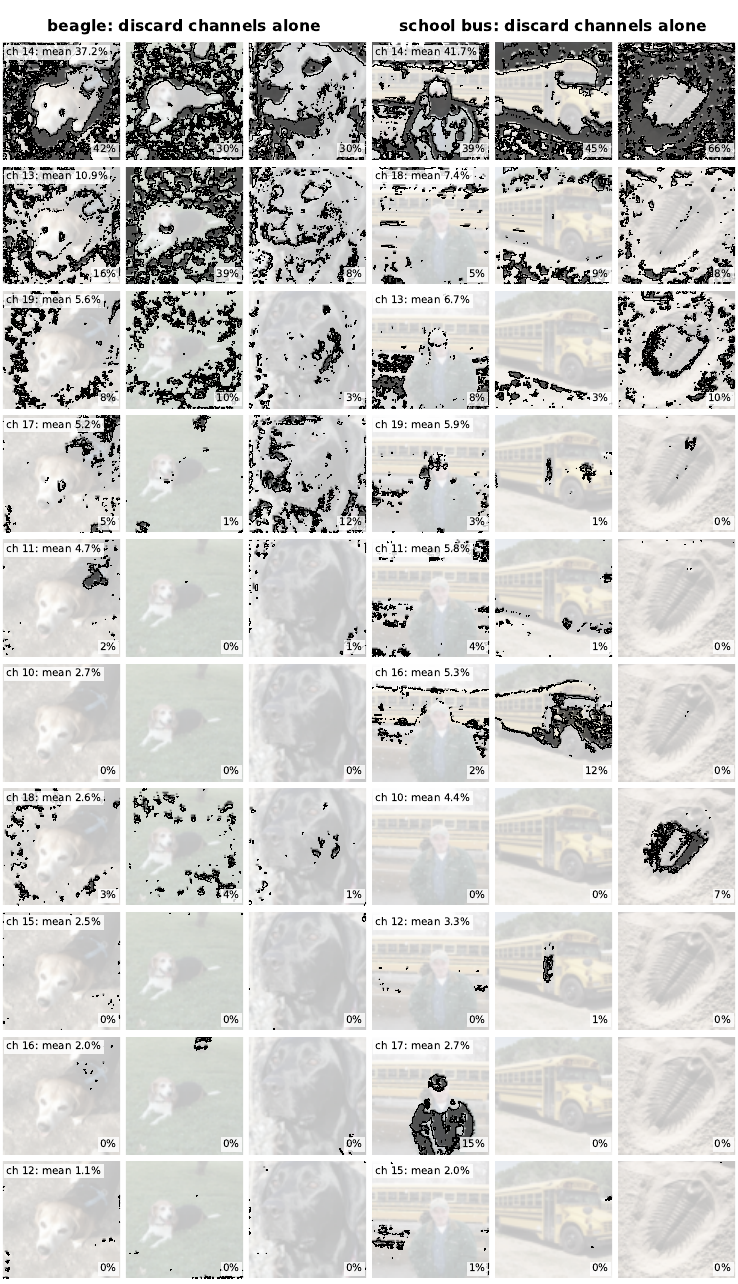}
\caption{Each discard channel alone (dark) on the images of \cref{fig:gallery-56} (dense maps of the $112\times112$ models); rows are ordered by the mean area over the balanced test set (label of the first tile), the number in a tile is the share of that image.}
\label{fig:discard}
\end{figure*}

The ten discard channels have no readout, and TV acts on no boundary between two of them, so nothing asks them to be compact or meaningful. The figures leave them uncolored; \cref{fig:discard} shows each discard channel of the $112\times112$ models of seed $101$ (\cref{app:finer}) alone on the images of \cref{fig:gallery-56}, ordered by its mean area. One discard channel takes what no used group claims (channel $14$: $37\%$ of the image on beagle and $42\%$ on school bus, as a mean over the balanced test set of the $112\times112$ models of seed $101$; ground, grass, sand, the sky and the background in general), and it is the largest on every image shown except the second beagle image, where channel $13$ takes the grass ($39\%$ against $30\%$). Of the others, five (beagle) and three (school bus) take less than $4\%$ on average and are empty on most of the images shown, and the rest are small, speckled sets of pixels that follow texture more than objects: some of them take a context region in a single image (on beagle channel $13$ the grass of the second image, on school bus channel $17$ the jacket of the man in the first, channel $16$ the contour of the bus in the second and channel $10$ the ring around the fossil). In \cref{fig:discard} these channels are fragmented into many small pieces, which no term penalizes. The discard channels therefore act as a catch-all with one main background class and a few texture-like classes; they may catch parts of the image that are interesting, but they carry no evidence for the decision and we do not interpret them.

\section{Does the teacher confirm the sign of the groups?}
\label{app:sign}
A group with a positive coefficient should carry evidence for the class in the frozen ViT-B/16 and a group with a negative coefficient evidence against it. We tested this on the surrogates of the main text (beagle and school bus, with and without TV, three seeds, the best-fidelity checkpoint). For every test image and every used group of at least $1\%$ of the crop, the pixels of the group are replaced by the same pixels of another image (deletion) or put into another image (insertion), as in the replacement tests of \cref{sec:compare}, and we measure the change $\Delta$ of the class margin of the teacher, $z_{\rm class}-\max_{c\neq{\rm class}}z_c$ (the quantity the surrogate describes: the class is the top-1 class). With $s$ the sign of the coefficient, the expected change is $-s\Delta>0$ for deletion and $s\Delta>0$ for insertion. Every group has a control, the same shape moved to another place (torus shift of the least overlap), because any removal of image content changes the margin a little. \Cref{tab:sign} gives the mean signed change of the groups and of their controls and, in brackets, the share of groups that beat their control.

\emph{Positive groups} pass: on the images of the class, deleting a positive group lowers the margin about twice as much as deleting the same shape elsewhere ($0.99$ against $0.42$ logits on beagle with TV, $1.20$ against $0.54$ on school bus), and the group beats its control for $61$ to $66\%$ of the groups. The groups of one image are correlated, so the unit of the uncertainty is the image: a bootstrap over the test images, which resamples an image with all its groups and seeds ($2{,}000$ draws, \texttt{sign\_test\_bootstrap.py}), gives $95\%$ intervals of $61$--$70\%$ (beagle) and $58$--$68\%$ (school bus) with TV and $56$--$67\%$ and $59$--$68\%$ without, and a two-level bootstrap that also resamples the three seeds gives intervals at most $3$ points wider. Putting a positive group into another image helps less ($52$ to $63\%$; the interval includes $50\%$ for beagle without TV). \emph{Negative groups} are confirmed only in part. Deleting them from images of a sibling class (dog breeds for beagle, road vehicles for school bus) raises the margin four times as much as the control on beagle with TV ($0.50$ against $0.13$, $61\%$, interval $57$--$64\%$) and less clearly in the other settings ($51$ to $57\%$; the interval includes $50\%$ for school bus without TV, and for school bus with TV once the seeds are resampled), and with the mean color as replacement the effect shrinks to $50$ to $55\%$ (every interval includes $50\%$). Pasting a negative group into a mean-color canvas lowers the margin more than the control does ($56$ to $68\%$, intervals between $53$ and $71\%$), but pasting it into another image does not (the change has the wrong sign for the groups and for the controls alike, $46$ to $52\%$; for beagle with TV the groups even lose to their controls, $43$--$50\%$). With TV, negative groups occur on the images of the class only two to five times per seed, so they cannot be tested there. We conclude that the teacher confirms the direction of the positive groups. For the negative groups the interventions agree with the sign only in part: clearly in the insertion into a mean-color canvas ($56$ to $68\%$), weakly in the deletion on sibling images ($51$ to $61\%$), and not with the mean color as replacement ($50$ to $55\%$) or in the insertion into another image ($46$ to $52\%$). The negative groups of the figures therefore describe the surrogate, and only in part the teacher.

\begin{table}[tp]
\centering
\small
\caption{Sign test of the groups against the teacher: mean signed change of the class margin (logits) for the groups / for their shifted-shape controls, with the share of groups that beat their control in brackets ($50$: no better than the control). Positive groups, on images of the class: deletion, the pixels replaced by another negative image ($173$ to $243$ groups per seed), and insertion of the group into another negative image. Negative groups: deletion on negative images of sibling classes, replaced by another image ($147$ to $597$ groups per seed) or by the mean color; insertion of the groups found on random negative images ($331$ to $635$ per seed) into a mean-color canvas or into another negative image. The signed change is positive when the teacher moves in the direction the sign of the group predicts. Means over three seeds; the win rates of single seeds differ by up to $12$ points; $95\%$ intervals over images in the text.}
\label{tab:sign}
\fitcolumn{\begin{tabular}{@{}llcccccc@{}}
\toprule
 & & \multicolumn{2}{c}{positive groups} & \multicolumn{4}{c}{negative groups}\\
\cmidrule(lr){3-4}\cmidrule(lr){5-8}
task & arm & deletion & insertion & \shortstack{deletion,\\ other image} & \shortstack{deletion,\\ mean color} & \shortstack{insertion,\\ mean-color canvas} & \shortstack{insertion,\\ other image}\\
\midrule
beagle & TV & $0.99/0.42$ (66) & $0.55/0.42$ (57) & $0.50/0.13$ (61) & $0.27/0.09$ (53) & $1.33/0.84$ (61) & $-0.22/-0.16$ (46)\\
 & without TV & $1.15/0.64$ (61) & $0.65/0.56$ (52) & $0.47/0.18$ (56) & $0.24/0.11$ (50) & $1.58/1.19$ (59) & $-0.40/-0.36$ (48)\\
school bus & TV & $1.20/0.54$ (63) & $1.58/0.99$ (63) & $0.37/0.16$ (57) & $0.23/0.04$ (55) & $1.38/0.77$ (68) & $-0.08/-0.06$ (48)\\
 & without TV & $1.28/0.73$ (64) & $1.57/1.33$ (59) & $0.41/0.35$ (51) & $0.22/0.17$ (53) & $1.90/1.61$ (56) & $-0.24/-0.27$ (52)\\
\bottomrule
\end{tabular}}
\end{table}

\emph{Independent check by signed attribution.} The intervention test changes the image; a second test leaves it alone and asks an attribution method that shares nothing with the surrogate. For every test image and every used group of at least $1\%$ of the crop we sum the signed AttnLRP relevance of the class logit~\cite{achtibat2024attnlrp} (conserved: summed over all pixels it gives the logit up to the bias terms) over the pixels of the group and over its shifted-shape control, and compare both with the sign $s$ of the coefficient: for a correct sign $sR>0$, and $sR$ of the group exceeds that of its control. Signed integrated gradients (zero baseline, $32$ steps) serve as a second, noisier method, and LeGrad~\cite{bousselham2025legrad}, which keeps only positive evidence, as a reference for the positive groups. \Cref{tab:signattr} gives the result for the TV models (three seeds). Positive groups pass clearly: on the images of the class the relevance is positive for $93\%$ (beagle) and $99\%$ (school bus) of the groups against $65\%$ and $72\%$ of their controls, and the group beats its control for $84\%$ and $85\%$ of the groups (integrated gradients $60\%$ and $69\%$, LeGrad $87\%$ and $89\%$). On negative images of sibling classes the relevance per pixel of a positive group exceeds that of a negative group of the same image in $76\%$ (beagle) and $70\%$ (school bus) of the pairs. Negative groups pass only there: on sibling-class images their relevance is negative for $85\%$ and $81\%$ of the groups and beats the control for $70\%$ and $65\%$ (against $61\%$ and $57\%$ in the deletion test), whereas on random negative images, where the teacher is far from the class, it shows nothing ($53\%$ and $49.5\%$), and integrated gradients does not separate negative groups from their controls on any of the negative images ($47$ to $54\%$). The routes agree for the positive groups. For the negative groups they differ: the attribution test confirms them only on sibling-class images ($65$ to $70\%$), the deletion test only weakly there ($51$ to $61\%$), and the insertion test into a mean-color canvas shows an effect on random negative images ($56$ to $68\%$) where the attribution shows none ($49.5$ to $53\%$). We therefore claim the direction of the positive groups only.

\begin{table}[tp]
\centering
\small
\caption{Signed attribution against the sign of the groups (TV models, three seeds). $sR$: the relevance of the class logit in the group times the sign of its coefficient, in logits, as a mean over groups for the groups and for their shifted-shape controls (positive: the direction the sign predicts), with the share of groups that beat their control in brackets; $sR>0$: share of groups / of controls with the predicted sign; IG win: the same win rate for signed integrated gradients. Positive and negative groups; images of the class, negative images of sibling classes (dog breeds, road vehicles) and random negative images.}
\label{tab:signattr}
\fitcolumn{\begin{tabular}{@{}llrccc@{}}
\toprule
task & groups (images) & per seed & AttnLRP $sR$: group / control (win \%) & $sR>0$ (\%) & IG win (\%)\\
\midrule
beagle & positive (class) & $173$ & $0.163/0.034$ (84) & $93/65$ & $60$\\
 & positive (sibling) & $135$ & $0.006/{-}0.007$ (60) & $52/31$ & $62$\\
 & negative (sibling) & $385$ & $0.039/0.009$ (70) & $85/72$ & $52$\\
 & negative (random) & $460$ & $0.005/0.002$ (53) & $59/61$ & $47$\\
school bus & positive (class) & $203$ & $0.333/0.075$ (85) & $99/72$ & $69$\\
 & positive (sibling) & $129$ & $0.015/{-}0.001$ (61) & $54/35$ & $61$\\
 & negative (sibling) & $147$ & $0.024/0.010$ (65) & $81/76$ & $54$\\
 & negative (random) & $331$ & $0.001/0.000$ (49.5) & $49/55$ & $48$\\
\bottomrule
\end{tabular}}
\end{table}

\emph{Does the surrogate respond to an edit as the teacher does?} The two tests above ask the teacher. A surrogate that explains the teacher should also move as the teacher moves when the image changes, which a fit to the teacher's decisions does not guarantee. For every deletion and insertion of the intervention test (TV models, three seeds, two negative images per edit, each group with its control) the edited image therefore also goes through the surrogate, recomputed from the edited pixels: its sixteen shifted views pass the frozen trunk, the prototypes and the readout, as in the store (\texttt{intervention\_fidelity.py}, \texttt{teacher\_agreement.py}). \Cref{tab:agree} compares the change of the surrogate's logit with the change of the teacher's class margin, with $95\%$ intervals from a bootstrap over the edited test images. For the positive groups on images of the class the two move in the same direction in $72\%$ (beagle) and $77\%$ (school bus) of the deletions, their changes have a rank correlation of $0.35$ and $0.41$ ($0.34$ and $0.41$ within an image, where it ranks the groups of one image), and for $64\%$ and $63\%$ of the groups they agree on whether the group moves the score more than its control. The surrogate, however, rarely follows a change of the teacher's decision: deleting one positive group flips the teacher in $84$ of $1{,}040$ (beagle) and $79$ of $1{,}216$ edits (school bus), and the surrogate flips in $2$ to $4\%$ of these; its decision rests on several groups at once and survives the loss of one, where the teacher's does not. On the teacher's own false positives the picture is similar ($63$ and $68\%$ in the same direction; the surrogate follows $10$ and $30\%$ of the teacher's flips). For the negative groups on sibling-class images the agreement is weak: $60\%$ and $62\%$ of the deletions move the two in the same direction (rank correlation $0.16$ and $0.23$), and on the third of these images closest to the teacher's decision boundary $56\%$ and $50\%$, with intervals that reach $50\%$; insertions of negative groups into another image agree at chance ($45$ to $57\%$). In every row the surrogate moves in the direction of the sign far more often ($69$ to $98\%$ of the deletions) than the teacher does ($45$ to $77\%$). What the surrogate says a positive group does, the teacher thus does in direction, and in the order of the groups of an image, more often than not, but not in its decisions; what it says a negative group does is not established for the teacher.

\emph{Training on the teacher's confidence.} The surrogate is fitted to the teacher's decision only. A natural remedy for its weak response to edits is to fit the teacher's confidence instead: the binary cross-entropy against the soft target $\sigma(m/T)$, with $m$ the teacher's class margin and $T=1$, which is the Kullback--Leibler divergence between the teacher's and the surrogate's Bernoulli distributions up to a constant, and which keeps the decision boundary at the teacher's top-1 rule ($m>0$). With everything else unchanged (three seeds, beagle and school bus), the fidelity is $98.7\%$ and $98.9\%$, the used area $24\%$ and $16\%$, and the box precision $97.3\%$ and $94.9\%$. Under the edits of \cref{tab:agree} the surrogate now follows $12\%$ of the teacher's decision flips on both tasks (instead of $2.4\%$ and $3.8\%$), but the agreement in direction does not improve: $73\%$ and $72\%$ for the positive groups on images of the class (rank correlation $0.32$ and $0.33$), $58\%$ and $66\%$ for the negative groups on sibling-class images, and of the images of the five nearest classes (\cref{tab:siblings}) it rejects $32.5\%$ and $71.7\%$, as before. $T=2$ on beagle gives the same picture. Fitting the teacher's confidence thus makes the surrogate less sure of itself but does not by itself make it respond to edits as the teacher does.

\emph{Why negative groups are hard to confirm.} We see one reason in the asymmetry of the task. The positive class is one object, the negative class is the $999$ other classes, and the teacher's ``no'' means that another class wins: the class margin $z_{\rm class}-\max_{c\neq{\rm class}}z_c$ depends on the evidence for whichever class wins, and that class changes from image to image. A negative group of the surrogate can therefore only stand for evidence that many competing classes share, and its effect on the teacher's margin is small and depends on the competitor, whereas a positive group stands for the evidence of one class. This fits the pattern of \cref{tab:sign,tab:signattr,tab:agree}: negative groups are confirmed in part on sibling-class images, where the competitor is close and specific, and not on random negatives. A decision between a finding and its absence, as in the chest X-ray teacher of \cref{app:cxr} (one logit per finding), is closer to symmetric: evidence against the finding acts on the same logit as evidence for it. We therefore ran the same edits on that teacher (\cref{tab:agreecxr}; \texttt{cxr\_interventions.py}): the $28\times28$ TV surrogates of \cref{app:cxr}, three seeds, up to $150$ teacher-positive and $150$ teacher-negative CheXlocalize test images per finding, and every used group of at least $1\%$ of the crop, with its shifted-shape control, replaced by the same pixels of another teacher-negative image (two per group) or by the mean intensity of the image, or put into another teacher-negative image. The surrogate now moves with the teacher for both signs. For the negative groups on teacher-negative images the two change in the same direction in $81\%$ (effusion) and $79\%$ (edema) of the deletions, with rank correlations of $0.75$ and $0.76$, against $56$ to $62\%$ and $0.10$ to $0.23$ on ImageNet; for the positive groups on teacher-positive images in $85\%$ and $90\%$ ($0.87$ and $0.82$, against $72$ to $77\%$ and $0.35$ to $0.41$), and the surrogate follows $72\%$ and $59\%$ of the teacher's decision flips, against $2$ to $4\%$ on ImageNet. The insertions agree as well ($77$ to $86\%$). The confirmation of the sign by the teacher itself, however, is not much stronger than on ImageNet. Deleting a negative group raises the teacher's logit more than deleting its control for $61\%$ (effusion) and $56\%$ (edema) of the groups with another image as replacement, and for $61\%$ and $57\%$ with the mean intensity (ImageNet: $51$ to $61\%$ and $50$ to $55\%$), against $68$ to $72\%$ for the positive groups; putting a negative group into another teacher-negative image does not lower the logit more than its control does ($47\%$ and $55\%$). The symmetric decision thus removes most of the disagreement between the surrogate and the teacher under edits, which on ImageNet is largest for the negative groups, but not the weak confirmation of their sign. We read the latter as a property of the evidence against a finding: in a radiograph without effusion it is spread over the normal anatomy, and replacing one region by the same region of another normal radiograph, or by its mean intensity, removes little of it. The two settings also differ in the teacher, the domain and the lattice, so the result is consistent with the hypothesis rather than a test of it in isolation.

\begin{table}[tp]
\centering
\small
\caption{The edits of \cref{tab:sign,tab:agree} on the chest X-ray teacher, whose decision is binary (Medical\_MAE, one logit per finding; the $28\times28$ TV surrogates of \cref{app:cxr}, three seeds pooled; \texttt{cxr\_interventions.py}, \texttt{teacher\_agreement.py}). Images: teacher-positive CheXlocalize test images with the positive groups, teacher-negative ones with the negative groups (at most $150$ images each); near bound.: the third of the teacher-negative images with the highest logit. Edits: deletion of a group, its pixels replaced by the same pixels of another teacher-negative image, two per group. Same dir.\ and $\rho$ (rank correlation) as in \cref{tab:agree}; flips: deletions that change the teacher's decision (logit above zero), and the share of them that the surrogate follows. Teacher wins: share of the groups whose deletion moves the teacher's logit more than the deletion of the shifted-shape control does, in the direction of the sign, with another image or the mean intensity of the image as replacement (as in \cref{tab:sign}). Brackets: $95\%$ bootstrap intervals over the edited test images; the intervals of the teacher's win rates are at most $\pm3$ points for the positive and the negative rows and $\pm5$ for the rows near the boundary.}
\label{tab:agreecxr}
\setlength{\tabcolsep}{3pt}
\fitcolumn{\begin{tabular}{@{}llrcccc@{}}
\toprule
 & & & \multicolumn{3}{c}{surrogate vs.\ teacher} & teacher wins (\%)\\
\cmidrule(lr){4-6}\cmidrule(lr){7-7}
 & images & edits & same dir.\ (\%) & $\rho$ & flips (follows) & image / mean\\
\midrule
effusion & positive & $2{,}628$ & $84.6\ [82.9, 86.1]$ & $0.87$ & $577\ (72\%)$ & $68.2$ / $69.0$\\
 & negative & $5{,}870$ & $81.4\ [79.6, 83.1]$ & $0.75$ & $97\ (31\%)$ & $61.0$ / $60.7$\\
 & near bound. & $1{,}732$ & $78.8\ [75.9, 81.5]$ & $0.74$ & $94\ (29\%)$ & $57.9$ / $60.3$\\
\midrule
edema & positive & $2{,}242$ & $90.4\ [88.5, 92.2]$ & $0.82$ & $540\ (59\%)$ & $69.7$ / $71.5$\\
 & negative & $5{,}112$ & $79.1\ [77.5, 80.7]$ & $0.76$ & $95\ (25\%)$ & $56.0$ / $57.2$\\
 & near bound. & $1{,}588$ & $76.8\ [73.8, 79.6]$ & $0.75$ & $95\ (25\%)$ & $57.6$ / $59.7$\\
\bottomrule
\end{tabular}}
\end{table}

\begin{table*}[tp]
\centering
\small
\caption{The surrogate under the same edits as the teacher (TV models, three seeds pooled; \texttt{teacher\_agreement.py}). Edits: the deletions of a group from a test image (its pixels replaced by those of another negative image), two per group; insertions: the group pasted into two other negative images. Same direction: share of edits in which the surrogate's logit and the teacher's class margin change with the same sign; rank corr.: Spearman correlation of the two changes over all edits, and its mean within an image (images with at least three groups); vs.\ control: share of edits in which the two agree on whether the group changes the score more than its shifted-shape control does, in the direction of its sign. Near boundary: the third of the sibling images with the highest teacher margin. Teacher flips: deletions that change the teacher's top-1 decision, and the share of them in which the surrogate's decision changes too (on negative images the teacher's decision never changes). Brackets: $95\%$ bootstrap intervals over the edited test images ($1{,}000$ draws).}
\label{tab:agree}
\resizebox{\textwidth}{!}{\begin{tabular}{@{}lllrcccccc@{}}
\toprule
 & & & & \multicolumn{4}{c}{deletion} & insertion & \\
\cmidrule(lr){5-8}\cmidrule(lr){9-9}
task & images & groups & edits & same direction (\%) & rank corr. & within image & vs.\ control (\%) & same dir.\ (\%) & teacher flips (surrogate follows)\\
\midrule
beagle & class & positive & $1{,}040$ & $71.9\ [65.7, 77.8]$ & $0.35\ [0.26, 0.43]$ & $0.34$ & $63.7$ & $61.1$ & $84\ (2.4\%)$\\
 & teacher's false positives & positive & $476$ & $63.2\ [55.1, 72.0]$ & $0.32\ [0.20, 0.43]$ & $0.24$ & $63.4$ & $56.1$ & $146\ (9.6\%)$\\
 & sibling & positive & $652$ & $63.7\ [59.3, 68.4]$ & $0.10\ [0.02, 0.21]$ & $0.14$ & $58.3$ & $52.6$ & --\\
 & sibling & negative & $1{,}494$ & $60.1\ [55.0, 64.9]$ & $0.16\ [0.08, 0.25]$ & $0.24$ & $56.4$ & $52.5$ & --\\
 & sibling, near boundary & negative & $414$ & $55.8\ [48.1, 64.3]$ & $0.03\ [-0.15, 0.22]$ & $0.11$ & $58.5$ & $57.0$ & --\\
 & random & negative & $1{,}838$ & $56.1\ [52.2, 59.7]$ & $0.10\ [0.01, 0.19]$ & $0.17$ & $52.0$ & $54.2$ & --\\
\midrule
school bus & class & positive & $1{,}216$ & $77.2\ [72.7, 81.6]$ & $0.41\ [0.27, 0.53]$ & $0.41$ & $63.2$ & $72.8$ & $79\ (3.8\%)$\\
 & teacher's false positives & positive & $150$ & $68.0\ [60.4, 75.0]$ & $0.28\ [0.03, 0.56]$ & $0.25$ & $66.0$ & $73.3$ & $37\ (29.7\%)$\\
 & sibling & positive & $772$ & $53.9\ [49.0, 58.5]$ & $0.13\ [0.02, 0.25]$ & $0.06$ & $56.6$ & $58.8$ & --\\
 & sibling & negative & $880$ & $61.9\ [55.4, 68.0]$ & $0.23\ [0.11, 0.33]$ & $0.18$ & $55.3$ & $49.3$ & --\\
 & sibling, near boundary & negative & $248$ & $50.4\ [39.2, 62.2]$ & $0.15\ [-0.03, 0.31]$ & $0.04$ & $51.2$ & $45.6$ & --\\
 & random & negative & $1{,}374$ & $56.8\ [52.3, 61.3]$ & $0.11\ [0.03, 0.19]$ & $0.11$ & $54.1$ & $49.4$ & --\\
\bottomrule
\end{tabular}}
\end{table*}

\section{Localization against boxes: spread over seeds}
\label{app:loc}
\Cref{tab:loc} gives the precision, lift, recall and positive area of the two arms against the object boxes with the standard deviation over the three seeds (\cref{sec:loc}). A precision is computed per image and seed, $|P\cap B|/|P|$ on the pixels of the $224\times224$ crop, and averaged over images and seeds; every positive region is nonempty, so the order of the two means does not matter and the means are those of \cref{sec:loc,tab:compare}. Lattice site $u$ covers the crop pixels $[4u+2,4u+6)$ around its center $3.5+4u$ (the outermost sites extend to the border), in every box measure of the paper (\texttt{loc\_table.py}, \texttt{box\_from\_predictions.py}).

\begin{table}[tp]
\centering
\small
\caption{Positive groups against the ImageNet object boxes, teacher-positive test images that show the class ($43$ and $49$ images), mean $\pm$ standard deviation over three seeds, best-fidelity checkpoint. Precision: share of positive-group area inside the box; lift: precision divided by the mean box share (last column); recall: share of the box covered by positive groups.}
\label{tab:loc}
\fitcolumn{\begin{tabular}{@{}llccccc@{}}
\toprule
task & arm & prec.\ (\%) & lift & rec.\ (\%) & pos.\ area (\%) & box (\%)\\
\midrule
beagle & $\rho=0$ & $91.1\pm8.3$ & 1.27 & $44.3\pm4.8$ & 35.1 & 71.8\\
 & $\rho=0.05$ & $95.0\pm5.5$ & 1.32 & $32.5\pm7.3$ & 24.1 & 71.8\\
school bus & $\rho=0$ & $81.1\pm1.3$ & 1.35 & $57.0\pm4.3$ & 38.7 & 60.0\\
 & $\rho=0.05$ & $94.7\pm0.7$ & 1.58 & $45.7\pm3.0$ & 25.5 & 60.0\\
\bottomrule
\end{tabular}}
\end{table}

\section{Comparison with attribution methods: details}
\label{app:compare}
\begin{table}[tp]
\centering
\small
\caption{Our groups against re-run attribution methods on the same teacher, at equal area, ours is the mean of three seeds. Prec., rec., lift: against the object boxes ($43$ and $49$ images). AUC: threshold-free area under the ROC curve of the continuous map against the boxes ($39$ and $45$ images whose box does not fill the crop); for our binary regions it is the mean of recall and specificity. Del.\ and ins.: net flip in points, the share of cases in which the teacher's top-1 decision changes when the region is replaced by a negative image (deletion) or put into one (insertion), minus the same for the same shape elsewhere (control). Centered disk: a disk in the middle of the crop with the area of our groups. The other maps (Grad-CAM, rollout, integrated gradients, RISE, DAVE, ViT-CX), the $k$-means groups, intervals and variants: \cref{tab:compare-full,tab:compare-ci,app:compare}.}
\label{tab:compare}
\fitcolumn{\begin{tabular}{@{}llcccccc@{}}
\toprule
task & method & prec.\ (\%) & rec.\ (\%) & lift & AUC (\%) & del. & ins.\\
\midrule
beagle & AtomEx (TV) & 95.0 & 32.5 & 1.32 & 64.1 & +70 & +12\\
 & Chefer & 92.7 & 32.4 & 1.29 & 75.5 & +71 & +35\\
 & LeGrad & 98.3 & 34.5 & 1.37 & 84.7 & +72 & +29\\
 & AttnLRP & 90.1 & 31.2 & 1.26 & 63.9 & +88 & +17\\
 & centered disk & 95.7 & 33.2 & 1.33 & 65.5 & +24 & +33\\
\midrule
school bus & AtomEx (TV) & 94.7 & 45.7 & 1.58 & 72.5 & +70 & +45\\
 & Chefer & 82.5 & 39.9 & 1.38 & 78.3 & +43 & +41\\
 & LeGrad & 96.5 & 47.3 & 1.61 & 88.3 & +74 & +48\\
 & AttnLRP & 87.2 & 41.5 & 1.45 & 70.3 & +83 & +59\\
 & centered disk & 81.6 & 36.2 & 1.36 & 65.2 & +6 & +35\\
\bottomrule
\end{tabular}}
\end{table}

\begin{table*}[tp]
\centering
\small
\caption{All methods and tasks of \cref{sec:compare}: precision, recall, IoU (intersection over union of the region and the boxes) and lift against the object boxes (beagle $43$ and school bus $49$ images; valley $39$ images without boxes), and the deletion and insertion tests (flip rate of the teacher's decision, flip rate of the control, and the specificity in percentage points with the $95\%$ bootstrap interval over images). Maps are binarized to the area of our TV groups on each image; ``without TV'' has a larger area and is not at equal area. AUC box and AUC mask: threshold-free area under the ROC curve of the continuous map against the boxes ($39$ and $45$ images) and the ImageNet-S masks ($11$ and $15$ images); for our binary regions it is the mean of recall and specificity. LeGrad and ViT-CX were run on beagle and school bus only.}
\label{tab:compare-full}
\resizebox{\textwidth}{!}{\begin{tabular}{@{}llcccccccccccc@{}}
\toprule
task & method & prec.\ (\%) & rec.\ (\%) & IoU (\%) & lift & AUC box (\%) & AUC mask (\%) & del.\ flip (\%) & ctrl.\ (\%) & del.\ spec.\ & ins.\ flip (\%) & ctrl.\ (\%) & ins.\ spec.\\
\midrule
beagle & AtomEx, TV 0.05 & 95.0 & 32.5 & 31.5 & 1.32 & 64.1 & 71.7 & 83 & 14 & +70 [+61, +78] & 12 & 0 & +12 [+7, +17]\\
 & AtomEx, without TV & 91.1 & 44.3 & 41.6 & 1.27 & 65.7 & 74.6 & 98 & 36 & +62 [+52, +72] & 36 & 2 & +33 [+24, +42]\\
 & Grad-CAM & 74.7 & 26.0 & 24.2 & 1.04 & 52.5 & 33.9 & 40 & 26 & +14 [-1, +29] & 13 & 5 & +8 [-2, +20]\\
 & Grad-CAM, lattice & 74.8 & 25.7 & 23.9 & 1.04 & 48.4 & 35.4 & 50 & 37 & +13 [-5, +31] & 3 & 0 & +3 [+0, +9]\\
 & rollout & 82.5 & 28.4 & 26.9 & 1.15 & 65.5 & 74.0 & 33 & 17 & +15 [+2, +28] & 12 & 1 & +10 [+3, +19]\\
 & rollout, lattice & 81.0 & 27.9 & 26.2 & 1.13 & 63.1 & 73.8 & 42 & 17 & +24 [+10, +40] & 2 & 1 & +1 [-2, +5]\\
 & Chefer & 92.7 & 32.4 & 31.4 & 1.29 & 75.5 & 83.5 & 83 & 12 & +71 [+58, +81] & 35 & 0 & +35 [+23, +47]\\
 & Chefer, lattice & 89.8 & 31.2 & 29.8 & 1.25 & 71.6 & 81.9 & 81 & 17 & +64 [+50, +77] & 15 & 0 & +15 [+7, +24]\\
 & Integrated gradients & 77.5 & 26.1 & 24.1 & 1.08 & 52.3 & 53.3 & 88 & 38 & +50 [+37, +63] & 0 & 0 & +0 [+0, +0]\\
 & RISE & 90.6 & 31.4 & 30.1 & 1.26 & 63.1 & 65.7 & 50 & 8 & +42 [+28, +55] & 36 & 5 & +31 [+21, +42]\\
 & DAVE & 94.8 & 33.1 & 32.3 & 1.32 & 80.1 & 80.7 & 86 & 15 & +71 [+55, +85] & 36 & 0 & +36 [+23, +48]\\
 & LeGrad & 98.3 & 34.5 & 34.2 & 1.37 & 84.7 & 93.3 & 83 & 10 & +72 [+59, +84] & 30 & 1 & +29 [+17, +41]\\
 & ViT-CX & 95.1 & 33.3 & 32.9 & 1.32 & 78.3 & 83.2 & 66 & 15 & +51 [+37, +65] & 23 & 0 & +23 [+14, +34]\\
 & AttnLRP & 90.1 & 31.2 & 29.8 & 1.26 & 63.9 & 66.4 & 99 & 10 & +88 [+80, +95] & 17 & 0 & +17 [+8, +27]\\
 & k-means groups & 82.7 & 27.7 & 26.1 & 1.15 & 58.8 & 65.8 & 57 & 21 & +36 [+20, +53] & 2 & 0 & +2 [+0, +6]\\
 & centered disk & 95.7 & 33.2 & 32.5 & 1.33 & 65.5 & 72.7 & 35 & 10 & +24 [+13, +36] & 33 & 0 & +33 [+21, +44]\\
 & random disk & 90.3 & 31.5 & 30.5 & 1.26 & 61.5 & 63.3 & 19 & 7 & +12 [+1, +23] & 27 & 0 & +27 [+17, +37]\\
\midrule
school bus & AtomEx, TV 0.05 & 94.7 & 45.7 & 43.6 & 1.58 & 72.5 & 76.1 & 77 & 7 & +70 [+60, +81] & 47 & 3 & +45 [+34, +56]\\
 & AtomEx, without TV & 81.1 & 57.0 & 47.4 & 1.35 & 70.6 & 74.5 & 89 & 18 & +72 [+62, +81] & 65 & 14 & +51 [+41, +62]\\
 & Grad-CAM & 64.1 & 28.9 & 25.9 & 1.07 & 50.0 & 51.1 & 21 & 12 & +9 [-4, +22] & 37 & 10 & +27 [+14, +39]\\
 & Grad-CAM, lattice & 58.6 & 25.8 & 22.0 & 0.98 & 48.9 & 48.9 & 20 & 15 & +5 [-7, +16] & 22 & 9 & +13 [+4, +23]\\
 & rollout & 69.8 & 33.3 & 29.1 & 1.16 & 65.4 & 53.5 & 9 & 6 & +3 [-5, +12] & 7 & 16 & -9 [-19, +0]\\
 & rollout, lattice & 69.3 & 32.7 & 28.3 & 1.16 & 64.6 & 54.3 & 12 & 10 & +2 [-8, +12] & 10 & 16 & -6 [-14, +2]\\
 & Chefer & 82.5 & 39.9 & 36.3 & 1.38 & 78.3 & 76.7 & 44 & 1 & +43 [+31, +54] & 49 & 8 & +41 [+29, +53]\\
 & Chefer, lattice & 80.2 & 38.3 & 34.2 & 1.34 & 76.4 & 74.6 & 57 & 13 & +44 [+31, +58] & 39 & 2 & +37 [+24, +49]\\
 & Integrated gradients & 64.3 & 28.5 & 24.1 & 1.07 & 52.9 & 51.3 & 58 & 21 & +37 [+24, +49] & 17 & 0 & +17 [+9, +28]\\
 & RISE & 74.8 & 34.9 & 30.6 & 1.25 & 59.1 & 59.0 & 26 & 3 & +22 [+11, +35] & 26 & 14 & +11 [+3, +19]\\
 & DAVE & 73.6 & 33.1 & 29.2 & 1.23 & 69.1 & 70.5 & 41 & 7 & +34 [+22, +45] & 38 & 2 & +36 [+24, +47]\\
 & LeGrad & 96.5 & 47.3 & 45.7 & 1.61 & 88.3 & 93.1 & 78 & 3 & +74 [+62, +85] & 48 & 0 & +48 [+36, +59]\\
 & ViT-CX & 86.2 & 42.0 & 39.4 & 1.44 & 77.3 & 82.5 & 50 & 2 & +48 [+35, +60] & 31 & 9 & +21 [+9, +35]\\
 & AttnLRP & 87.2 & 41.5 & 38.0 & 1.45 & 70.3 & 70.1 & 93 & 10 & +83 [+72, +92] & 59 & 0 & +59 [+47, +71]\\
 & k-means groups & 66.7 & 29.5 & 26.0 & 1.11 & 47.8 & 47.4 & 31 & 9 & +21 [+9, +34] & 19 & 9 & +10 [+0, +21]\\
 & centered disk & 81.6 & 36.2 & 33.9 & 1.36 & 65.2 & 69.0 & 9 & 3 & +6 [-2, +14] & 44 & 9 & +35 [+20, +48]\\
 & random disk & 73.6 & 32.4 & 29.7 & 1.23 & 61.5 & 64.7 & 11 & 1 & +10 [+4, +18] & 40 & 16 & +23 [+10, +37]\\
\midrule
valley & AtomEx, TV 0.05 & -- & -- & -- & -- & -- & -- & 73 & 57 & +16 [+10, +23] & 6 & 1 & +5 [+2, +8]\\
 & AtomEx, without TV & -- & -- & -- & -- & -- & -- & 100 & 98 & +2 [+0, +4] & 31 & 12 & +19 [+12, +27]\\
 & Grad-CAM & -- & -- & -- & -- & -- & -- & 81 & 67 & +14 [+4, +24] & 1 & 0 & +1 [+0, +4]\\
 & Grad-CAM, lattice & -- & -- & -- & -- & -- & -- & 83 & 76 & +8 [-3, +18] & 3 & 1 & +1 [+0, +4]\\
 & rollout & -- & -- & -- & -- & -- & -- & 73 & 64 & +9 [-5, +23] & 0 & 3 & -3 [-6, +0]\\
 & rollout, lattice & -- & -- & -- & -- & -- & -- & 64 & 69 & -5 [-17, +6] & 0 & 5 & -5 [-12, +0]\\
 & Chefer & -- & -- & -- & -- & -- & -- & 91 & 68 & +23 [+12, +35] & 4 & 1 & +3 [+0, +6]\\
 & Chefer, lattice & -- & -- & -- & -- & -- & -- & 92 & 73 & +19 [+6, +33] & 4 & 1 & +3 [+0, +6]\\
 & Integrated gradients & -- & -- & -- & -- & -- & -- & 85 & 53 & +32 [+21, +45] & 0 & 0 & +0 [+0, +0]\\
 & RISE & -- & -- & -- & -- & -- & -- & 72 & 69 & +3 [-9, +14] & 1 & 0 & +1 [+0, +4]\\
 & DAVE & -- & -- & -- & -- & -- & -- & 63 & 45 & +18 [+6, +29] & 3 & 0 & +3 [+0, +6]\\
 & AttnLRP & -- & -- & -- & -- & -- & -- & 86 & 41 & +45 [+31, +59] & 15 & 0 & +15 [+6, +27]\\
 & k-means groups & -- & -- & -- & -- & -- & -- & 92 & 58 & +35 [+22, +50] & 4 & 1 & +3 [+0, +6]\\
 & centered disk & -- & -- & -- & -- & -- & -- & 79 & 64 & +15 [+5, +27] & 3 & 1 & +1 [+0, +4]\\
 & random disk & -- & -- & -- & -- & -- & -- & 63 & 71 & -8 [-19, +5] & 5 & 0 & +5 [+0, +12]\\
\bottomrule
\end{tabular}}
\end{table*}

\Cref{tab:compare-full} gives all methods and tasks of \cref{sec:compare} with the IoU against the boxes and the specificity of the two replacement tests, the effect of the region minus the effect of its control, paired per image, with $95\%$ bootstrap intervals over images ($2{,}000$ resamples; for our model the three seeds are averaged per image first). Settings: Grad-CAM on the input of the last block, attention rollout with the residual and the class-token row, Chefer's relevance over all blocks, integrated gradients with a zero (mean-color) baseline and $32$ steps, RISE with $4{,}000$ masks of $7\times7$ cells and $p=0.5$, DAVE on the class logit with $50$ steps, LeGrad on the last two blocks as in its code, ViT-CX on the $768$ channel maps of the tokens at the input of the last block, clustered by complete linkage (distance threshold $0.1$), AttnLRP as pixel relevance with the rules of the paper. DAVE has no lattice variant. The rows with ``lattice'' run the token-level method on the sixteen shifted views and interleave the maps into our $56\times56$ lattice; they are not better than the plain versions, so the finer lattice is not an advantage of our model only. The model without TV has a larger area ($35\%$ on beagle, $39\%$ on school bus and $63\%$ on valley), so its rows are not at equal area and are shown for reference. Valley has no boxes. Random disk: a disk with the area of our groups on the same image and a random center such that the disk lies inside the crop. The replacement tests use two negative images per test image, drawn from the negatives of the study, and the transplant is made on the $224\times224$ crop that the teacher sees.

\begin{table*}[tp]
\centering
\small
\caption{$95\%$ bootstrap intervals over images of the box precision, the recall, and the AUC against the boxes ($39$ and $45$ images whose box does not fill the crop) and the ImageNet-S masks ($11$ and $15$ images), at the area of our groups; our model is the mean of three seeds per image, its AUC is that of a binary region.}
\label{tab:compare-ci}
\begin{tabular}{@{}llcccc@{}}
\toprule
task & method & prec.\ (\%) & rec.\ (\%) & AUC box (\%) & AUC mask (\%)\\
\midrule
beagle & AtomEx, TV 0.05 & 95.0 [92.8, 96.8] & 32.5 [30.2, 34.8] & 64.1 [63.1, 65.0] & 71.7 [68.5, 75.0]\\
 & AtomEx, without TV & 91.1 [87.9, 93.8] & 44.3 [42.1, 46.5] & 65.7 [64.2, 67.1] & 74.6 [70.9, 78.0]\\
 & Grad-CAM & 74.7 [67.2, 81.4] & 26.0 [23.0, 29.2] & 52.5 [45.1, 59.7] & 33.9 [17.9, 53.1]\\
 & Grad-CAM, lattice & 74.8 [68.1, 80.8] & 25.7 [22.9, 28.6] & 48.4 [43.4, 53.3] & 35.4 [23.5, 47.9]\\
 & rollout & 82.5 [76.8, 87.3] & 28.4 [25.8, 31.2] & 65.5 [59.9, 70.5] & 74.0 [60.3, 86.3]\\
 & rollout, lattice & 81.0 [75.4, 86.1] & 27.9 [25.3, 30.4] & 63.1 [58.0, 68.1] & 73.8 [60.5, 85.2]\\
 & Chefer & 92.7 [90.6, 94.6] & 32.4 [29.7, 35.2] & 75.5 [72.1, 78.6] & 83.5 [75.3, 90.7]\\
 & Chefer, lattice & 89.8 [87.4, 92.0] & 31.2 [28.9, 33.7] & 71.6 [68.8, 74.3] & 81.9 [74.5, 87.9]\\
 & Integrated gradients & 77.5 [72.6, 82.1] & 26.1 [24.4, 27.9] & 52.3 [51.6, 53.1] & 53.3 [51.1, 55.7]\\
 & RISE & 90.6 [87.1, 93.5] & 31.4 [28.7, 34.0] & 63.1 [58.7, 67.7] & 65.7 [58.3, 73.4]\\
 & DAVE & 94.8 [92.7, 96.6] & 33.1 [30.4, 36.1] & 80.1 [77.9, 82.1] & 80.7 [76.7, 84.7]\\
 & LeGrad & 98.3 [96.8, 99.3] & 34.5 [31.6, 37.8] & 84.7 [82.2, 87.2] & 93.3 [89.9, 96.3]\\
 & ViT-CX & 95.1 [91.3, 98.2] & 33.3 [30.3, 37.1] & 78.3 [73.7, 82.6] & 83.2 [72.2, 92.4]\\
 & AttnLRP & 90.1 [87.6, 92.4] & 31.2 [28.9, 33.7] & 63.9 [62.2, 65.4] & 66.4 [63.9, 69.1]\\
 & k-means groups & 82.7 [76.4, 88.6] & 27.7 [25.4, 30.0] & 58.8 [53.8, 63.5] & 65.8 [53.3, 76.9]\\
 & centered disk & 95.7 [92.6, 98.4] & 33.2 [30.6, 35.9] & 65.5 [64.1, 67.0] & 72.7 [68.4, 77.7]\\
 & random disk & 90.3 [85.3, 94.5] & 31.5 [28.5, 34.7] & 61.5 [60.2, 62.7] & 63.3 [60.8, 65.8]\\
\midrule
school bus & AtomEx, TV 0.05 & 94.7 [91.0, 97.4] & 45.7 [41.7, 49.8] & 72.5 [70.4, 74.6] & 76.1 [72.1, 80.2]\\
 & AtomEx, without TV & 81.1 [75.7, 86.2] & 57.0 [53.2, 61.0] & 70.6 [68.6, 72.6] & 74.5 [70.6, 77.7]\\
 & Grad-CAM & 64.1 [55.0, 72.8] & 28.9 [24.4, 33.2] & 50.0 [43.1, 56.7] & 51.1 [38.7, 63.2]\\
 & Grad-CAM, lattice & 58.6 [51.3, 66.3] & 25.8 [22.7, 28.8] & 48.9 [44.7, 53.1] & 48.9 [40.2, 57.5]\\
 & rollout & 69.8 [62.7, 76.7] & 33.3 [29.2, 38.4] & 65.4 [59.3, 71.2] & 53.5 [41.5, 65.5]\\
 & rollout, lattice & 69.3 [62.6, 76.1] & 32.7 [28.8, 36.5] & 64.6 [59.1, 69.8] & 54.3 [42.7, 64.7]\\
 & Chefer & 82.5 [76.7, 87.3] & 39.9 [35.6, 44.6] & 78.3 [74.3, 81.6] & 76.7 [66.4, 85.2]\\
 & Chefer, lattice & 80.2 [74.5, 85.1] & 38.3 [34.5, 42.3] & 76.4 [73.0, 79.5] & 74.6 [65.2, 82.6]\\
 & Integrated gradients & 64.3 [57.8, 70.8] & 28.5 [26.7, 30.5] & 52.9 [51.7, 53.9] & 51.3 [48.9, 53.5]\\
 & RISE & 74.8 [68.5, 81.0] & 34.9 [31.2, 39.3] & 59.1 [53.7, 64.8] & 59.0 [50.7, 67.9]\\
 & DAVE & 73.6 [66.4, 80.1] & 33.1 [30.2, 36.2] & 69.1 [65.0, 73.0] & 70.5 [65.0, 76.3]\\
 & LeGrad & 96.5 [93.1, 98.9] & 47.3 [42.5, 52.4] & 88.3 [85.6, 90.6] & 93.1 [90.4, 95.3]\\
 & ViT-CX & 86.2 [80.0, 92.0] & 42.0 [37.0, 47.3] & 77.3 [71.4, 83.0] & 82.5 [70.6, 92.7]\\
 & AttnLRP & 87.2 [83.2, 90.5] & 41.5 [38.1, 45.1] & 70.3 [68.3, 72.3] & 70.1 [65.1, 74.8]\\
 & k-means groups & 66.7 [58.4, 74.7] & 29.5 [25.8, 33.1] & 47.8 [43.2, 52.6] & 47.4 [37.8, 56.7]\\
 & centered disk & 81.6 [73.4, 89.0] & 36.2 [32.3, 39.7] & 65.2 [62.5, 67.7] & 69.0 [64.6, 72.5]\\
 & random disk & 73.6 [64.7, 82.1] & 32.4 [28.4, 36.2] & 61.5 [59.4, 63.4] & 64.7 [61.3, 67.3]\\
\bottomrule
\end{tabular}
\end{table*}

\begin{figure*}[tp]
\centering
\includegraphics[width=\textwidth]{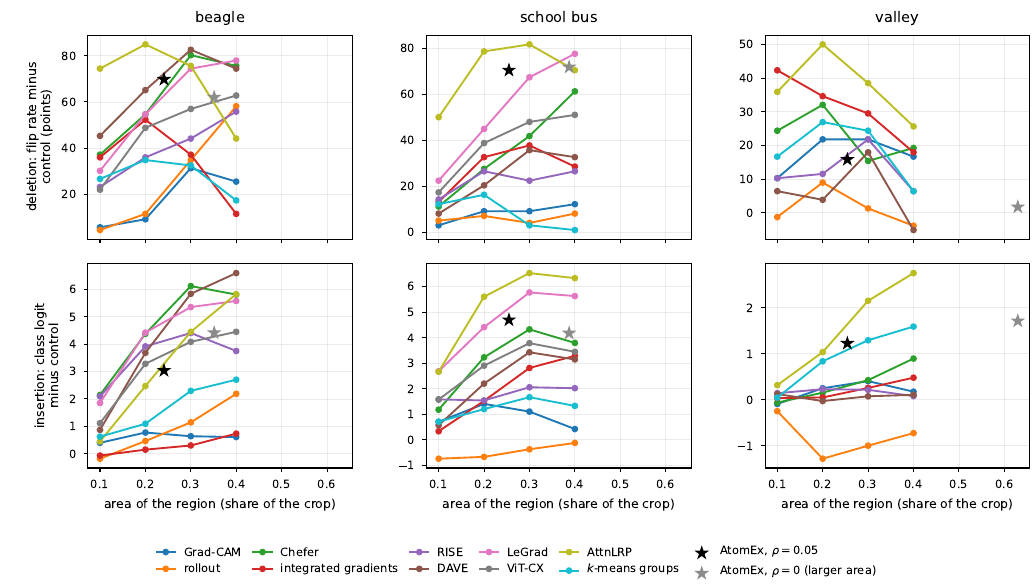}
\caption{Deletion (top) and insertion (bottom) against the area of the region: effect of the region minus the effect of its control, for every map binarized to $10$ to $40\%$ of the crop and for our models at their mean area (stars; the model without TV has a larger area). Deletion: flip rate of the teacher's decision, in points; insertion: change of the class logit. LeGrad and ViT-CX were run on beagle and school bus only.}
\label{fig:compare-sweep}
\end{figure*}

\Cref{fig:compare-sweep} binarizes every map to the same share of the crop on all images, $10$ to $40\%$ (in \cref{tab:compare-full} the share is that of our groups on each image), and plots the specificity of the two tests against that share; the stars are our models at their mean area. In deletion our TV model lies among the curves of Chefer's relevance, DAVE and LeGrad on beagle and below AttnLRP, above every curve except AttnLRP on school bus, and in the lower half on valley (below integrated gradients, AttnLRP, the $k$-means groups, Chefer's relevance and Grad-CAM). In insertion it lies below six of the ten curves on beagle, above all but LeGrad and AttnLRP on school bus, and below AttnLRP only on valley. Maps of Chefer's relevance and LeGrad next to our groups are shown in \cref{fig:teaser}.

An explanation should also not change when the input changes in a way that leaves the decision alone; heat maps are known to fail this~\cite{ghorbani2019fragile,dombrowski2019manipulated}. We perturbed the same teacher-positive images ($43$ beagle and $49$ school-bus images) in natural, non-adversarial ways: Gaussian pixel noise ($\sigma=0.02$ and $0.05$ of the intensity range), brightness $\times0.9$ and $\times1.1$, JPEG compression (quality $75$) and a translation of the content by $2$ and $4$ pixels (the crop window moved inside the resized image). Keeping only the cases in which the teacher's top-1 decision stays, we compare the region of every explanation before and after by its IoU, in the coordinates of the content: for AtomEx the positive groups of the three TV surrogates, recomputed on the perturbed image; for the maps the top pixels at the area of our groups on the unperturbed image (\texttt{explanation\_stability.py}). \Cref{tab:stability} gives the result. Our regions are more stable than those of Chefer's relevance, Grad-CAM and integrated gradients under every perturbation, and about as stable as LeGrad's, $0.01$ to $0.08$ lower under noise, brightness, JPEG and the $2$-pixel shift, but much more stable under the $4$-pixel shift ($0.91$ against $0.70$--$0.73$), which moves the content by exactly one site of the lattice. Rollout, which is not specific to the class, is among the most stable under noise and JPEG. The surrogate keeps its decision in all but one case, and $93$ to $99.6\%$ of the pixels that are positive before and after keep their group, so the identity of the components is stable too, which no map can show.

\begin{table}[tp]
\centering
\small
\caption{Stability under natural perturbations that keep the teacher's decision: IoU of the region of each explanation before and after the perturbation, in the coordinates of the content, mean over the images (kept: images whose teacher decision stays, of $43$ and $49$). AtomEx: positive groups, mean over three seeds; maps: top pixels at the area of our groups on the unperturbed image.}
\label{tab:stability}
\fitcolumn{\begin{tabular}{@{}llccccccc@{}}
\toprule
task & perturbation & kept & AtomEx & LeGrad & rollout & Chefer & Grad-CAM & IG\\
\midrule
beagle & noise $0.02$ & $43$ & $0.88$ & $0.90$ & $0.94$ & $0.79$ & $0.66$ & $0.58$\\
 & noise $0.05$ & $43$ & $0.76$ & $0.81$ & $0.83$ & $0.65$ & $0.48$ & $0.39$\\
 & brightness $0.9$ & $43$ & $0.97$ & $0.98$ & $0.86$ & $0.95$ & $0.88$ & $0.74$\\
 & brightness $1.1$ & $41$ & $0.95$ & $0.96$ & $0.87$ & $0.91$ & $0.84$ & $0.71$\\
 & JPEG $75$ & $34$ & $0.84$ & $0.88$ & $0.95$ & $0.68$ & $0.53$ & $0.51$\\
 & shift $2$ px & $40$ & $0.73$ & $0.81$ & $0.78$ & $0.67$ & $0.53$ & $0.31$\\
 & shift $4$ px & $37$ & $0.91$ & $0.73$ & $0.69$ & $0.57$ & $0.41$ & $0.24$\\
\midrule
school bus & noise $0.02$ & $49$ & $0.90$ & $0.93$ & $0.94$ & $0.79$ & $0.69$ & $0.61$\\
 & noise $0.05$ & $48$ & $0.82$ & $0.88$ & $0.86$ & $0.68$ & $0.57$ & $0.46$\\
 & brightness $0.9$ & $49$ & $0.97$ & $0.98$ & $0.86$ & $0.94$ & $0.91$ & $0.70$\\
 & brightness $1.1$ & $49$ & $0.95$ & $0.97$ & $0.87$ & $0.91$ & $0.85$ & $0.67$\\
 & JPEG $75$ & $48$ & $0.88$ & $0.90$ & $0.93$ & $0.74$ & $0.63$ & $0.56$\\
 & shift $2$ px & $48$ & $0.74$ & $0.79$ & $0.71$ & $0.61$ & $0.50$ & $0.30$\\
 & shift $4$ px & $48$ & $0.91$ & $0.70$ & $0.60$ & $0.51$ & $0.41$ & $0.23$\\
\bottomrule
\end{tabular}}
\end{table}

\section{Sensitivity, the lattice and the shape of the groups}
\label{app:sweep}
\Cref{tab:sweep} varies one setting at a time, with one seed ($101$), the protocol of \cref{sec:protocol} on the $56\times56$ lattice and the best-fidelity checkpoint; \emph{alive channels} are used channels whose balanced mean area is at least $1\%$ of the sites, a reporting threshold: a channel below it is rare, not removed, and it is not shown to be causally unnecessary. The first block varies the TV weight at $p=0.25$, the second the dropout at $\rho=0.05$; the last rows give a model trained with an unconstrained readout (a CNN with free signs, otherwise the base setting $\rho=0.05$, $p=0.25$; the prototypes are trained together with the readout, so a difference is not due to the readout alone) and a logistic probe on the mean descriptors of the images (the same fitting images, no groups). \Cref{tab:sweep} shows that the fidelity is within about a point of the base setting in the first two blocks except for dropout $0$ and $0.1$ on beagle ($-2.3$ and $-1.4$ points), that without dropout school bus keeps only three alive channels and $8\%$ of the area, and that the constrained surrogate is not worse than the unconstrained one. One test positive moves the balanced accuracy by $0.78$ (beagle) and $0.89$ points (school bus), so differences below a point carry no ordering.

\begin{table}[tp]
\centering
\small
\caption{Sensitivity to the TV weight and the dropout, and the cost of the constraints (balanced fidelity, used area and number of alive channels; one seed). The linear probe is fitted on the mean-pooled descriptors and evaluated on the same test set.}
\label{tab:sweep}
\fitcolumn{\begin{tabular}{@{}llcccccc@{}}
\toprule
 & & \multicolumn{3}{c}{beagle} & \multicolumn{3}{c}{school bus}\\
\cmidrule(lr){3-5}\cmidrule(lr){6-8}
setting & value & fid.\ (\%) & area (\%) & alive & fid.\ (\%) & area (\%) & alive\\
\midrule
TV weight $\rho$ & $0$ & $98.84$ & $44.2$ & $10$ & $99.74$ & $43.4$ & $10$\\
 & $0.02$ & $98.88$ & $33.3$ & $10$ & $99.74$ & $26.5$ & $10$\\
 & $0.05$ (base) & $99.39$ & $23.6$ & $9$ & $99.83$ & $17.4$ & $7$\\
 & $0.1$ & $99.67$ & $22.4$ & $7$ & $98.89$ & $10.8$ & $5$\\
\midrule
dropout $p$ & $0$ & $97.09$ & $26.3$ & $8$ & $99.74$ & $8.1$ & $3$\\
 & $0.1$ & $98.01$ & $21.4$ & $8$ & $99.68$ & $19.4$ & $7$\\
 & $0.25$ (base) & $99.39$ & $23.6$ & $9$ & $99.83$ & $17.4$ & $7$\\
 & $0.5$ & $99.36$ & $26.3$ & $10$ & $99.67$ & $25.5$ & $9$\\
\midrule
readout & unconstrained & $97.47$ & $22.8$ & $7$ & $99.56$ & $19.1$ & $7$\\
 & linear probe & $98.5$ & -- & -- & $99.4$ & -- & --\\
\bottomrule
\end{tabular}}
\end{table}

\begin{table*}[tp]
\centering
\small
\caption{The lattice and the shape of the groups, mean over three seeds, best-fidelity checkpoint, $p=0.25$. Lattice: $16$, $4$ or $1$ shifted views per image ($56\times56$, $28\times28$, $14\times14$ sites); area penalty: the TV term replaced by the same weight times the share of the image covered by the used channels; readout dilated: the kernels of the monotone CNN dilated by $4$ or $2$, so that the readout spans as many pixels of the image as on the $14\times14$ lattice. TV$/\sqrt{\text{area}}$: hard TV divided by the square root of the used area (free of scale; not comparable across lattices). Pieces per group: connected pieces of a used group present in an image; in specks: share of the used area in pieces smaller than one token ($R\times R$ sites); both balanced over teacher-positive and teacher-negative test images (\texttt{fragmentation.py}). Box and disk precision as in \cref{tab:compare} (\texttt{box\_precision\_models.py}).}
\label{tab:shape}
\resizebox{\textwidth}{!}{\begin{tabular}{@{}llcccccccc@{}}
\toprule
task & setting & bal.\ fid.\ (\%) & area (\%) & alive & TV$/\sqrt{\text{area}}$ & pieces per group & in specks (\%) & box prec.\ (\%) & disk prec.\ (\%)\\
\midrule
beagle & $56\times56$, TV (base) & $99.24$ & $22.1$ & $8.7$ & $0.93$ & $13.9$ & $27$ & $95.0$ & $95.7$\\
 & $56\times56$, no TV & $98.71$ & $44.1$ & $10.0$ & $1.17$ & $31.5$ & $32$ & $91.1$ & $92.9$\\
 & $56\times56$, area penalty $0.03$ & $98.70$ & $42.2$ & $10.0$ & $1.16$ & $29.6$ & $31$ & $89.9$ & $92.8$\\
 & $56\times56$, area penalty $0.1$ & $98.66$ & $33.3$ & $9.7$ & $1.17$ & $25.6$ & $32$ & $92.7$ & $94.3$\\
 & $56\times56$, area penalty $0.3$ & $98.95$ & $22.6$ & $8.3$ & $1.04$ & $18.2$ & $31$ & $94.7$ & $95.8$\\
 & $56\times56$, TV, readout dilated $\times4$ & $99.00$ & $24.4$ & $8.3$ & $0.91$ & $14.0$ & $24$ & $96.1$ & $95.6$\\
 & $28\times28$, TV & $98.83$ & $30.1$ & $8.7$ & -- & -- & -- & $94.4$ & $94.6$\\
 & $28\times28$, TV, readout dilated $\times2$ & $98.87$ & $29.5$ & $9.0$ & -- & -- & -- & $94.1$ & $94.5$\\
 & $28\times28$, no TV & $98.58$ & $49.3$ & $10.0$ & -- & -- & -- & $87.7$ & $91.6$\\
 & $14\times14$, TV & $98.94$ & $36.8$ & $9.7$ & -- & -- & -- & $92.4$ & $93.2$\\
 & $14\times14$, no TV & $98.43$ & $50.9$ & $10.0$ & -- & -- & -- & $87.1$ & $91.0$\\
\midrule
school bus & $56\times56$, TV (base) & $98.89$ & $18.2$ & $7.0$ & $0.81$ & $10.8$ & $29$ & $94.7$ & $81.6$\\
 & $56\times56$, no TV & $99.49$ & $45.5$ & $10.0$ & $1.20$ & $34.8$ & $34$ & $81.1$ & $77.2$\\
 & $56\times56$, area penalty $0.03$ & $98.96$ & $34.3$ & $10.0$ & $1.28$ & $35.0$ & $42$ & $90.7$ & $80.0$\\
 & $56\times56$, area penalty $0.1$ & $98.37$ & $24.6$ & $10.0$ & $1.25$ & $29.1$ & $46$ & $91.9$ & $81.3$\\
 & $56\times56$, area penalty $0.3$ & $98.39$ & $9.1$ & $3.0$ & $1.00$ & $15.4$ & $60$ & $95.3$ & $84.8$\\
 & $56\times56$, TV, readout dilated $\times4$ & $99.48$ & $16.6$ & $6.7$ & $0.82$ & $11.3$ & $32$ & $95.8$ & $82.4$\\
 & $28\times28$, TV & $99.51$ & $22.1$ & $9.0$ & -- & -- & -- & $93.9$ & $81.1$\\
 & $28\times28$, TV, readout dilated $\times2$ & $98.55$ & $17.9$ & $6.3$ & -- & -- & -- & $95.3$ & $81.9$\\
 & $28\times28$, no TV & $99.49$ & $43.9$ & $10.0$ & -- & -- & -- & $90.1$ & $79.0$\\
 & $14\times14$, TV & $99.15$ & $25.7$ & $10.0$ & -- & -- & -- & $92.1$ & $80.2$\\
 & $14\times14$, no TV & $99.45$ & $46.3$ & $10.0$ & -- & -- & -- & $88.5$ & $78.4$\\
\midrule
valley & $56\times56$, TV (base) & $99.67$ & $20.0$ & $7.7$ & $1.10$ & $18.5$ & $45$ & -- & --\\
 & $56\times56$, no TV & $99.35$ & $67.3$ & $10.0$ & $1.13$ & $53.2$ & $41$ & -- & --\\
\bottomrule
\end{tabular}}
\end{table*}

Two further controls, with three seeds each, bear on the lattice and on what TV does (\cref{tab:shape}). The first trains the same model on fewer shifted views: one view (the native $14\times14$ grid) or four ($28\times28$). The fidelity does not change ($98.4$ to $99.5\%$), but with TV the used area stays larger on the coarser lattices (beagle $36.8$, $30.1$ and $22.1\%$ for $14$, $28$ and $56$ sites per axis; school bus $25.7$, $22.1$ and $18.2\%$) and the positive groups are $0.6$ to $2.6$ points less precise in the boxes; on school bus they stay above a centered disk on every lattice, on beagle level with it or below. The lattice thus yields finer regions rather than higher fidelity. The change of lattice also changes the span of the readout: its three $3\times3$ kernels cover $7$ sites, which is $96$ pixels between the outermost site centers on the $14\times14$ lattice but only $48$ and $24$ on the finer ones. Dilating the kernels by $2$ on the $28\times28$ lattice and by $4$ on the $56\times56$ lattice restores the span of $96$ pixels, and the ordering remains: the used area is $36.8$, $29.5$ and $24.4\%$ on beagle and $25.7$, $17.9$ and $16.6\%$ on school bus for $14$, $28$ and $56$ sites per axis, at the same fidelity and with box precisions of $94.1$ to $96.1\%$. The smaller groups of the finer lattices are therefore not an effect of the shorter span of the readout, although on school bus the difference between $28$ and $56$ sites shrinks to $1.3$ points. The second control replaces the TV term by a penalty on the used area. It shrinks the groups as well, at no cost in fidelity, but it does not simplify them: at a similar area on beagle ($22.6\%$ with weight $0.3$ against $22.1\%$ with TV) a group falls into $18.2$ pieces against $13.9$, the boundaries are longer for their size (TV per square root of the used area $1.04$ against $0.93$) and $31\%$ of the used area lies in specks smaller than one token against $27\%$; on school bus the area penalty either keeps more area ($24.6\%$ at weight $0.1$, with $29.1$ pieces per group and $1.25$ against $0.81$ per square root of area) or shrinks most groups below the alive threshold (three alive at weight $0.3$, with $60\%$ of the area in specks). TV therefore shortens the boundaries beyond what a smaller area gives, and its groups are as precise in the boxes as those of the area penalty or more. It does not remove the specks: with TV $27$ to $29\%$ of the used area of beagle and school bus ($45\%$ on valley) lies in pieces smaller than one token, against $32$ to $34\%$ ($41\%$) without, so the hard $56\times56$ maps remain fragmented, as \cref{fig:gallery-56} shows. The hard TV of the tables merges the groups of one sign, so the boundary between two positive groups does not count. Counted per group, as the trained term does ($n\,\TV_\Delta$ of the hard maps of the used channels, \texttt{group\_tv.py}), the boundary length per $16$-pixel patch is $2.60$, $2.75$ and $4.61$ without TV and $1.10$, $0.84$ and $1.16$ with TV (beagle, school bus, valley; $2.4$, $3.3$ and $4.0$ times shorter), and the share of these boundaries that separates two groups of the same sign, which the hard TV by sign does not see, falls from $36$, $37$ and $58\%$ to $19$, $16$ and $13\%$. TV thus also simplifies the cut between the groups of one sign.

The monotone CNN of the readout could be replaced by something simpler, and two checks show what it adds. First, on the test images it is nearly additive in the groups. For the three TV surrogates of the main text and test images of four kinds (teacher-positive images of the class, the teacher's false positives, sibling-class and random negatives; $792$ beagle and $714$ school-bus image--seed pairs) we computed the exact Shapley values of the used groups present in an image for the surrogate's logit, switching a group off by moving its sites to a discard channel, the move of \cref{eq:sign} ($2^j$ evaluations for the $j\le10$ groups present; \texttt{intervention\_fidelity.py}). An additive model explains $99.97\%$ of the variance of the logit over all subsets of the groups, the Shapley value of every group has the sign of its coefficient (as \cref{eq:sign} implies), and the largest group carries on average half of the total absolute contribution. The decision on an image can therefore be read as a sum of signed contributions of its groups; on the ambulance of \cref{fig:teaser}, for instance, the dark glass and the wheels add $+2.2$ and $+1.5$ to the school-bus logit and the two main negative groups $-6.4$ and $-10.7$. Second, a readout that is additive by construction, $F=g\sum_k\gamma_k\,\mathrm{area}_k+b$ with a fixed gain $g=10$, the same model and training otherwise, is as faithful but makes worse explanations (\cref{tab:readout}; three seeds, compared with a rerun of the base model with the same code on the same GPU). It rewards area, since every site of a positive group adds evidence, so TV has to work against the readout: the groups stay larger (beagle $36.6\%$ against $26.1\%$, school bus $30.5\%$ against $16.2\%$), fewer of them are switched off, they fall into more pieces, and they are $3.5$ and $4.0$ points less precise in the boxes. We read this as follows: the saturating nonlinearities of the CNN let a compact piece of a part carry the evidence, whereas a sum of areas asks for more of it. The CNN thus yields more compact explanations rather than higher fidelity.

Two further ablations in \cref{tab:readout} concern the discard channels and the training of the prototypes. Without discard channels ($m=0$) every token has to join a used group. The fidelity does not drop ($98.6$ and $99.8\%$), and the groups do not merge into one dominant group ($9.0$ and $8.3$ stay alive), but the positive groups cover $44\%$ and $30\%$ of the image instead of $14\%$ and $11\%$, each group falls into more pieces, and the box precision of the positive groups drops to $68.5$ and $76.6\%$, below a centered disk of the same area on beagle ($77.6\%$) and only $4$ points above it on school bus ($72.5\%$). Five discard channels give intermediate values. The discard channels are therefore what makes the explanations compact, while the fidelity does not need them. The second ablation freezes the prototypes at their $k$-means initialization and trains only the readout; with the additive readout this is the baseline ``cluster the descriptors, then fit a signed linear model on the cluster areas''. Its fidelity is lower ($96.6$ and $97.7\%$, against $99.0$ and $98.6\%$), the used groups cover $65\%$ and $59\%$ of the image in about $39$ pieces each, and their box precision is $88.6$ and $86.4\%$. The monotone CNN on the same frozen maps recovers part of the fidelity ($97.4$ and $98.0\%$) but cannot change the maps. Training the prototypes together with the readout and TV is therefore necessary for both the fidelity and the compactness of the groups.

\begin{table}[tp]
\centering
\small
\caption{Ablations of the readout, the discard channels and the prototypes, base setting otherwise ($\rho=0.05$, $p=0.25$, $56\times56$), mean over three seeds, all fitted with the same code on the same GPU (the base rows are a rerun of the base model, so they differ slightly from \cref{tab:tasks}). Additive readout: $F=g\sum_k\gamma_k\,\mathrm{area}_k+b$. $m$: number of discard channels ($10$ in the base model); with $m=0$ every site belongs to a used group, so TV$/\sqrt{\text{area}}$ is not comparable. Frozen prot.: the prototypes stay at their $k$-means initialization and only the readout is trained, so the maps are the same for both readouts. Columns as in \cref{tab:shape}.}
\label{tab:readout}
\fitcolumn{\begin{tabular}{@{}llccccccc@{}}
\toprule
task & variant & fid.\ (\%) & area (\%) & alive & TV$/\sqrt{\text{area}}$ & pieces & prec.\ (\%) & disk (\%)\\
\midrule
beagle & base (monotone CNN) & $99.01$ & $26.1$ & $9.0$ & $0.93$ & $14.4$ & $95.8$ & $95.2$\\
 & additive readout & $99.14$ & $36.6$ & $10.0$ & $1.05$ & $18.6$ & $92.3$ & $93.7$\\
 & $m=5$ discard ch. & $99.03$ & $29.5$ & $8.7$ & $0.97$ & $14.7$ & $92.4$ & $93.8$\\
 & $m=0$ discard ch. & $98.61$ & $100.0$ & $9.0$ & -- & $20.6$ & $68.5$ & $77.6$\\
 & frozen prot., CNN & $97.36$ & $65.0$ & $10.0$ & $1.38$ & $39.4$ & $88.6$ & $91.3$\\
 & frozen prot., additive & $96.62$ & $65.0$ & $10.0$ & $1.38$ & $39.4$ & $88.6$ & $91.3$\\
\midrule
school bus & base (monotone CNN) & $98.60$ & $16.2$ & $6.7$ & $0.80$ & $10.6$ & $94.6$ & $82.2$\\
 & additive readout & $99.70$ & $30.5$ & $9.3$ & $0.92$ & $15.7$ & $90.6$ & $79.8$\\
 & $m=5$ discard ch. & $98.87$ & $21.6$ & $7.7$ & $0.84$ & $11.2$ & $92.2$ & $81.1$\\
 & $m=0$ discard ch. & $99.76$ & $100.0$ & $8.3$ & -- & $25.7$ & $76.6$ & $72.5$\\
 & frozen prot., CNN & $97.97$ & $58.8$ & $10.0$ & $1.26$ & $38.8$ & $86.4$ & $76.6$\\
 & frozen prot., additive & $97.66$ & $58.8$ & $10.0$ & $1.26$ & $38.8$ & $86.4$ & $76.6$\\
\bottomrule
\end{tabular}}
\end{table}

\section{Twenty classes drawn at random}
\label{app:random}
\begin{table*}[tp]
\centering
\small
\caption{Twenty classes drawn at random (mean over the seeds $101$, $211$ and $307$, $56\times56$ lattice, best-fidelity checkpoint), with ($\rho=0.05$) and without TV ($\rho=0$). Card: the integer scorecard of \cref{eq:card} read from the groups of the $\rho=0.05$ surrogate (\cref{app:card}). Images: teacher-positive test images of the class with an ImageNet box that does not fill the crop. Alive groups: used channels with at least $1\%$ of the balanced area. Box prec.: share of the positive-group area inside the boxes, averaged over the images; disk: the same for a centered disk of the same area on the same images.}
\label{tab:random}
\setlength{\tabcolsep}{5pt}
\begin{tabular}{@{}lcccccccccccc@{}}
\toprule
 & & \multicolumn{3}{c}{bal.\ fid.\ (\%)} & \multicolumn{2}{c}{area (\%)} & \multicolumn{2}{c}{hard TV} & \multicolumn{2}{c}{alive groups} & \multicolumn{2}{c}{box prec.\ (\%), $\rho=0.05$}\\
\cmidrule(lr){3-5}\cmidrule(lr){6-7}\cmidrule(lr){8-9}\cmidrule(lr){10-11}\cmidrule(l){12-13}
class (in the order drawn) & images & $\rho=0$ & $0.05$ & card & $\rho=0$ & $0.05$ & $\rho=0$ & $0.05$ & $\rho=0$ & $0.05$ & groups & disk\\
\midrule
guinea pig & $43$ & $98.6$ & $99.5$ & $99.5$ & $42.6$ & $15.0$ & $0.78$ & $0.32$ & $10.0$ & $4.3$ & $94.6$ & $95.2$\\
bullet train & $49$ & $99.3$ & $99.4$ & $98.6$ & $56.8$ & $17.0$ & $0.90$ & $0.33$ & $10.0$ & $4.7$ & $67.9$ & $78.5$\\
Old English sheepdog & $40$ & $99.1$ & $99.5$ & $99.5$ & $38.4$ & $17.3$ & $0.61$ & $0.32$ & $10.0$ & $7.3$ & $99.6$ & $84.5$\\
hot pot & $31$ & $99.5$ & $99.3$ & $98.8$ & $48.4$ & $17.3$ & $0.90$ & $0.48$ & $10.0$ & $6.7$ & $80.3$ & $75.4$\\
fly & $36$ & $99.4$ & $99.4$ & $98.4$ & $47.6$ & $11.0$ & $0.95$ & $0.29$ & $10.0$ & $4.3$ & $87.5$ & $91.1$\\
abacus & $39$ & $97.1$ & $99.3$ & $98.8$ & $41.2$ & $18.0$ & $0.77$ & $0.35$ & $10.0$ & $5.3$ & $95.8$ & $86.9$\\
go-kart & $47$ & $98.5$ & $98.5$ & $98.1$ & $53.3$ & $16.1$ & $0.93$ & $0.41$ & $10.0$ & $6.3$ & $60.7$ & $58.4$\\
robin & $47$ & $99.8$ & $99.9$ & $99.1$ & $38.4$ & $10.7$ & $0.86$ & $0.23$ & $10.0$ & $4.3$ & $89.8$ & $82.2$\\
frying pan & $20$ & $97.0$ & $97.3$ & $95.5$ & $41.2$ & $19.6$ & $0.88$ & $0.45$ & $10.0$ & $8.7$ & $93.7$ & $89.8$\\
shoe shop & $18$ & $96.1$ & $97.0$ & $96.1$ & $40.8$ & $13.2$ & $1.02$ & $0.43$ & $10.0$ & $4.0$ & $79.3$ & $72.1$\\
paddle & $37$ & $98.0$ & $98.6$ & $96.7$ & $42.6$ & $17.6$ & $0.78$ & $0.40$ & $10.0$ & $7.7$ & $57.4$ & $63.5$\\
beer bottle & $40$ & $98.6$ & $98.7$ & $98.9$ & $44.3$ & $20.9$ & $0.92$ & $0.48$ & $10.0$ & $8.0$ & $80.9$ & $72.6$\\
prairie chicken & $49$ & $99.6$ & $99.6$ & $99.1$ & $48.5$ & $13.5$ & $0.97$ & $0.32$ & $10.0$ & $5.0$ & $78.7$ & $77.1$\\
vault & $33$ & $98.4$ & $98.6$ & $97.9$ & $59.7$ & $40.7$ & $1.05$ & $0.81$ & $10.0$ & $9.3$ & $63.3$ & $60.8$\\
bearskin & $46$ & $98.2$ & $97.6$ & $96.9$ & $44.7$ & $16.7$ & $0.82$ & $0.37$ & $10.0$ & $6.7$ & $39.3$ & $17.8$\\
pinwheel & $39$ & $99.1$ & $99.3$ & $98.9$ & $44.3$ & $22.7$ & $0.75$ & $0.47$ & $10.0$ & $8.7$ & $89.8$ & $82.9$\\
rapeseed & $46$ & $100.0$ & $99.7$ & $99.8$ & $60.7$ & $20.7$ & $0.79$ & $0.38$ & $10.0$ & $4.0$ & $76.1$ & $39.0$\\
nematode & $35$ & $99.8$ & $99.4$ & $99.6$ & $60.4$ & $16.4$ & $0.97$ & $0.34$ & $9.3$ & $3.3$ & $83.8$ & $71.8$\\
tench & $48$ & $99.6$ & $98.5$ & $98.9$ & $50.4$ & $16.2$ & $0.97$ & $0.33$ & $10.0$ & $6.3$ & $69.6$ & $61.3$\\
African elephant & $26$ & $99.3$ & $99.3$ & $99.4$ & $45.4$ & $23.6$ & $0.81$ & $0.45$ & $10.0$ & $7.0$ & $94.6$ & $81.4$\\
\midrule
mean & $38$ & $98.7$ & $98.9$ & $98.4$ & $47.5$ & $18.2$ & $0.87$ & $0.40$ & $10.0$ & $6.1$ & $79.1$ & $72.1$\\
\bottomrule
\end{tabular}
\end{table*}

\begin{figure*}[p]
\centering
\includegraphics[width=.77\textwidth]{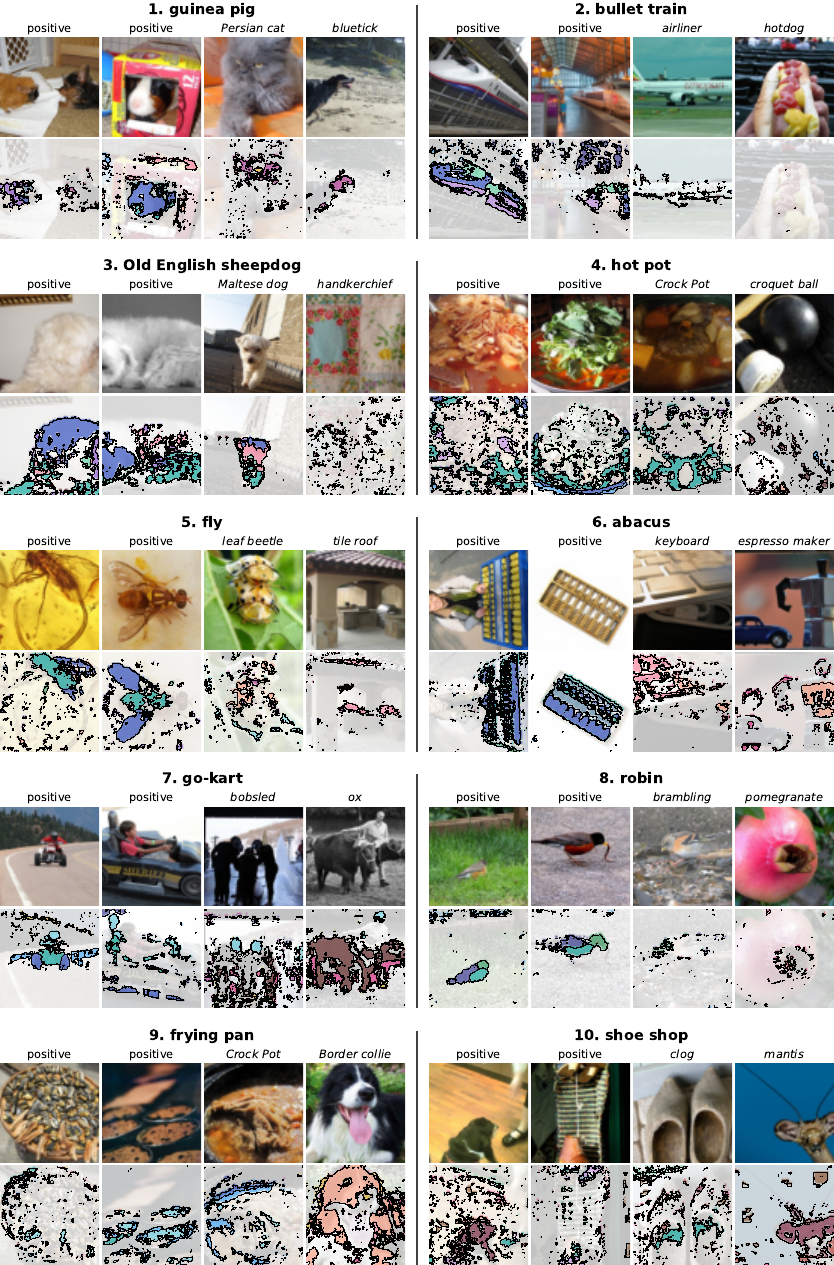}
\caption{Draws $1$--$10$ of the twenty random classes of \cref{app:random}: the $56\times56$ TV model of each class (seed $101$) on its first two teacher-positive test images, the first teacher-negative image of a sibling class and one random teacher-negative image of another class (labels: ImageNet class of the negatives), chosen before any map was drawn; dense maps drawn as in \cref{fig:gallery-56} (cool colors: positive groups, warm: negative). A color names a group of one model only.}
\label{fig:random-a}
\end{figure*}

\begin{figure*}[p]
\centering
\includegraphics[width=.8\textwidth]{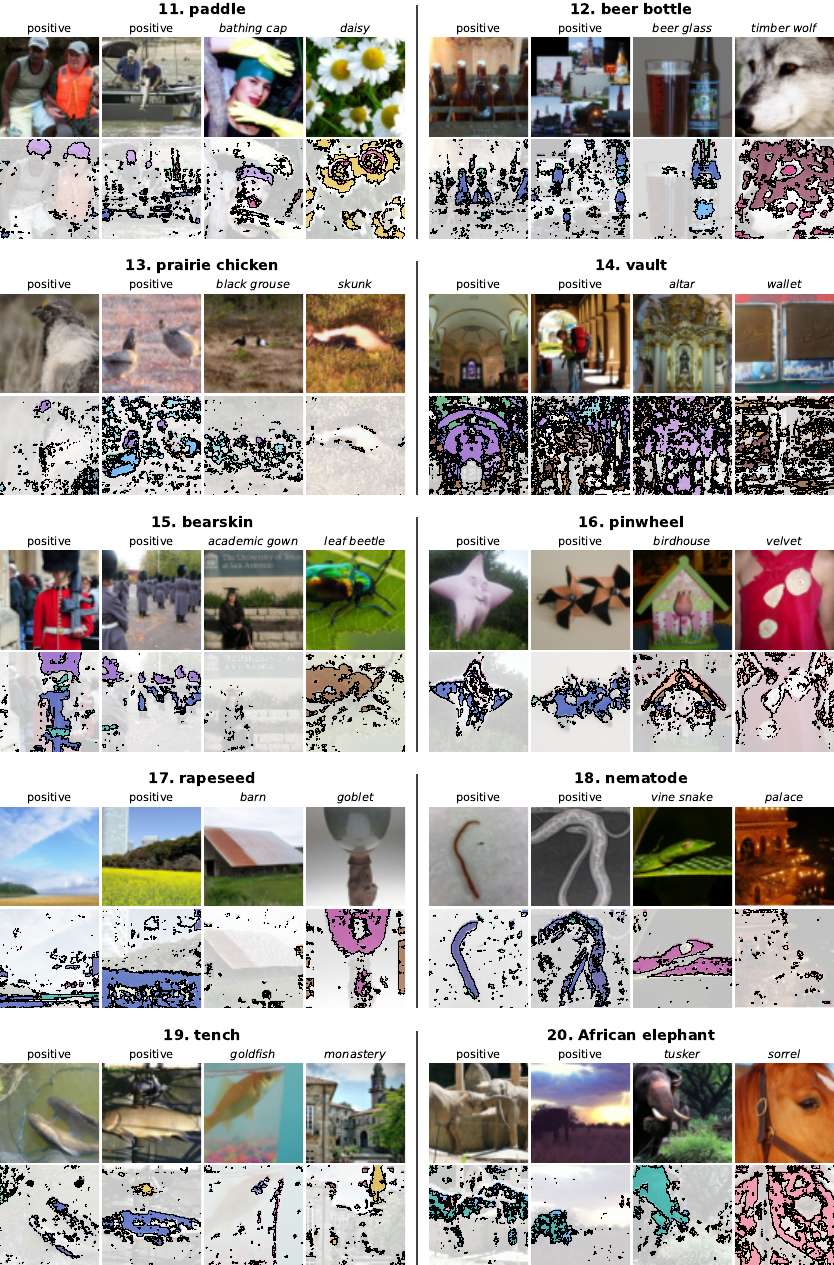}
\caption{Draws $11$--$20$ of the twenty random classes, as in \cref{fig:random-a}.}
\label{fig:random-b}
\end{figure*}

The three classes of the main text and the ten of \cref{app:classes} were chosen by us as common, familiar categories with clearly recognizable evidence: a dog breed, a vehicle and a landscape in the main text, and everyday objects and animals with public ImageNet-S masks in \cref{app:classes}. To make sure that the results do not depend on this choice, we drew twenty further classes at random and ran them with the protocol unchanged. The rule was fixed before the draw: the $987$ ImageNet classes not used before, in the order of their class index, are sampled with \texttt{random.Random(20261108).sample}, and the first twenty draws are taken in the order drawn (\texttt{draw20.py}); no class was skipped or replaced. All twenty classes were run with the three seeds of the main text ($101$, $211$ and $307$). Every class gets the base setting ($\rho=0.05$, $p=0.25$) and its $\rho=0$ partner on the $56\times56$ lattice, with the sets built as in \cref{app:store} (the same $4{,}000$ random validation images in the test set, plus the teacher-positive validation images of the class), and we report the best-fidelity checkpoint.

\paragraph{Result.} \Cref{tab:random} gives every class as the mean over three seeds. The fidelity with TV is $98.9\%$ on average, from $97.0\%$ (shoe shop) to $99.9\%$ (robin), and $98.7\%$ without it; over single seeds it ranges from $96.3\%$ to $100\%$. TV shrinks the used area from $47.5\%$ to $18.2\%$ of the image and more than halves the hard TV ($0.87$ to $0.40$), and it shrinks groups below the alive threshold of $1\%$: without TV all ten used channels are alive (nine in one seed of nematode), with TV $3.3$ (nematode) to $9.3$ (vault) on average over the seeds, $6.1$ in all. One class follows only in part: on vault, whose images are ceilings and arcades that fill the crop, TV leaves the area at $63.5\%$ in seed $101$ and shrinks it to $35\%$ and $23\%$ in the other two. Against the ImageNet boxes ($18$ to $49$ teacher-positive images per class with a box that does not fill the crop, $769$ in all) the positive groups with TV have a mean precision of $79.1\%$ against $72.1\%$ for a centered disk of the same area; they are above the disk on $16$ of the $20$ classes and in $47$ of the $60$ class--seed pairs. Fly, bullet train and paddle are below it in every seed ($4$, $11$ and $6$ points on average); guinea pig is level ($94.6\%$ against $95.2\%$, above in one seed), and Old English sheepdog is above it in every seed ($99.3$--$99.8\%$ against $84.0$--$84.9\%$). The margin is a few points on most classes and large where the box is small against the crop (bearskin: $39.3\%$ against $17.8\%$; rapeseed: $76.1\%$ against $39.0\%$). Without TV the precision is $67.5\%$ against $68.0\%$ for the disk (above it on $10$ classes), so here too TV is what moves the groups onto the object. The random draw thus reproduces what the chosen classes show (\cref{tab:classes}: fidelity $99.1\%$, area $53\%$ to $20\%$, above the disk on seven of ten).

\paragraph{Scorecards.} For every class and seed we read an integer card (\cref{eq:card}) from the groups of the TV surrogate exactly as for the three main tasks (\cref{app:card}: areas on the fit, selection and test images of the class, card fitted to the teacher's decisions on the groups with $\gamma_k>0$, integer points). Over the $60$ models the cards reach $98.42\%$ balanced fidelity against $98.92\%$ for their surrogates and decide like them on $98.6\%$ of the test images (\cref{tab:random}, column card). Per class the card is within $0.5$ points of the surrogate on $9$ of the $20$ classes; it is lowest on frying pan ($95.5\%$, surrogate $97.3\%$), shoe shop ($96.1\%$) and paddle ($96.7\%$), and the largest losses against the surrogate are on frying pan and paddle ($1.8$ points) and fly ($1.0$). $58$ of the $60$ cards have four to six groups (one has three, one two), $37$ have points of at most $5$, the largest point is $11$, and the cards need $1$ to $8$ points. No card accepts an image on which no group of the class counts. The two-sided card is more faithful by $0.33$ points on average and less faithful on seven of the twenty classes (\cref{app:sides}).

\paragraph{Images.} \Cref{fig:random-a,fig:random-b} show every class on four test images chosen by a fixed rule before any map was drawn (\texttt{pick\_four.py}): the first two teacher-positive test images in the official validation order; the first teacher-negative test image of a sibling class; and one teacher-negative test image of any other class, drawn with \texttt{random.Random(20261108 + class index)} (the first such image in the official order would be the same tench for almost every class). The sibling classes of a class are the five classes whose rows of the teacher's linear head are closest to its row by cosine (\texttt{head\_siblings.py}), so they are close in the teacher's own decision, \eg airliner for bullet train and computer keyboard for abacus (for beagle and school bus, \cref{app:hard} uses taxonomic sets as well as these nearest classes). The maps are dense maps (\cref{app:finer}) of the $56\times56$ TV models of seed $101$, drawn as in \cref{fig:gallery-56}. With its threshold from the selection set the surrogate accepts $39$ of the $40$ positives shown (not the first shoe-shop image), rejects all $20$ other negatives and $14$ of the $20$ sibling negatives; the six it accepts are a crock pot for hot pot and for frying pan, a beer glass next to a beer bottle, a black grouse for prairie chicken, an altar for vault and a tusker for African elephant. This is the weakness of \cref{app:hard}: sibling-class negatives are where the surrogate departs from the teacher most.

On most classes the positive groups lie on the object or its defining part: the beads and frame of the abacus, the body of the fly, the robin and the nematode, the fish of tench, the pinwheels, the yellow field of rapeseed, the black hats and red tunics of bearskin (the class is named after the hat). Some classes are explained by context: on paddle the positive groups lie on the caps of the people and along the boat, and no paddle is visible in either image; on vault TV leaves groups on the whole ceiling and the walls. As with the wheels of the school bus (\cref{sec:groups}), positive groups also appear on negatives that share parts with the class: the bottle in the beer-glass image, the body of the tusker, the pot of the crock pot, the bathing cap for paddle. On the other negatives most of the colored area is negative (warm colors): the keyboard for abacus, the collie for frying pan, the wolf for beer bottle, the horse for African elephant, the ox for go-kart. As in \cref{fig:gallery-56}, the dense maps of the $56\times56$ models are fragmented, with small islands around the main groups.

\section{Ten further classes chosen by us}
\label{app:classes}
\begin{table}[tp]
\centering
\small
\caption{Ten further classes, mean over seeds $101$, $211$ and $307$, $56\times56$ lattice, best-fidelity checkpoint. Images: teacher-positive test images of the class with an ImageNet box that does not fill the crop. Prec.\ and AUC, both against these boxes: share of the positive-group area inside the boxes, and the ROC AUC of the binary region ($(\mathrm{recall}+\mathrm{specificity})/2$), averaged over the images; the disk columns give the same for a centered disk of the same area on the same images. The mean rows average the ten classes.}
\label{tab:classes}
\fitcolumn{\begin{tabular}{@{}llcccccccc@{}}
\toprule
class & $\rho$ & images & bal.\ fid.\ (\%) & area (\%) & hard TV & prec.\ (\%) & disk prec.\ (\%) & AUC (\%) & disk AUC (\%)\\
\midrule
tabby cat & $0$ & $24$ & $99.6$ & $52.2$ & $0.82$ & $93.9$ & $95.2$ & $67.5$ & $69.3$\\
 & $0.05$ & $24$ & $99.7$ & $27.9$ & $0.53$ & $95.2$ & $96.9$ & $63.1$ & $64.4$\\
zebra & $0$ & $37$ & $97.9$ & $50.5$ & $0.86$ & $80.1$ & $77.6$ & $65.9$ & $68.8$\\
 & $0.05$ & $37$ & $98.2$ & $23.6$ & $0.35$ & $93.3$ & $81.3$ & $70.6$ & $65.2$\\
lion & $0$ & $34$ & $98.7$ & $48.3$ & $0.78$ & $94.5$ & $92.6$ & $65.5$ & $67.0$\\
 & $0.05$ & $34$ & $99.6$ & $11.7$ & $0.27$ & $98.4$ & $94.7$ & $59.0$ & $57.9$\\
fire engine & $0$ & $34$ & $99.2$ & $49.8$ & $0.77$ & $89.7$ & $86.1$ & $71.6$ & $69.6$\\
 & $0.05$ & $34$ & $98.2$ & $19.9$ & $0.45$ & $95.0$ & $89.3$ & $64.4$ & $62.7$\\
airliner & $0$ & $43$ & $99.6$ & $56.5$ & $0.92$ & $57.2$ & $68.3$ & $57.7$ & $70.6$\\
 & $0.05$ & $43$ & $99.8$ & $19.4$ & $0.42$ & $80.3$ & $83.5$ & $63.8$ & $65.7$\\
speedboat & $0$ & $47$ & $99.4$ & $55.7$ & $0.87$ & $51.3$ & $51.4$ & $67.0$ & $68.5$\\
 & $0.05$ & $47$ & $99.2$ & $18.4$ & $0.42$ & $75.2$ & $63.9$ & $70.9$ & $65.0$\\
banana & $0$ & $39$ & $99.5$ & $45.4$ & $0.84$ & $71.9$ & $66.9$ & $68.8$ & $64.0$\\
 & $0.05$ & $39$ & $99.3$ & $22.9$ & $0.48$ & $75.6$ & $69.0$ & $65.4$ & $60.4$\\
pizza & $0$ & $26$ & $99.4$ & $51.2$ & $0.95$ & $98.5$ & $84.1$ & $76.4$ & $68.5$\\
 & $0.05$ & $26$ & $99.1$ & $19.1$ & $0.47$ & $98.9$ & $85.3$ & $65.4$ & $60.3$\\
mountain bike & $0$ & $42$ & $99.1$ & $54.7$ & $0.92$ & $55.3$ & $56.2$ & $64.3$ & $65.1$\\
 & $0.05$ & $42$ & $98.9$ & $21.4$ & $0.49$ & $68.1$ & $62.3$ & $63.0$ & $59.9$\\
steam locomotive & $0$ & $44$ & $98.9$ & $62.0$ & $0.88$ & $70.7$ & $76.6$ & $62.9$ & $71.6$\\
 & $0.05$ & $44$ & $99.2$ & $15.6$ & $0.37$ & $78.6$ & $87.2$ & $59.0$ & $64.3$\\
\midrule
mean & $0$ & & $99.1$ & $52.6$ & $0.86$ & $76.3$ & $75.5$ & $66.8$ & $68.3$\\
 & $0.05$ & & $99.1$ & $20.0$ & $0.42$ & $85.9$ & $81.3$ & $64.5$ & $62.6$\\
\bottomrule
\end{tabular}}
\end{table}

\begin{figure*}[tp]
\centering
\includegraphics[width=\textwidth]{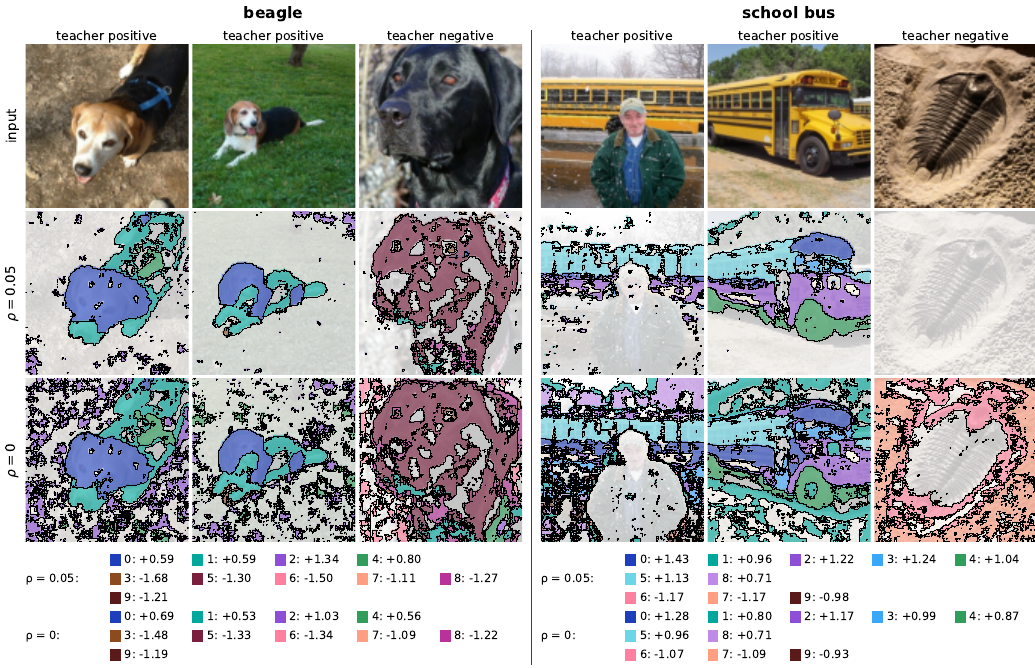}
\caption{DeiT-III-B/16 groups on the images of \cref{fig:gallery-56}: the $56\times56$ models of seed $101$ with TV ($\rho=0.05$) and without it ($\rho=0$), as dense maps drawn as in \cref{fig:gallery-56}, which shows the ViT-B/16 models on the same images. On beagle a group of the head and one of the body, as with the ViT-B/16; on school bus the groups cover $24\%$ and $44\%$ of the two bus images ($23\%$ and $31\%$ for the ViT-B/16) and nothing of the fossil; without TV the groups spread over the background.}
\label{fig:deit}
\end{figure*}

To check that the results are not specific to the three classes of the main text we ran the base setting ($\rho=0.05$, $p=0.25$) and its $\rho=0$ partner on ten further classes that we chose among those with public ImageNet-S masks (tabby cat, zebra, lion, fire engine, airliner, speedboat, banana, pizza, mountain bike and steam locomotive), seeds $101$, $211$ and $307$, the same protocol and lattice ($56\times56$, best-fidelity checkpoint); \cref{tab:classes} gives the result as a mean over the three seeds. The fidelity is $99.1\%$ with and without TV; TV shrinks the used area from $53\%$ to $20\%$ of the image and halves the hard TV ($0.86$ to $0.42$). Against the ImageNet boxes ($24$ to $47$ teacher-positive images per class) the positive groups with TV have a mean precision of $85.9\%$ against $81.3\%$ for a centered disk of the same area, and they are above the disk on seven of the ten classes (below it on tabby cat, airliner and steam locomotive; per seed on eight, eight and seven classes); without TV the mean precision is $76.3\%$ against $75.5\%$ for the disk. The AUC of the binary region is $64.5\%$ against $62.6\%$ with TV and $66.8\%$ against $68.3\%$ without. As on school bus (and unlike beagle), TV keeps the fidelity, makes the regions compact and, on most classes, more precise than a disk of the same area; the mean margin ($85.9\%$ against $81.3\%$) is moderate, and a difference of a few points on one class should not be over-read.

\section{A second plain ViT: DeiT-III-B/16}
\label{app:deit}
We repeated the protocol of \cref{sec:protocol} with DeiT-III-B/16~\cite{touvron2022deit3} (timm, trained on ImageNet-1k only like our ViT-B/16, $83.4\%$ top-1 with our preprocessing). It has the same $14\times14$ tokens of stride $16$, so four views per axis give the same $56\times56$ lattice and the same reflection-padded views; the descriptor is the layer-normed input of the last block at the patches, and everything else is as in the main text (the same three tasks, three seeds, $\rho\in\{0,0.05\}$, $p=0.25$, $18$ fits). \Cref{tab:deit} gives the result next to the ViT-B/16 of the main text, and \cref{fig:deit} shows the groups on the images of \cref{fig:gallery-56}. The fidelity is $98.7$--$99.4\%$ with TV (mean $99.1\%$) and $97.9$--$99.7\%$ without, and TV reduces the used area from $54$, $57$ and $60\%$ to $21$, $20$ and $26\%$ and the hard TV from $0.81$, $0.69$ and $0.81$ to $0.39$, $0.28$ and $0.52$, as with the ViT-B/16. In the boxes the groups with TV are above a centered disk of the same area on school bus ($87.0\%$ against $79.4\%$, lift $1.45$; ViT-B/16: $94.7\%$ against $81.6\%$) and below it on beagle ($89.9\%$ against $93.4\%$; ViT-B/16: $95.0\%$ against $95.7\%$); their precision is $5.1$ (beagle) and $7.7$ points (school bus) lower than that of the ViT-B/16. The groups do not follow a position template more than those of the ViT-B/16: the correlation of the positive map of an image with the mean map of the other images is $0.27$ on both tasks ($0.28$ and $0.23$ for the ViT-B/16), and with its own box $0.29$ on beagle and $0.43$ on school bus ($0.29$ and $0.46$). The findings of the main text thus carry over to a second plain ViT with a different training recipe: the fidelity and the effect of TV on the area are similar, and the groups lie on the object more than a disk does on school bus but not on beagle, at a precision $5$ to $8$ points lower.

\begin{table}[tp]
\centering
\small
\caption{DeiT-III-B/16 next to the ViT-B/16 of the main text on the $56\times56$ lattice, best-fidelity checkpoint, mean $\pm$ standard deviation over three seeds. Prec.: share of the positive-group area inside the object box ($43$ beagle and $49$ school-bus images for each backbone; the teacher-positive images are those of each backbone); disk prec.: the same for a centered disk of the same area on the same images.}
\label{tab:deit}
\fitcolumn{\begin{tabular}{@{}lllccccc@{}}
\toprule
task & backbone & $\rho$ & bal.\ fid.\ (\%) & hard TV & area (\%) & prec.\ (\%) & disk prec.\ (\%)\\
\midrule
beagle & ViT-B/16 & $0$ & $98.71\pm0.22$ & $0.78\pm0.08$ & $44.1\pm4.2$ & $91.1\pm8.3$ & $92.9\pm2.7$\\
& ViT-B/16 & $0.05$ & $99.24\pm0.38$ & $0.44\pm0.04$ & $22.1\pm3.1$ & $95.0\pm5.5$ & $95.7\pm1.1$\\
& DeiT-III-B/16 & $0$ & $97.87\pm0.41$ & $0.81\pm0.03$ & $54.3\pm9.7$ & $75.9\pm2.0$ & $85.3\pm1.7$\\
& DeiT-III-B/16 & $0.05$ & $98.69\pm1.16$ & $0.39\pm0.15$ & $21.3\pm12.0$ & $89.9\pm1.6$ & $93.4\pm2.7$\\
\midrule
school bus & ViT-B/16 & $0$ & $99.49\pm0.43$ & $0.81\pm0.03$ & $45.5\pm4.1$ & $81.1\pm1.3$ & $77.2\pm0.6$\\
& ViT-B/16 & $0.05$ & $98.89\pm1.48$ & $0.35\pm0.04$ & $18.2\pm0.9$ & $94.7\pm0.7$ & $81.6\pm0.3$\\
& DeiT-III-B/16 & $0$ & $99.65\pm0.50$ & $0.69\pm0.06$ & $57.2\pm2.1$ & $67.0\pm2.8$ & $67.5\pm1.2$\\
& DeiT-III-B/16 & $0.05$ & $99.35\pm0.53$ & $0.28\pm0.07$ & $19.7\pm3.7$ & $87.0\pm5.7$ & $79.4\pm1.4$\\
\midrule
valley & ViT-B/16 & $0$ & $99.35\pm0.43$ & $0.93\pm0.05$ & $67.3\pm0.9$ & -- & --\\
& ViT-B/16 & $0.05$ & $99.67\pm0.17$ & $0.48\pm0.15$ & $20.0\pm10.2$ & -- & --\\
& DeiT-III-B/16 & $0$ & $99.08\pm0.87$ & $0.81\pm0.01$ & $60.0\pm1.2$ & -- & --\\
& DeiT-III-B/16 & $0.05$ & $99.37\pm0.52$ & $0.52\pm0.09$ & $25.5\pm8.2$ & -- & --\\
\bottomrule
\end{tabular}}
\end{table}

\section{A hierarchical backbone: Swin-B}
\label{app:swin}
We repeated the protocol of \cref{sec:protocol} with the torchvision Swin-B (ImageNet-1k, $82.0\%$ top-1 with our preprocessing). Its last stage is a $7\times7$ map of stride $32$ that a linear head reads after layer norm and mean pooling, so the descriptor of a position is that layer norm applied at the position, and four shifted views per axis give a $28\times28$ lattice (sites $8$ pixels apart); everything else is as in the main text (the same three tasks, three seeds, $p=0.25$, four TV weights $\rho\in\{0,0.05,0.1,0.2\}$, $36$ fits; $\rho=0.05$ was chosen for the ViT-B/16, so $0.1$ and $0.2$ were added to retune the TV weight for Swin-B). \Cref{tab:swin} gives the result. The fidelity ($98.8$--$99.6\%$) and the area reduction with TV carry over, but the hard TV does not fall at the base weight: at $\rho=0.05$ it is $0.66$ against $0.63$ on beagle, $0.60$ against $0.59$ on valley and $0.48$ against $0.48$ on school bus; for $\rho\ge0.1$ it falls on school bus ($0.31$ and $0.27$) and valley ($0.50$ and $0.46$) but hardly on beagle ($0.58$ and $0.60$), and TV per area rises for every $\rho>0$. The alignment with the object does not carry over either: on beagle the positive groups are less precise than a centered disk of the same area for every $\rho$, and on school bus only $\rho=0.05$ is above the disk ($79.7\%$ against $73.8\%$). Larger $\rho$ shrinks the groups but the disk gains more from the smaller area. The groups of Swin-B follow a position template more than those of the ViT-B/16: the correlation of the positive map of an image with the mean map of the other images is $0.50$ and $0.54$, against $0.28$ and $0.23$. We read this as a limit of the method on this backbone. Swin-B differs from the ViT-B/16 in the size of the cells ($32$ pixels), the architecture (windowed attention, patch merging), the descriptor and the spacing of the lattice at once, so the failure cannot be attributed to the $32$-pixel cells alone; we did not run a control that changes one of them. The box precision of this table places a site on its block of pixels without the half-step shift of \cref{app:loc}, a difference below a point in the ViT-B/16 tables.

\begin{table}[tp]
\centering
\small
\caption{Swin-B on a $28\times28$ lattice, best-fidelity checkpoint, mean $\pm$ standard deviation over three seeds. Prec.: share of the positive-group area inside the object box ($43$ beagle and $48$ school-bus images); disk prec.: the same for a centered disk of the same area on the same images. The teacher-positive images are those of the Swin teacher, so the images differ from those of the ViT tables. For school bus at $\rho=0.2$ one seed has a nonempty positive map on only $18$ of the $48$ images, which gives the large spread of the precision there.}
\label{tab:swin}
\fitcolumn{\begin{tabular}{@{}llcccccc@{}}
\toprule
task & $\rho$ & bal.\ fid.\ (\%) & hard TV & area (\%) & TV/area & prec.\ (\%) & disk prec.\ (\%)\\
\midrule
beagle & $0$ & $99.42\pm0.44$ & $0.63\pm0.11$ & $59.9\pm7.9$ & $2.16$ & $68.7\pm3.9$ & $74.2\pm3.6$\\
 & $0.05$ & $99.15\pm0.78$ & $0.66\pm0.08$ & $49.9\pm9.5$ & $2.77$ & $73.3\pm5.4$ & $79.4\pm5.0$\\
 & $0.1$ & $99.40\pm0.47$ & $0.58\pm0.16$ & $35.9\pm7.7$ & $3.35$ & $81.5\pm10.9$ & $86.1\pm5.9$\\
 & $0.2$ & $99.35\pm0.49$ & $0.60\pm0.21$ & $35.3\pm7.2$ & $3.44$ & $76.5\pm9.4$ & $86.2\pm3.3$\\
\midrule
school bus & $0$ & $99.38\pm0.51$ & $0.48\pm0.03$ & $67.7\pm4.3$ & $1.43$ & $60.4\pm2.2$ & $64.4\pm1.0$\\
 & $0.05$ & $99.64\pm0.49$ & $0.48\pm0.09$ & $36.5\pm4.8$ & $2.59$ & $79.7\pm0.8$ & $73.8\pm0.4$\\
 & $0.1$ & $99.27\pm0.62$ & $0.31\pm0.09$ & $18.2\pm4.9$ & $3.68$ & $74.5\pm7.0$ & $82.4\pm6.8$\\
 & $0.2$ & $98.92\pm0.18$ & $0.27\pm0.04$ & $18.3\pm6.5$ & $3.17$ & $71.6\pm31.4$ & $80.1\pm6.4$\\
\midrule
valley & $0$ & $99.09\pm0.01$ & $0.59\pm0.04$ & $64.7\pm4.7$ & $1.85$ & -- & --\\
 & $0.05$ & $98.82\pm1.29$ & $0.60\pm0.03$ & $39.7\pm11.9$ & $3.21$ & -- & --\\
 & $0.1$ & $99.10\pm0.84$ & $0.50\pm0.09$ & $35.0\pm5.9$ & $2.87$ & -- & --\\
 & $0.2$ & $99.08\pm0.84$ & $0.46\pm0.02$ & $31.7\pm7.1$ & $2.97$ & -- & --\\
\bottomrule
\end{tabular}}
\end{table}

\section{Fidelity on hard negatives}
\label{app:hard}
The balanced fidelity of the main text is measured on a test set enriched with positives against mostly random negatives, and most of these negatives are easy. The test images also contain the teacher-negative images of classes close to the task class: the other dog breeds for beagle ($470$ images, ImageNet classes $151$ to $268$ except beagle) and road vehicles for school bus ($82$ images, $19$ classes such as minibus, trolleybus, fire engine, police van and limousine). \Cref{tab:hard} separates them from the rest (true positive rate TPR and true negative rate TNR at the selection threshold), from the stored best-fidelity predictions of the models of the main text. The true negative rate on the hard negatives is $92.7\%$ (beagle) and $87.4\%$ (school bus) with TV, against $99.8\%$ on all other negatives, so the surrogate decides positive for $7\%$ and $13\%$ of the sibling images that the teacher does not call the class. The balanced accuracy against the hard negatives alone is $96.1\%$ and $92.8\%$ with TV; TV changes it by less than the spread over seeds. The $99\%$ of the main text is therefore a statement about the decision between the class and images that are mostly unrelated to it; between the class and its close siblings the surrogate reproduces the teacher less well. At a natural prevalence $\pi=0.1\%$ the precision $\mathrm{TPR}\,\pi/(\mathrm{TPR}\,\pi+(1-\mathrm{TNR})(1-\pi))$ with the TPR and the TNR over all negatives of \cref{tab:hard} would be $9\%$ for beagle and $18$ to $19\%$ for school bus; this is an extrapolation from the rates, not a measured precision. On the test set itself the average precision of the surrogate's score against the teacher's decision is $0.948$ (beagle) and $0.954$ (school bus) with TV, and the precision at the selection threshold $0.61$ and $0.76$ (positives are enriched, $64$ and $56$ of $4{,}158$).

\begin{table*}[tp]
\centering
\small
\caption{The surrogates (TV, best-fidelity checkpoint) on the $250$ validation images of the five classes nearest to each class in the teacher's head. Teacher FP: share of these images that the teacher calls the class; TNR: share of the others that the surrogate rejects (mean over the seeds); TPR: true positive rate on the test set; bal.: mean of TNR and TPR.}
\label{tab:siblings}
\resizebox{\textwidth}{!}{\begin{tabular}{@{}llccccc@{}}
\toprule
class & five nearest classes (teacher's head) & seeds & teacher FP (\%) & TNR (\%) & TPR (\%) & bal.\ (\%)\\
\midrule
guinea pig & hamster, Angora, beaver, marmoset, Persian cat & $3$ & $0.0$ & $69.3$ & $99.3$ & $84.3$\\
bullet train & electric locomotive, passenger car, airliner, streetcar, police van & $3$ & $0.0$ & $88.8$ & $99.4$ & $94.1$\\
Old English sheepdog & komondor, Tibetan terrier, Dandie Dinmont, Maltese dog, otterhound & $3$ & $1.2$ & $70.6$ & $99.4$ & $85.0$\\
hot pot & consomme, wok, Crock Pot, soup bowl, potpie & $3$ & $6.8$ & $67.8$ & $99.5$ & $83.7$\\
fly & bee, leaf beetle, dragonfly, damselfly, cicada & $3$ & $0.8$ & $54.2$ & $99.3$ & $76.8$\\
abacus & computer keyboard, menu, confectionery, slide rule, marimba & $3$ & $0.0$ & $95.1$ & $99.3$ & $97.2$\\
go-kart & racer, lawn mower, bobsled, tricycle, sports car & $3$ & $0.8$ & $56.5$ & $98.1$ & $77.3$\\
robin & junco, house finch, brambling, coucal, bulbul & $3$ & $0.0$ & $75.5$ & $100.0$ & $87.7$\\
frying pan & wok, Dutch oven, spatula, waffle iron, Crock Pot & $3$ & $2.0$ & $31.4$ & $98.3$ & $64.9$\\
shoe shop & sandal, toyshop, clog, tobacco shop, running shoe & $3$ & $2.4$ & $45.5$ & $96.2$ & $70.9$\\
paddle & canoe, speedboat, gondola, bathing cap, boathouse & $3$ & $0.4$ & $29.3$ & $100.0$ & $64.7$\\
beer bottle & pop bottle, wine bottle, beer glass, bottlecap, water bottle & $3$ & $3.6$ & $49.5$ & $98.7$ & $74.1$\\
prairie chicken & black grouse, ruffed grouse, partridge, ptarmigan, bustard & $3$ & $0.4$ & $54.2$ & $100.0$ & $77.1$\\
vault & monastery, altar, church, dome, triumphal arch & $3$ & $4.0$ & $46.4$ & $98.3$ & $72.3$\\
bearskin & pickelhaube, military uniform, academic gown, cuirass, palace & $3$ & $0.8$ & $70.3$ & $95.8$ & $83.1$\\
pinwheel & umbrella, maypole, flagpole, mortarboard, birdhouse & $3$ & $0.0$ & $82.9$ & $99.3$ & $91.1$\\
rapeseed & hay, barn, yellow lady's slipper, daisy, stone wall & $3$ & $0.0$ & $95.5$ & $99.3$ & $97.4$\\
nematode & flatworm, jellyfish, lacewing, vine snake, isopod & $3$ & $0.0$ & $93.9$ & $99.3$ & $96.6$\\
tench & barracouta, sturgeon, goldfish, coho, gar & $3$ & $0.4$ & $72.6$ & $97.3$ & $84.9$\\
African elephant & tusker, Indian elephant, zebra, warthog, water buffalo & $3$ & $4.8$ & $57.1$ & $99.2$ & $78.2$\\
\midrule
mean of the twenty & & & $1.4$ & $65.3$ & $98.8$ & $82.1$\\
\midrule
beagle & basset, Walker hound, English foxhound, Pembroke, basenji & $3$ & $6.4$ & $33.5$ & $99.5$ & $66.5$\\
school bus & trolleybus, cab, minibus, traffic light, snowplow & $3$ & $1.2$ & $71.3$ & $98.2$ & $84.7$\\
\bottomrule
\end{tabular}}
\end{table*}

\begin{figure*}[tp]
\centering
\includegraphics[width=.9\textwidth]{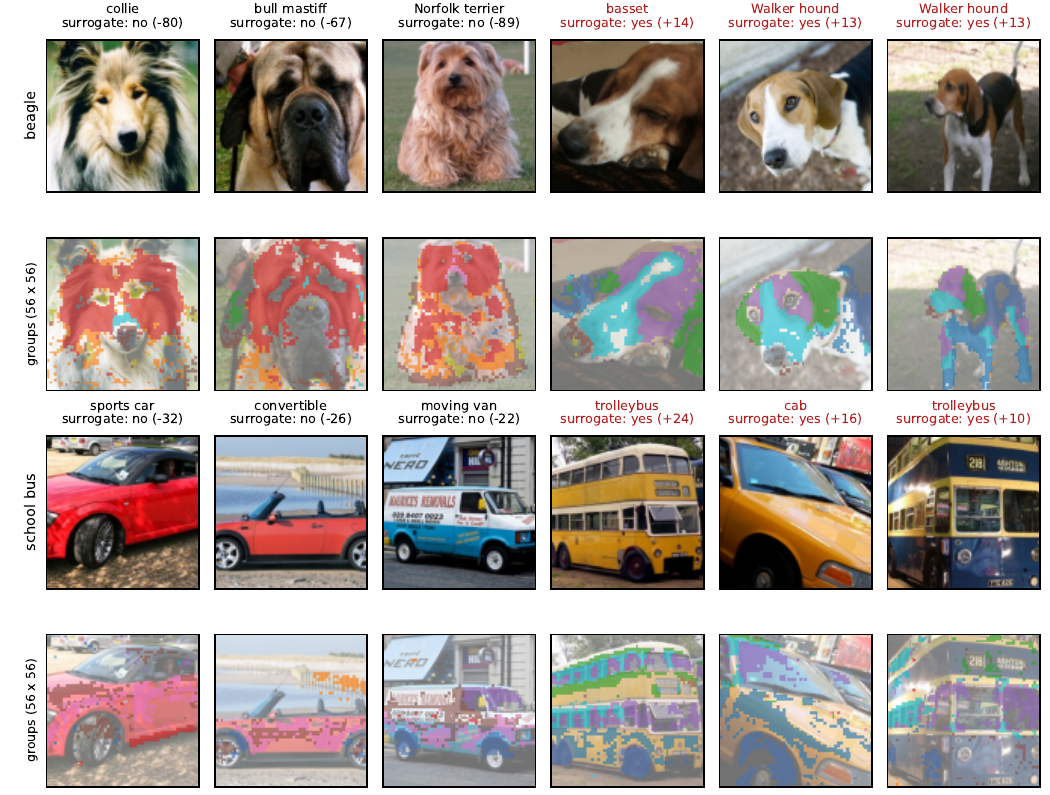}
\caption{Explanations of a ``no'' on teacher-negative sibling-class images, drawn on the native $56\times56$ lattice that the readout reads (TV models of seed $101$; positive groups in cool colors, negative groups in warm colors, discard channels uncolored; colors name the groups of one model only). Left: the three images that the surrogate rejects with the largest share of negative groups; right: the three with the highest surrogate score among those it accepts although the teacher does not ($37$ of $470$ beagle and $7$ of $82$ school-bus sibling negatives for this seed). Number: the surrogate's logit minus its threshold. The images were chosen by these rules only (\texttt{fig\_negatives.py}).}
\label{fig:negatives}
\end{figure*}

\emph{Where the disagreements lie} (\texttt{fidelity\_strata.py}, TV models, three seeds). Of the surrogate's false positives per seed, $34$ of $41$ (beagle) and $10$ of $18$ (school bus) are sibling-class images, and false negatives are rare ($0.3$ and $1.0$ per seed). By the teacher's class margin, the agreement is $99$--$100\%$ for negatives with a margin below $-7$ (most of the test set), $96$--$97\%$ between $-7$ and $-5$, and $51$--$80\%$ between $-5$ and $-3$ ($13$ and $5$ images), and on teacher-positive images with a margin below $3$ it is $89$--$99\%$ ($29$ and $8$ images): the disagreements concentrate where the teacher is close to its decision, as for any surrogate of a decision. On the teacher's own errors against the ImageNet label, its false positives ($21$ and $7$ images of other classes that it calls the class), the surrogate agrees with the teacher on $98\%$ and $91\%$; the test set holds no class image that the teacher misses. \Cref{fig:negatives} shows explanations of a ``no'' that succeed and fail, on the native $56\times56$ lattice that the readout reads. Where the surrogate rejects a sibling image as the teacher does, negative groups cover most of the dog or the car; where it accepts one that the teacher rejects, positive groups take the face of a dog of beagle-like build and coat (basset, Walker hound) or the yellow body of a trolleybus or a cab. The surrogate thus names what it confuses, but whether the teacher rejects these images for the reasons the negative groups give is not tested by this figure (\cref{app:sign}).

\emph{The five nearest classes} (\texttt{sibling\_eval.py}, \texttt{sibling\_summary.py}). The sibling sets above are broad ($118$ dog breeds, $19$ road vehicles). A stricter rule, the same for every class and fixed without looking at any image, takes the five ImageNet classes whose rows of the teacher's head are closest to the row of the class (cosine) and all their $250$ validation images. The teacher calls $0$ to $7\%$ of these images the class ($1.4\%$ on average over the twenty random classes), but the surrogate rejects far fewer of them than of other negatives (\cref{tab:siblings}): $65.3\%$ on average over the twenty random classes (median $68.6\%$, from $29\%$ for paddle to $96\%$ for rapeseed, below $50\%$ for five classes), $33.5\%$ for beagle (basset, Walker hound, English foxhound, Pembroke, basenji) and $71.3\%$ for school bus, against $96$ to $100\%$ of all other test negatives. With the true positive rate of the test set, the balanced accuracy against these classes is $82.1\%$ on average ($65$ to $97\%$); it combines two populations (the sibling images and the positives of the test set), not one naturally sampled set, and since the test set holds only images that the teacher calls the class, explanations of the teacher's false negatives are not evaluated. The classes whose siblings the teacher itself places closest to the class (the share of sibling images within $5$ logits of the class) are those the surrogate rejects least (rank correlation $-0.78$). The fidelity of the main text thus holds between a class and the rest of ImageNet, not between a class and its nearest neighbors: there the surrogate accepts many images that the teacher rejects, and its explanation of such an image (\cref{fig:negatives}) is an explanation of the surrogate's error, not of the teacher's decision. Every class has three seeds in this test.

\begin{table}[tp]
\centering
\small
\caption{True positive rate and true negative rate of the surrogates at the selection threshold (best-fidelity checkpoint, mean over three seeds; standard deviation over seeds for the hard negatives). Hard: teacher-negative test images of sibling classes (dog breeds for beagle, road vehicles for school bus); other: all remaining teacher-negative test images.}
\label{tab:hard}
\fitcolumn{\begin{tabular}{@{}llccccc@{}}
\toprule
task & arm & TPR (\%) & TNR all (\%) & TNR hard (\%) & TNR other (\%) & hard images\\
\midrule
beagle & TV & $99.5$ & $99.01$ & $92.70\pm0.65$ & $99.83$ & $470$\\
 & without TV & $98.4$ & $98.97$ & $92.34\pm2.59$ & $99.83$ & $470$\\
school bus & TV & $98.2$ & $99.56$ & $87.40\pm7.04$ & $99.81$ & $82$\\
 & without TV & $99.4$ & $99.57$ & $84.96\pm4.93$ & $99.87$ & $82$\\
\bottomrule
\end{tabular}}
\end{table}

\section{A chest X-ray classifier}
\label{app:cxr}
AtomEx is not tied to ImageNet. We explain a chest X-ray classifier, Medical\_MAE~\cite{xiao2023delving}, for pleural effusion and edema, and test the groups against the radiologists' masks of CheXlocalize~\cite{saporta2022benchmarking}. In short: the surrogates reach a balanced accuracy of $94$--$95\%$ against the teacher; the groups are anatomical regions whose sign says whether the tissue looks abnormal; for effusion the region of the positive groups overlaps the masks within $0.03$ in IoU of LeGrad, Chefer and Grad-CAM at the same area (Grad-CAM is higher), for edema it is not separated from a centered box while those three methods are above it. A second teacher, BiomedCLIP used zero-shot, gives the same picture with lower fidelity (three seeds).

\paragraph{Teacher.} Medical\_MAE~\cite{xiao2023delving} is a timm ViT-B/16 that its authors pretrained as a masked autoencoder on chest radiographs and fine-tuned for $75$ epochs on CheXpert~\cite{irvin2019chexpert} (the checkpoint of epoch $30$). The decision is a linear head on the layer-normalized mean of the final patch tokens, with no class token, for five labels (cardiomegaly, edema, consolidation, atelectasis, pleural effusion); the input is the grayscale image resized to $256$ pixels (bicubic), center-cropped to $224$ and normalized with mean $0.5056$ and standard deviation $0.252$. On the $413$ frontal CheXlocalize test images its AUROC is $0.956$ for pleural effusion and $0.894$ for edema. We use the released checkpoint as it is, and we explain one output at a time with the teacher's own decision, the logit above zero; at this threshold the head works for effusion and edema (sensitivity $0.87$--$0.90$ and $0.71$--$0.75$), while it is conservative for the other three labels, which we therefore leave out.

\paragraph{Descriptors and training.} As on ImageNet, the descriptor of a site is the layer-normalized input of the last attention block (class token dropped). Four views, windows of the $256$-pixel image shifted by $0$ or $8$ pixels along each axis, are interleaved into a $28\times28$ lattice (a site covers $[8u+4,\,8u+12)$ pixels of the crop). The windows are real windows of the resized image, not mirror-padded, because reflected lung could change the finding. The surrogate has the same form as in \cref{sec:model} (ten used and ten discard channels, channel dropout $0.25$, TV weight $0$ or $0.05$, $40$ epochs, the checkpoint of best fidelity on a patient-disjoint selection part), is fitted to the teacher decisions on $12{,}000$ frontal CheXpert training images ($3{,}919$ teacher-positive of $9{,}999$ for effusion, $2{,}601$ for edema; the remaining $2{,}001$ form the selection part) and is tested on the $413$ CheXlocalize images. The fidelity (three seeds, TV, range over seeds) is $0.943$ ($0.940$--$0.945$) balanced accuracy and $0.992$ AUROC for effusion and $0.949$ ($0.941$--$0.954$) and $0.993$ for edema; without TV $0.940$ and $0.946$. The tables use the $28\times28$ lattice. For seed $101$ we also fitted the $56\times56$ lattice of the ImageNet tables (sixteen views, shifts $-4$ to $8$ pixels); it changes neither the fidelity (balanced accuracy $0.940$ for effusion and $0.937$ for edema against $0.940$ and $0.941$, TV) nor the overlap with the masks (IoU $0.228$ against $0.236$ for effusion and $0.270$ against $0.272$ for edema, TV; without TV $0.237$ and $0.291$ against $0.237$ and $0.300$), so the resolution of the lattice is not what limits the agreement with the radiologists.

\paragraph{Evaluation.} CheXlocalize provides segmentations by radiologists for the images with a finding. We evaluate the images on which the teacher is positive, the mask is not empty and at least half of the mask lies inside the $224$ crop (effusion $87$ of $136$ teacher-positive images, $11\%$ of the mask area lies outside the crop on average; edema $58$ of $99$); a crop pixel belongs to the mask if at least half of its footprint in the original image does. We report precision, recall, IoU and lift (precision divided by the share of the mask in the crop) of the region of the positive groups, and the same for the nulls on the same images: a centered box of the same area, and the mask of another image (mean of $20$ draws). The other methods are compared at equal area: for every image the region of a method is its top pixels, as many as the positive groups of AtomEx (TV) cover in that image, and the numbers are the mean over the three seeds; we add the ROC AUC of the continuous map against the mask (for AtomEx the AUC of a binary region) and bootstrap intervals over images ($1000$ draws).

\paragraph{Attribution methods.} All are written for this teacher, whose decision pools the patch tokens instead of reading the class token, and are checked on random weights (forward values, determinism, class dependence); our pipeline reproduces the stored teacher decisions on every evaluated image. Grad-CAM: gradient of the logit with respect to the final patch tokens, weighted channels, ReLU. Chefer~\cite{chefer2021transformer}: gradient-times-attention relevance propagated through the blocks, read as the mean over all patch rows of the relevance matrix instead of the class-token row. LeGrad~\cite{bousselham2025legrad}: last two blocks, with the block score given by the head of the model applied to the mean of the patch tokens of the block. AttnLRP~\cite{achtibat2024attnlrp}: the rules of the paper on the timm blocks. Integrated gradients: $32$ steps from a black image. The two last methods are less reliable here: the relevance that AttnLRP delivers to the pixels is only $48\%$ (effusion) and $17\%$ (edema) of the logit on average, and the integrated gradients miss the logit by a large gap (completeness), so their low scores say little about the methods.

\begin{table*}[tp]
\centering
\small
\caption{Chest X-ray: overlap of the region of each method with the radiologists' masks (CheXlocalize, teacher-positive images with a mask). The regions of the baselines and of the centered box have the area of the positive groups of AtomEx (TV) in the same image (mean over three seeds; $27.6\%$ and $23.1\%$ of the crop); the AtomEx rows are the regions of the groups at their own area ($27.6$ and $29.7\%$, $23.1$ and $27.1\%$). Lift: precision divided by the share of the mask in the crop ($0.106$ and $0.247$). AUC: threshold-free ROC AUC of the map against the mask. Intervals: bootstrap over images.}
\label{tab:cxr}
\setlength{\tabcolsep}{4pt}
\begin{tabular}{lcccccccc}
\toprule
 & \multicolumn{3}{c}{pleural effusion ($87$ images)} & & \multicolumn{3}{c}{edema ($58$ images)} \\
\cmidrule{2-4}\cmidrule{6-8}
 & IoU [95\% interval] & lift & AUC & & IoU [95\% interval] & lift & AUC\\
\midrule
AtomEx, TV & $.225\ [.190,.259]$ & $2.50$ & $.733$ & & $.251\ [.225,.279]$ & $1.79$ & $.609$\\
AtomEx, no TV & $.224\ [.190,.258]$ & $2.43$ & $.738$ & & $.268\ [.243,.297]$ & $1.71$ & $.619$\\
Grad-CAM & $.251\ [.217,.287]$ & $2.71$ & $.854$ & & $.322\ [.293,.356]$ & $2.13$ & $.756$\\
Chefer & $.210\ [.183,.238]$ & $2.39$ & $.839$ & & $.347\ [.318,.373]$ & $2.25$ & $.832$\\
LeGrad & $.216\ [.187,.247]$ & $2.43$ & $.802$ & & $.321\ [.295,.349]$ & $2.16$ & $.764$\\
AttnLRP & $.113\ [.100,.127]$ & $1.46$ & $.575$ & & $.172\ [.160,.185]$ & $1.33$ & $.536$\\
Integrated gradients & $.119\ [.104,.134]$ & $1.54$ & $.574$ & & $.188\ [.177,.201]$ & $1.44$ & $.549$\\
centered box & $.085\ [.063,.106]$ & $1.07$ & $.478$ & & $.235\ [.210,.261]$ & $1.59$ & $.606$\\
\bottomrule
\end{tabular}
\end{table*}

\paragraph{Result.} \Cref{tab:cxr} gives the numbers. For pleural effusion the regions of AtomEx, LeGrad, Chefer and Grad-CAM overlap the masks to a similar degree (IoU $0.21$--$0.25$) and $2.5$ to $3$ times as much as a centered box; AttnLRP and the integrated gradients are lower. Paired per image and seed (the difference of the IoU on the same image at the same area, with a bootstrap over images, \texttt{cxr\_paired.py}), Grad-CAM is above AtomEx by $0.026$ $[0.012, 0.040]$, Chefer below it by $0.020$ $[0.002, 0.037]$, and the difference to LeGrad, $-0.009$ $[-0.023, 0.004]$, is not separated from zero, which does not show that the two are equal. For edema Chefer, LeGrad and Grad-CAM are above AtomEx by $0.07$ to $0.09$ (every interval above zero), and AtomEx ($0.251$) is not separated from a centered box of the same area ($0.235$; difference $0.015$ $[-0.015, 0.046]$). The differences are computed from the stored maps, which reproduce the table to within $0.01$ (Chefer, whose stored map has ties). The threshold-free AUC says the same, and AtomEx, as a binary region, has the lowest AUC of the methods that work. The three seeds of AtomEx differ by up to $0.03$ in IoU for effusion ($0.208$--$0.236$) and by up to $0.06$ for edema ($0.214$--$0.272$). Without TV the numbers are the same within this spread.

\begin{table}[tp]
\centering
\small
\caption{Chest X-ray with BiomedCLIP as the teacher (mean of three seeds, $28\times28$ lattice): balanced accuracy of the surrogate on the test images, and overlap (IoU) of the positive groups with the masks for $73$ effusion and $59$ edema images. The nulls are a centered box with the area of the positive groups in the same image and the mask of another image (mean of $20$ draws). Lift: precision divided by the share of the mask in the crop.}
\label{tab:cxr-bm}
\setlength{\tabcolsep}{2pt}
\begin{tabular}{lccccc}
\toprule
 & bal.\ acc. & IoU & centered box & other image & lift\\
\midrule
effusion, TV & $.89$ & $.165$ & $.026$ & $.069$ & $3.73$\\
effusion, no TV & $.88$ & $.178$ & $.053$ & $.089$ & $2.71$\\
edema, TV & $.88$ & $.185$ & $.150$ & $.130$ & $1.85$\\
edema, no TV & $.88$ & $.226$ & $.226$ & $.170$ & $1.63$\\
\bottomrule
\end{tabular}
\end{table}

\paragraph{A second teacher.} To see whether the picture depends on Medical\_MAE, we repeated the experiment with BiomedCLIP~\cite{zhang2025biomedclip}, the released ViT-B/16 image tower trained on image--text pairs from the literature, used zero-shot. Its decision follows the example of the model card and was fixed before any evaluation: the logit of a finding is the scaled difference between the cosine similarity of the image embedding (from the class token) to the text ``this is a photo of <finding> chest x-ray'' and to ``this is a photo of normal chest x-ray'', and the finding is positive when the logit is above zero. Nothing is trained or tuned. On the $413$ CheXlocalize images the AUROC is about $0.84$ for effusion and $0.76$ for edema, against $0.956$ and $0.894$ of Medical\_MAE. At logit zero the effusion decision works (sensitivity $0.72$, specificity $0.80$), but the edema decision calls $82\%$ of the test images positive (sensitivity $0.99$, specificity $0.22$); for edema we therefore moved the threshold to $2.65$, the Youden point on the $202$ CheXpert validation images (the CheXlocalize images do not enter; on the test images the sensitivity is $0.77$, the specificity $0.57$ and the AUROC is unchanged). This deviates from the zero-threshold rule, and the edema teacher remains weak. We fitted the surrogates as above (same $28\times28$ lattice, same $12{,}000$ training images, TV weight $0$ or $0.05$), for three seeds ($101$, $211$, $307$).

The surrogates agree less with this teacher than with Medical\_MAE: balanced accuracy $0.89$ (TV) and $0.88$ (without TV) for effusion and $0.88$ for edema, against $0.94$--$0.95$, probably because the teacher decides from the class token, which the surrogate does not see. \Cref{tab:cxr-bm} gives the overlap with the masks. For effusion the positive groups are small ($6.6\%$ of the crop) and selective: precision $0.42$, lift $3.7$, IoU $0.165$ against $0.026$ for a centered box of the same area and $0.069$ for the mask of another image; without TV the region is larger ($13\%$) and the IoU is $0.178$ against $0.053$ and $0.089$. For edema the IoU is $0.185$ with TV, $0.035$ above the centered box ($0.150$), and $0.226$ without TV, not above it ($0.226$), as with Medical\_MAE. The two teachers thus give the same picture, effusion yes and edema no, but this check is weaker than the first: three seeds (the IoU of the seeds lies within $0.01$ for effusion and spreads by $0.04$--$0.06$ for edema), $73$ and $59$ images, no attribution baselines for this teacher, and an edema decision that is weak to begin with.

\paragraph{What the groups are.} \Cref{fig:cxr-groups} shows the used groups of one surrogate of each task. Chest radiographs are taken in a standard pose, so the tokens carry the position of the tissue strongly and the groups are regions of the thorax with an appearance, not objects: for effusion the positive groups are the lateral lung zones (half of the pixels of the first one lie in a radiologist's mask), the region below the lungs and the zone behind the heart; the seven negative groups are normal-looking lung in different places, each at about $4$--$5\%$ of the image, present in most images and mostly in teacher-negative ones. For edema the positive groups are the right mid and lower zone ($74\%$ of its pixels in a mask), the left costophrenic angle and the zone beside the heart. A negative group may also lie inside the masks (the left lower zone, $75\%$): what makes a group evidence against is where and when it occurs, not where the radiologists mark. The positive groups of effusion also take the diaphragm and the cardiac shadow (\cref{fig:cxr-compare}), so the precision against the masks is $0.26$ (lift $2.5$): they mark the region where fluid sits, not the fluid alone.

\paragraph{Two one-sided cards.} On ImageNet a class is set against the rest of the world, and the card of \cref{eq:card} counts the evidence for it. For a finding on a chest radiograph the other side has a meaning of its own: the negative groups are normal-looking lung, and the absence of a finding can be read from normal tissue as well as its presence from abnormal tissue. Both may matter to a radiologist, who has to know what distinguishes the signs of a disease from the signs of normality. We therefore fit both one-sided cards for the chest X-ray surrogates (\cref{tab:cxr-cards}): the card of the finding (\cref{eq:card}) and the card of normality, which reports the finding unless the points of the normal-looking groups reach a threshold. For Medical\_MAE and effusion the two are equally faithful ($93.7$ and $93.9\%$): three or four groups of the zones where fluid sits (effusion if their points reach $4$) against six or seven groups of normal lung (effusion unless their points reach $6$ to $8$). For edema the card of normality is more faithful ($94.5$ against $92.0\%$) and as faithful as the two-sided card: the teacher's edema is recognized better by the absence of normal lung than by the presence of one abnormal region. For BiomedCLIP the card of the finding is better for effusion and the card of normality for edema. All cards lose $0.4$ to $4.8$ points against the surrogates, more than on ImageNet.

\begin{figure}[!htbp]
\centering
\includegraphics[width=\columnwidth]{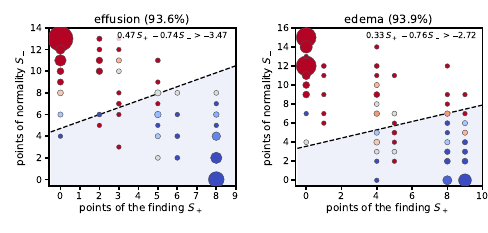}
\caption{Chest X-ray, the two one-sided cards read together (Medical\_MAE, seed $101$, TV): every test image is a point ($S_+$, $S_-$) of the points of the card of the finding and of the card of normality; a disk per occupied point, its area growing with the number of images, its color the share of the images that the teacher calls positive (blue: positive, red: negative). Below the dashed line $aS_+-bS_-=c$ (shaded) the combination reports the finding; the balanced fidelity of the combination is in the title (\texttt{analysis/card\_combine.py}).}
\label{fig:cxr-combined}
\end{figure}

\begin{figure*}[tp]
\centering
\includegraphics[width=.44\textwidth]{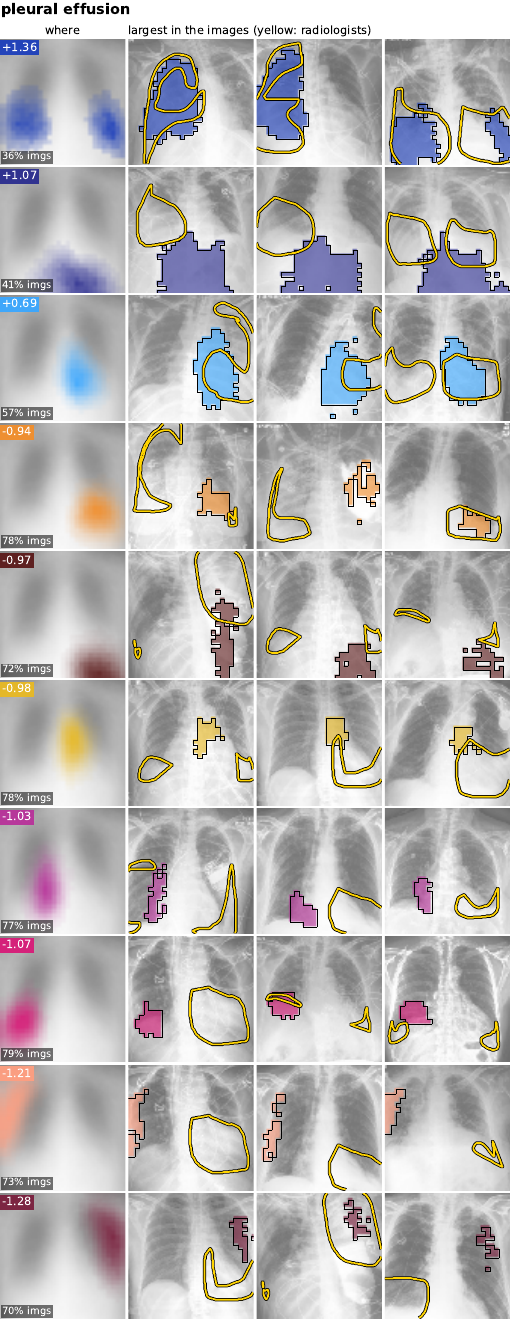}\hspace{.04\textwidth}%
\includegraphics[width=.44\textwidth]{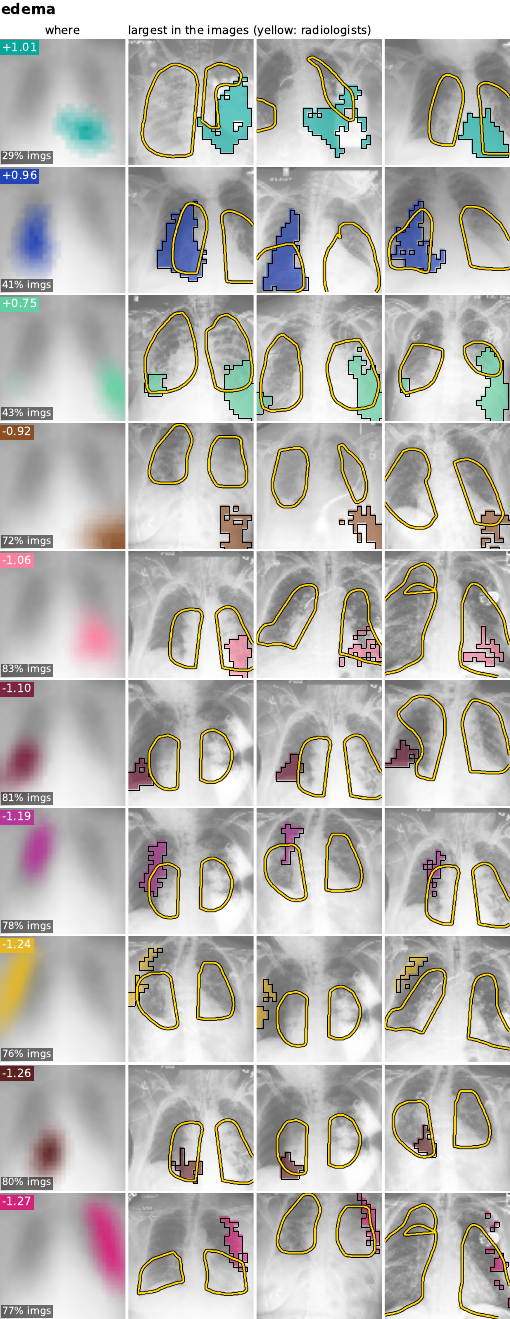}
\caption{The used groups of the chest X-ray surrogates (seed $101$, TV, $28\times28$ lattice), pleural effusion left and edema right, one group per row, positive groups first, each in its own color as in \cref{fig:teaser} (cool: positive coefficient, warm: negative; the label gives the coefficient). First column: the share of the $413$ test images in which a crop pixel belongs to the group (opacity), over the mean of the evaluated X-rays, and the share of images in which the group appears; next three: the evaluated images in which the group is largest, with the group (black outline) and the radiologists' mask (yellow). Left and right in the text are the patient's.}
\label{fig:cxr-groups}
\end{figure*}

\begin{figure*}[tp]
\centering
\includegraphics[width=\textwidth]{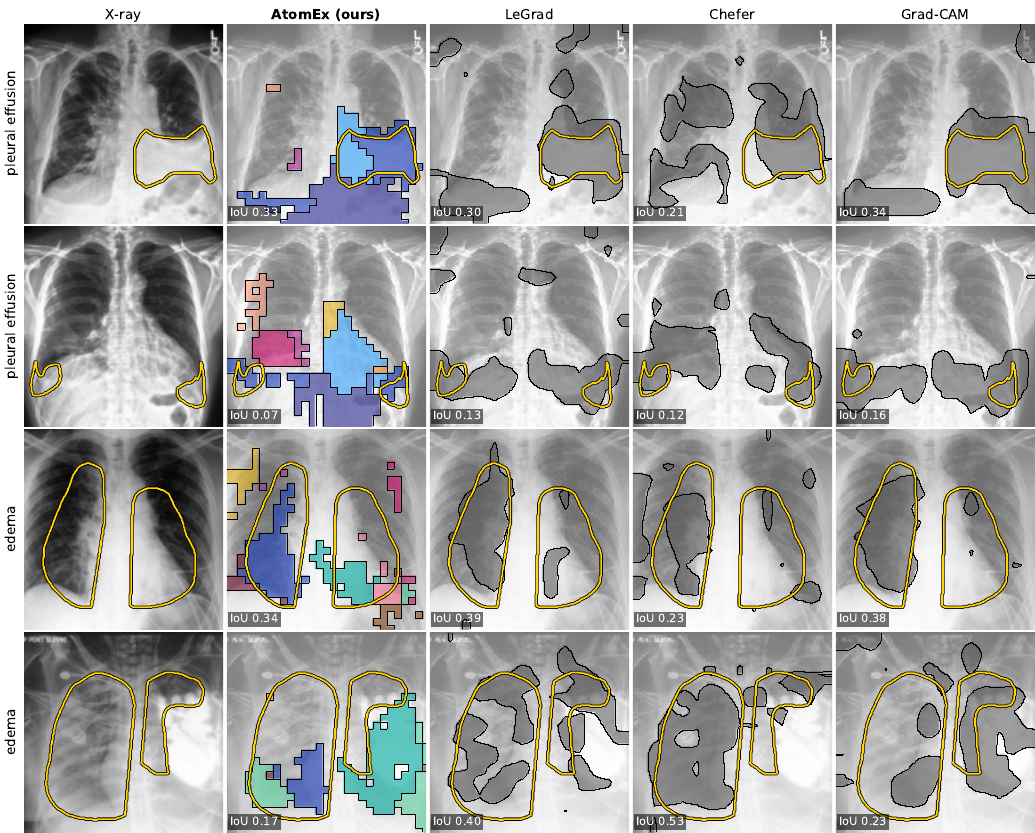}
\caption{Chest X-ray, two evaluated images per task (the $75$th and the $25$th percentile of the IoU of AtomEx, seed $101$): the radiologists' mask (yellow outline), the used groups of AtomEx drawn as in \cref{fig:gallery-56} (cool colors: positive coefficient, warm: negative; the IoU is that of the positive groups) and the regions of LeGrad, Chefer and Grad-CAM (gray) at the area of the positive groups of AtomEx, with the IoU of each region.}
\label{fig:cxr-compare}
\end{figure*}

The two cards can also be read together. Let $S_+$ be the points of the card of the finding and $S_-$ those of the card of normality on an image; a linear decision $aS_+-bS_->c$ with two real weights, fitted by a class-balanced logistic regression on the fit images ($c$ on the selection images), uses the evidence of both sides while each card stays readable on its own (\cref{fig:cxr-combined}). One point of normality weighs $1.2$ to $2.3$ points of the finding for Medical\_MAE and $0.9$ to $2.2$ for BiomedCLIP. The combined rule is not an integer card: its two real weights trade some of the simplicity of a card for fidelity. It is the most faithful for BiomedCLIP ($88.0$ and $86.0\%$, against $87.2$ and $85.5\%$ for the two-sided card) and close to the better one-sided card for Medical\_MAE ($93.7$ and $94.3\%$); over all $24$ chest X-ray surrogates, with and without TV, it reaches $90.7\%$, against $90.2\%$ for the two-sided card and $91.3\%$ for the surrogates. On ImageNet, where the negative groups are general (\cref{app:sides}), it is about as faithful as the two-sided card ($99.24$ and $98.73\%$ on the main and the random classes, against $99.32$ and $98.75\%$).

\begin{table}[!htbp]
\centering
\small
\caption{Chest X-ray: balanced fidelity to the teacher (\%, $413$ test images, TV, mean over three seeds, $28\times28$ lattice) of the surrogate and of the integer cards read from its groups: two-sided, the card of the finding (\cref{eq:card}, the groups with $\gamma_k>0$), the card of normality (the groups with $\gamma_k<0$: the finding unless their points reach a threshold) and the two one-sided cards combined by $aS_+-bS_->c$ (\cref{fig:cxr-combined}). Areas from the stored descriptors (\texttt{analysis/card\_areas\_cxr.py}).}
\label{tab:cxr-cards}
\fitcolumn{\begin{tabular}{@{}llccccc@{}}
\toprule
teacher & finding & surrogate & two-sided & card of the finding & card of normality & combined\\
\midrule
Medical\_MAE & effusion & 94.28 & 92.76 & 93.74 & 93.93 & 93.74\\
 & edema & 94.93 & 94.45 & 92.04 & 94.53 & 94.28\\
BiomedCLIP & effusion & 89.11 & 87.18 & 86.38 & 84.28 & 88.02\\
 & edema & 87.64 & 85.46 & 83.52 & 84.32 & 85.95\\
\bottomrule
\end{tabular}}
\end{table}

\section{A scorecard read from the groups}
\label{app:card}
\Cref{sec:card} reads a scorecard (\cref{eq:card}) from the frozen groups of a trained surrogate. The card is fitted on the fit images of the surrogate, with the two classes balanced, as $b+\sum_k p_k\mathbf 1[a_k>m_k]>\theta$ over the groups with $\gamma_k>0$ and $p_k\ge0$; then $T$ is the smallest integer above $\theta-b$. For fixed thresholds it is a logistic regression on the binary features $\mathbf 1[a_k>m_k]$ with nonnegative points, a convex problem; the thresholds are found by coordinate descent over a grid of candidates (quantiles of the nonzero areas of each group and $0.25$, $0.5$, $1$, $2$ and $4\%$ of the image), starting from a fit of the same rule with sigmoids in place of the steps; while more than eight groups carry points, the group whose removal costs the least selection fidelity is dropped and the card refitted (for the card of \cref{eq:card} this never happens: our surrogates have three to six groups with $\gamma_k>0$; the limit matters for the two-sided card of \cref{app:sides}). The threshold $\theta$ is chosen on the selection images, and the points and $b$ are rounded to integers, which costs $0.01$ points of fidelity on average over the nine models ($98.96$ against $98.95\%$; \texttt{analysis/card\_areas.py}, \texttt{card\_fit.py --sides positive}). The coordinate descent finds better cards than the gradient fit alone (selection fidelity $98.0$ to $98.5$ against $96.7$ to $97.9\%$ on beagle) and is chosen on the selection images for eight of the nine models.

\paragraph{The executable rule.} The fit proceeds in a fixed order (\texttt{card\_fit.py}). (i)~A gradient fit of the same rule with sigmoids $\sigma((a_k-m_k)/t)$ in place of the steps: points started at $4$, thresholds at $2\%$, $b$ at $0$; Adam with learning rate $0.01$, $1{,}500$ steps at $t=0.0125$ and $1{,}500$ steps annealing $t$ to $0.00125$; class-balanced logistic loss on the fit images. (ii)~Coordinate descent on the thresholds, three sweeps: the candidates of a group are the $2$nd to $98$th percentiles ($40$ levels) of its nonzero areas on the fit images, each lowered by $10^{-7}$, and $0.25$, $0.5$, $1$, $2$ and $4\%$ of the lattice; for every candidate the points are refitted exactly (class-balanced logistic loss with an $L_2$ penalty of $10^{-4}$, points constrained to $p_k\ge0$, L-BFGS-B). (iii)~Of the two fits, the one with the higher balanced fidelity on the selection images is kept. (iv)~The points and $b$ are rounded to the nearest integer. (v)~Only then is $\theta$ chosen on the selection images, as the cut between the integer scores with the highest balanced fidelity (the lowest such cut on ties). The card then reads $\sum_k p_k\mathbf 1[a_k>m_k]\ge T$ with $T$ the smallest integer above $\theta-b$. No constraint forces $T\ge1$; every fitted card has it ($102$ cards on ImageNet and the chest X-rays), so none accepts an image on which no group of the class counts. An area is a number of lattice sites out of $3{,}136$, and every threshold lies just below an attained area or at a fixed share, so a condition reads ``at least $j_k$ sites'', with $j_k$ the smallest number of sites above $m_k$; the card figures show $j_k/3{,}136$ with two decimals and the sign $\ge$ (one site is $0.032\%$), which makes every comparison checkable (for the black dog of \cref{fig:card-beagle} the back covers exactly the $40$ sites it needs). The card is evaluated on these native lattice areas; the dense maps of the figures give the same conditions on $99.3\%$ of the $594$ conditions of $99$ school-bus test images (teacher-positive images and sibling-class negatives) and the same decision on all of them (\texttt{card\_dense\_native.py}).

\section{Two-sided and negative cards}
\label{app:sides}
Besides the card of \cref{eq:card}, which counts only the evidence for the class, the same procedure fits two other cards on the same frozen groups: a \emph{two-sided} card $b+\sum_k p_k\mathbf 1[a_k>m_k]>\theta$ over all used groups, with points of the sign of $\gamma_k$, and a \emph{negative} card over the groups with $\gamma_k<0$ alone, which accepts an image unless the points against the class reach a threshold. \Cref{tab:sides} compares the three. The two-sided card matches the CNN ($99.32\%$ against $99.27\%$ on the main tasks, $98.75\%$ against $98.92\%$ on the random classes); eight of the nine main cards keep eight groups and one keeps seven, with point magnitudes $1$--$4$, and $54$ of the $60$ random-class cards keep eight groups. The card of \cref{eq:card} gives up $0.37$ and $0.33$ points for counting only the evidence for the class; on school bus it is more faithful than the two-sided card. The negative card is the least faithful on beagle and school bus, and close to the others on valley, where evidence against a landscape (objects in front of it) is as informative as evidence for it. With a rule fixed before the results (the card of \cref{eq:card}, unless the negative card is better on the selection images by more than $0.5$ points), the card of \cref{eq:card} is chosen for $8$ of the $9$ main models and $59$ of the $60$ random-class models. A second rule, also fixed on the selection images, replaces the one-sided card by the two-sided card where the latter is more faithful there by more than $0.5$ points (\cref{app:stress}).

\begin{table}[!htbp]
\centering
\small
\caption{Balanced fidelity to the teacher (\%, test images, mean over three seeds) of the surrogates and of three integer cards read from their groups: two-sided (all used groups, points of the sign of $\gamma_k$), positive (\cref{eq:card}, the groups with $\gamma_k>0$) and negative (the groups with $\gamma_k<0$). 20 random: mean over the $60$ models of \cref{app:random}.}
\label{tab:sides}
\fitcolumn{\begin{tabular}{@{}llcccc@{}}
\toprule
task & arm & surrogate & two-sided & positive (\cref{eq:card}) & negative\\
\midrule
beagle & $\rho=0.05$ & 99.24 & 99.34 & 98.60 & 97.57\\
school bus & $\rho=0.05$ & 98.89 & 99.00 & 99.57 & 95.28\\
valley & $\rho=0.05$ & 99.67 & 99.62 & 98.67 & 99.13\\
\midrule
mean & $\rho=0.05$ & 99.27 & 99.32 & 98.95 & 97.33\\
 & $\rho=0$ & 99.18 & 98.95 & 98.85 & 98.59\\
20 random & $\rho=0.05$ & 98.92 & 98.75 & 98.42 & 95.70\\
\bottomrule
\end{tabular}}
\end{table}

\begin{figure}[!htbp]
\centering
\includegraphics[width=\columnwidth]{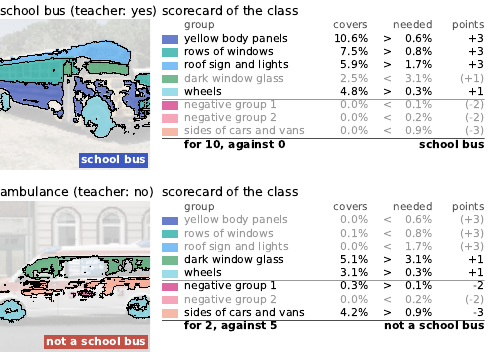}
\caption{The two-sided card of the school-bus surrogate of \cref{fig:teaser} (seed $101$) on the images of \cref{fig:teaser}. The bus earns $10$ points and none against; the ambulance shares bus-like wheels and dark glass ($+1$ each; the thresholds of this card differ from those of \cref{fig:teaser}), but two groups with $\gamma_k<0$ subtract $5$. Here $b=\theta$, so the class is accepted when the points for it exceed those against it.}
\label{fig:card-both}
\end{figure}

\begin{figure}[!htbp]
\centering
\includegraphics[width=\columnwidth]{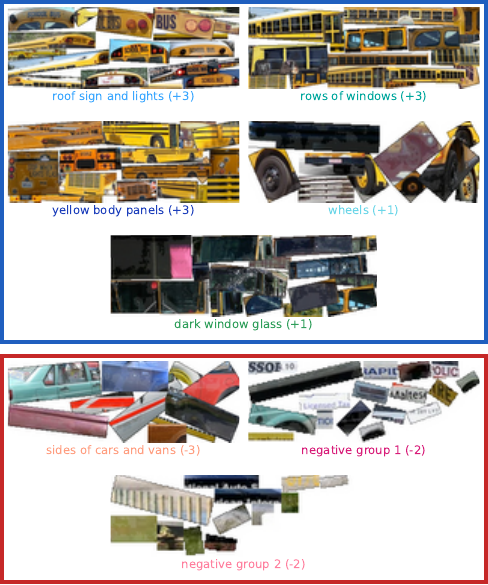}
\caption{The eight groups of the two-sided school-bus card of \cref{fig:card-both} (blue: they add points, red: they subtract them), drawn as in \cref{fig:collage}; the red groups take their pieces from the first $20$ images of other vehicles, and negative groups 1 and 2, with few pieces there, up to six pieces per image from all $50$ images of other vehicles (\cref{app:collage}).}
\label{fig:collage-both}
\end{figure}

The two-sided card makes rejection explicit (\cref{fig:card-both}), but its negative groups are rarely specific counterexamples: they count on $31$--$97\%$ of all test images, five of the nine main cards accept an image on which no group counts ($b>\theta$), and the red sheets of \cref{fig:collage-both} collect the sides of cars and vans and fragments of lettering rather than one recognizable object. Without TV the two-sided cards lose $0.2$ points against their surrogates ($98.95$ against $99.18\%$), need points up to $9$, and eight of nine accept an image on which no group counts.

\section{Stress tests of the final card}
\label{app:stress}
The card is fitted separately from the surrogate, so the hard cases studied for the surrogate are repeated here for the card itself, on the same images and edits (the analyses of \cref{app:hard,app:sign} and of the stability test of \cref{app:compare}, rerun with the areas of the groups saved; \texttt{card\_stress.py}).

\paragraph{When to use the two-sided card.} The card of \cref{eq:card} counts only the evidence for the class, and this has a price on images of similar classes. We therefore fix one rule on the selection images: the two-sided card of \cref{app:sides} replaces the one-sided card when it is more faithful on the selection images by more than $0.5$ points (\texttt{card\_side\_rule.py}). The rule picks the two-sided card for $5$ of the $9$ main models (the three beagle models and two of valley) and for $11$ of the $60$ random-class models, and the chosen cards reach $99.49\%$ and $98.61\%$ balanced fidelity on the test images, against $98.95\%$ and $98.42\%$ for the one-sided cards and $99.27\%$ and $98.92\%$ for the surrogates. Where a rejection has to be explained as well as an acceptance, as for medical findings, we report the two-sided card or both one-sided cards regardless of the rule (\cref{app:cxr}).

\begin{table*}[tp]
\centering
\small
\caption{The final rules on the five classes nearest to the target in the teacher's head (the protocol of \cref{tab:siblings}: their $250$ validation images, mean over the three seeds and, for the random classes, over the classes): balanced fidelity, the mean of the TNR on the sibling images the teacher rejects (in parentheses) and the TPR of the rule on its own test set. Chosen: the one-sided or the two-sided card by the selection rule above. Logistic: the regression on all ten areas of \cref{tab:card}. Scoring system and tree: the rules of \cref{tab:complexity} with six conditions.}
\label{tab:stress-siblings}
\begin{tabular}{@{}lccccccc@{}}
\toprule
 & surrogate & card (\cref{eq:card}) & two-sided & chosen & logistic & scoring system ($6$) & tree ($6$)\\
\midrule
20 random classes & $82.1$ ($65.3$) & $80.3$ ($62.3$) & $80.8$ ($62.9$) & $80.6$ & $79.7$ ($60.2$) & $80.1$ ($62.2$) & $84.1$ ($72.0$)\\
beagle & $66.5$ ($33.5$) & $57.8$ ($15.7$) & $65.7$ ($31.3$) & $65.7$ & $64.3$ ($29.2$) & $59.4$ ($19.4$) & $73.7$ ($49.4$)\\
school bus & $84.7$ ($71.3$) & $81.4$ ($62.9$) & $82.4$ ($66.0$) & $81.4$ & $81.2$ ($63.0$) & $83.5$ ($67.1$) & $85.7$ ($74.9$)\\
\bottomrule
\end{tabular}
\end{table*}

\begin{figure}[!htbp]
\centering
\includegraphics[width=\columnwidth]{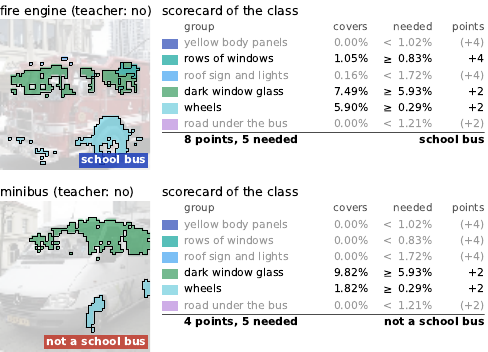}
\caption{A failure of the one-sided card (school bus, seed $101$, native $56\times56$ groups): a fire engine that the teacher and the surrogate reject earns $8$ points with windows, dark glass and wheels and is accepted; a minibus with dark glass and wheels earns $4$ and is rejected.}
\label{fig:card-hard}
\end{figure}

\paragraph{Nearest classes.} On the sibling images (\cref{tab:stress-siblings}) the one-sided card rejects fewer images than the surrogate: $62.3\%$ against $65.3\%$ on the random classes, and on beagle only $15.7\%$ against $33.5\%$, since the parts that the card counts also occur on other hounds and nothing on the card can take points away. The two-sided card recovers most of the gap ($31.3\%$ on beagle), because its negative groups, although they are general (\cref{app:sides}), are present on the other breeds. The same holds outside the head-nearest classes: of the $50$ road-vehicle negatives of \cref{app:collage} the one-sided school-bus card of seed $101$ accepts $12$ and the surrogate $3$; \cref{fig:card-hard} shows a fire engine that earns $8$ points with windows, dark glass and wheels. A small decision tree with six splits on the same areas rejects more sibling images than any card ($72.0\%$), although it is less faithful on the test set (\cref{tab:complexity}). The card and the surrogate agree on $76$ to $79\%$ of the teacher-negative sibling images.

\paragraph{Edits.} Under the deletions of \cref{app:sign} the card behaves like the surrogate. When a positive group is deleted from a class image, the teacher changes its decision on $8.1\%$ and $6.5\%$ of the edits (beagle, school bus); the card follows $2.4\%$ and $10.1\%$ of these changes, the surrogate $2.4\%$ and $3.8\%$, and the card never changes its decision when the same shape is deleted elsewhere. Where the card's points change, they move in the direction of the teacher's margin on $70\%$ and $76\%$ of the deletions. Pasting a positive group into a negative image changes the card's decision more often than the surrogate's ($2.5\%$ and $14.1\%$ of the insertions, against $0.2\%$ and $4.5\%$), while the teacher almost never changes ($0$ and $0.2\%$): the card takes a pasted part of a bus as evidence for a bus.

\paragraph{Natural perturbations.} On the teacher-positive images of the stability test (noise, brightness, JPEG and shifts of $2$ and $4$ pixels), whenever the teacher keeps its decision the card keeps it too, for every perturbation and seed. Between $0.3\%$ and $3.5\%$ of the conditions of the card change (most under JPEG and noise), but a class image has a median of $6$ (beagle) and $7$ (school bus) points more than it needs, and the closest threshold lies a median of $25$ and about $12$ lattice sites away.

\paragraph{Uncertainty of the headline numbers.} Paired bootstrap intervals for the difference of balanced fidelity between the one-sided card and the surrogate (\texttt{card\_uncertainty.py}; one resampling of the shared test images for the nine main models, and resampling of the classes and, within a class, of its test images, stratified by the teacher's decision, for the random classes): $-0.32$ points $[-0.76, +0.15]$ on the main tasks and $-0.50$ $[-0.88, -0.13]$ on the random classes. The two-sided card is within $+0.06$ $[-0.15, +0.35]$ and $-0.17$ $[-0.41, +0.08]$ of the surrogate. The one-sided card thus costs about half a point on the random classes, a difference that the intervals separate from zero. On the test images it agrees with the surrogate on $99.0\%$ of the teacher-positive and $98.4\%$ of the teacher-negative images (main tasks).

\begin{table}[!htbp]
\centering
\small
\caption{Fidelity against complexity of interpretable rules on the same frozen areas (balanced test fidelity, \%, mean over seeds and classes). Conditions: the number of area thresholds a rule reads. Card: \cref{eq:card} with at most $k$ groups, integer points. Scoring system: a conventional sparse scoring system, $L_1$-regularized logistic regression on threshold indicators of all ten groups (deciles of the nonzero areas and $0.25$ to $4\%$ of the lattice), cut to $k$ indicators, refitted, scaled to a largest weight of $3$, $5$ or $10$ points and rounded, signs free. Threshold model: the same with real weights. Tree: a class-balanced decision tree with $k$ splits. Fit, selection and test images as for the card (\texttt{card\_complexity.py}).}
\label{tab:complexity}
\fitcolumn{\begin{tabular}{@{}llcccccc@{}}
\toprule
 & conditions & $1$ & $2$ & $3$ & $4$ & $5$ & $6$\\
\midrule
main tasks & card (\cref{eq:card}) & $95.98$ & $98.12$ & $98.67$ & $98.75$ & $98.82$ & $98.95$\\
 & scoring system & $97.73$ & $98.03$ & $98.56$ & $98.65$ & $98.94$ & $99.11$\\
 & threshold model & $97.73$ & $98.03$ & $98.56$ & $98.65$ & $98.93$ & $98.89$\\
 & tree & $97.61$ & $98.43$ & $98.52$ & $98.18$ & $98.22$ & $98.26$\\
\midrule
20 random & card (\cref{eq:card}) & $97.59$ & $98.16$ & $98.24$ & $98.31$ & $98.45$ & $98.42$\\
 & scoring system & $97.45$ & $97.51$ & $97.98$ & $98.32$ & $98.33$ & $98.36$\\
 & threshold model & $97.45$ & $97.49$ & $97.96$ & $98.32$ & $98.33$ & $98.38$\\
 & tree & $97.38$ & $97.80$ & $97.84$ & $97.85$ & $97.72$ & $97.56$\\
\bottomrule
\end{tabular}}
\end{table}

\paragraph{Simpler rules on the same areas.} \Cref{tab:complexity} compares the card with other interpretable rules that read the same frozen areas and the same number of conditions. With four to six conditions a conventional sparse scoring system and a threshold model with real weights are as faithful as the card, although they may use the groups of both signs and weights of any sign, and a decision tree is less faithful. The fit of \cref{eq:card} therefore does not buy fidelity over standard rule learners; what it adds is the form of the rule, evidence for the class with a few small integer points. On the nearest classes the tree is the most specific (\cref{tab:stress-siblings}), which shows what a rule that can test for the absence of a group gains on hard negatives.

\section{Why the card is not trained directly}
\label{app:card-routes}
\begin{table}[!htbp]
\centering
\small
\caption{Routes to a card (beagle / school bus, three seeds, best-fidelity checkpoint, $\rho=0.05$, $p=0.25$). Below the line the alternatives, trained on a store of the two tasks (the rerun of the CNN surrogate is the same protocol on that store); the readout of the groups is then $b+\sum_k\gamma_kw_k[\sigma((a_k-m_k)/t_k)-\sigma(-m_k/t_k)]$ (one sigmoid per group of learned width $t_k$), its step is the card, or the step is trained directly with a temperature annealed to zero. Fidelity: of the card (for the one-sigmoid readout first the continuous model); area: used area; box prec.: precision of the positive groups against the boxes.}
\label{tab:card-routes}
\fitcolumn{\begin{tabular}{@{}lccc@{}}
\toprule
route to a card & fidelity (\%) & area (\%) & box prec.\ (\%)\\
\midrule
two-sided card read from the CNN groups (\cref{tab:sides}) & 99.34 / 99.00 & 22.1 / 18.2 & 95.0 / 94.7\\
\midrule
monotone CNN surrogate (rerun) & 99.01 / 98.60 & 26.1 / 16.2 & 95.8 / 94.6\\
one sigmoid per group & 99.31 / 99.74 & 23.4 / 18.1 & 97.6 / 96.8\\
\quad its step, same $b$ and $\theta$ & 98.85 / 99.12 & 23.4 / 18.1 & 97.6 / 96.8\\
\quad start fixed at $-4$, its step & 97.14 / 99.28 & 19.4 / 15.7 & 88.7 / 93.0\\
\quad fine-tuned, random temperature, its step & 98.78 / 99.12 & 21.4 / 15.5 & 98.2 / 97.1\\
step trained directly, threshold from $0.05$ & 98.14 / 98.76 & 32.3 / 25.6 & 91.3 / 95.7\\
step trained directly, thresholds from $0.02$, $0.1$ & 99.11 / 99.41 & 26.5 / 21.3 & 93.1 / 96.6\\
\bottomrule
\end{tabular}}
\end{table}

\begin{figure*}[tp]
\centering
\includegraphics[width=\textwidth]{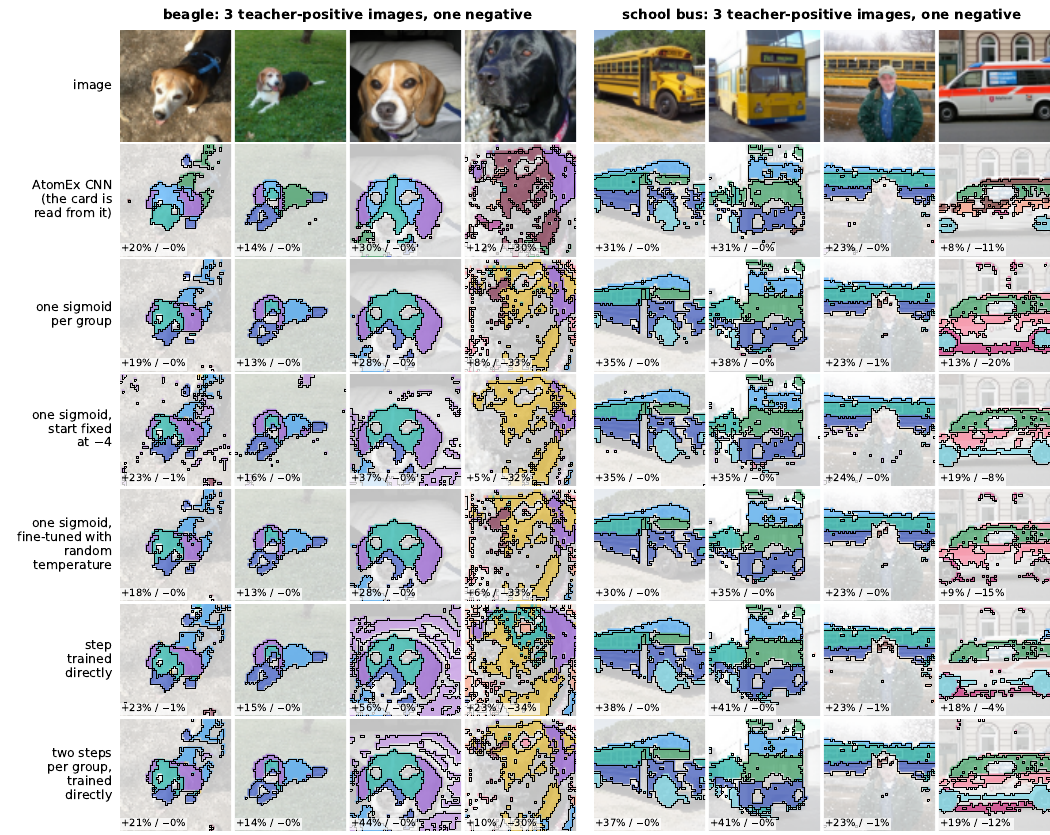}
\caption{The groups of the routes of \cref{tab:card-routes} on the same test images (seed $101$, hard $56\times56$ maps; cool: positive groups, warm: negative groups; colors mark the sign only, and the groups of different models are unrelated). Numbers: used area of the positive / negative groups. First row: the surrogate from which the cards of \cref{tab:card,tab:sides} are read. The negatives are the black dog of \cref{fig:gallery} and the ambulance of \cref{fig:teaser}. Trained directly, the step spreads the positive groups over the background (beagle close-up: $56\%$ of the image against $30\%$) and onto the ambulance; one sigmoid per group covers the ambulance with negative groups that count on most test images; with the start fixed at $-4$ the positive groups grow; fine-tuning with a random temperature removes the wheel group of the bus (\texttt{appendix\_figs/fig\_card\_routes.py}).}
\label{fig:card-routes}
\end{figure*}

\Cref{tab:card-routes,fig:card-routes} compare the card read from the CNN groups with routes that learn a card more directly; since these routes train signed readouts, the comparison uses the two-sided card of \cref{app:sides}. All of them reach a fidelity close to that of the CNN, but their groups are worse, and the groups are the explanation. Training the step directly fails: as the temperature falls, the gradient of a threshold vanishes except for images whose area lies near it, so the thresholds stay near their start and the groups grow to pass them ($32$ and $26\%$ of the image against $22$ and $18\%$ for the card, box precision $91$ and $96\%$). In \cref{fig:card-routes} the positive groups spread over the pillow of the beagle close-up ($56\%$ of the image against $30\%$) and cover the wheels and the lower body of the ambulance ($18\%$ against $8\%$); two steps per group, with thresholds started at $0.02$ and $0.1$, give a better card but the same growth ($44\%$ on the close-up). A readout of one sigmoid per group, trained in its continuous form and read as a step with the same start value and threshold, gives a card of about the same fidelity and precise positive groups, but its negative groups are catch-alls: the two that count on the ambulance of \cref{fig:teaser} count on $83$ and $61\%$ of all test images and cover $20\%$ of the ambulance, one of them along the lower edge of the van. Fixing the start value at $-4$, so that an image without positive evidence is a ``no'', makes the negative groups more specific (the most general one counts on $29$ to $39\%$ of the test images), but the positive groups grow and count on the ambulance (its wheels and windows, $19\%$ of the image), which costs fidelity of the card on beagle and precision in the boxes ($89$ and $93\%$). Fine-tuning the one-sigmoid model with a random temperature per group, so that it has to decide correctly anywhere between the sigmoid and the step, gives no better card and lets TV remove groups that the card does not use (the wheel group of the bus). We therefore read the card from the groups of the monotone CNN, which all other experiments of the paper evaluate.

\begin{table}[!htbp]
\centering
\small
\caption{The same one-sided integer card (\cref{eq:card}, fitted as in \cref{app:card}) read from the vocabulary of every route of \cref{tab:card-routes} (beagle / school bus, mean over three seeds; \texttt{card\_areas\_routes.py}, \texttt{card\_routes\_summary.py}). Card: balanced test fidelity of the integer card. Kept area: the area of the groups on the card on teacher-positive test images. Kept count on neg.: the share of teacher-negative test images on which at least one group of the card counts. The recomputed groups of all $42$ models equal their stored test labels.}
\label{tab:routes-cards}
\fitcolumn{\begin{tabular}{@{}lccc@{}}
\toprule
vocabulary of & card (\%) & kept area, pos.\ (\%) & kept count on neg.\ (\%)\\
\midrule
monotone CNN surrogate (rerun) & 98.43 / 98.77 & 27.1 / 21.8 & 7.6 / 3.6\\
one sigmoid per group & 99.08 / 98.74 & 27.0 / 22.8 & 7.6 / 6.7\\
\quad start fixed at $-4$ & 98.81 / 99.65 & 30.6 / 28.2 & 11.1 / 7.3\\
\quad fine-tuned, random temperature & 98.72 / 99.35 & 25.1 / 21.7 & 5.9 / 3.9\\
\quad the same, log-uniform temperature & 99.05 / 99.57 & 25.9 / 22.2 & 8.3 / 3.0\\
step trained directly, threshold from $0.05$ & 97.92 / 98.42 & 36.3 / 29.6 & 27.4 / 4.8\\
step trained directly, thresholds from $0.02$, $0.1$ & 98.11 / 98.83 & 31.9 / 27.2 & 16.0 / 8.6\\
\bottomrule
\end{tabular}}
\end{table}

Measured by the final explanation instead of by each route's own readout (\cref{tab:routes-cards}), the picture is more even than the comparison above suggests. With the same one-sided card, the vocabularies of the one-sigmoid routes give cards as faithful as that of the monotone CNN ($98.7$ to $99.7\%$ against $98.4$ and $98.8\%$), with kept groups of similar size and similar specificity; only the step trained directly keeps larger and more general groups (on beagle a kept group counts on $16$ to $27\%$ of the negatives, against $8\%$). For the one-sided card the monotone CNN is therefore one reasonable route rather than the only one; its advantage lies in the negative groups, which the two-sided card of \cref{app:sides,app:stress} needs and which the one-sigmoid routes turn into catch-alls.

\section{Echocardiography videos}
\label{app:echo}
AtomEx also explains video transformers whose decision reads several videos at once. We explain two echocardiography video ViTs, EchoJEPA-B~\cite{munim2026echojepa} and EchoFM~\cite{kim2024echofm}, each fine-tuned and probed for congenital heart defects in newborns. In short: the surrogates agree with the teachers on $87$--$94\%$ of the test samples and the two-sided integer cards read from their groups on $83$--$92\%$; the groups fall into three kinds, color Doppler, the black frame outside the scan sector and tissue inside it; in seven of the eight settings one to four groups are the color Doppler, with the sign of a defect, and every Doppler group is on the one-sided card of its sign. Of the two one-sided cards, the card of the evidence for the decision (\cref{eq:card}) is the more faithful for a defect in all three views wherever the surrogate has Doppler groups, and the card of the evidence against it for ``no defect'', as for edema on the chest X-rays. Removing the color from the input changes the decisions of all four teachers beyond a control of similar area, so the teachers do use the Doppler; the direction of the change agrees with the sign of the Doppler groups for one teacher (EchoJEPA-B, probe) and is the opposite for EchoFM and for the ``no defect'' decision of EchoJEPA-B, ft. Among the explanations we compared, the Doppler groups predict best which samples change most. Explained on every frame of whole videos, the Doppler group of EchoJEPA-B follows the color-flow jets, while two Doppler groups of EchoFM stay at one place of the frame. A pediatric cardiologist found that the Doppler group of EchoJEPA-B marks the abnormal flow of a hypoplastic left heart and that the card of \cref{fig:echo-card} covers most of the region of the defect, while no method, AtomEx included, marks a defect seen only in B-mode; the groups show where the teachers look: at the flow, at the acquisition and at fixed places in the frame (paragraph \emph{A cardiologist's reading}).

\paragraph{Teachers.} The data are echocardiograms of newborns with and without a congenital heart defect. A sample is a combination of one video of each of three views, short axis (sax), suprasternal (ssn) and four-chamber (4ch), and its class is the number of views whose video is marked as showing the defect: none, one or two, or all three. There are $1{,}158$ training, $209$ test and $99$ validation samples of $438$, $75$ and $57$ persons; the validation persons come partly from a newer acquisition source. A teacher encodes every view with one shared video ViT (a clip of $32$ frames), adds a learned embedding of the view to the tokens, and reads the tokens of all three views with one attentive-probe block (a learned query with cross-attention and an MLP, as in V-JEPA) followed by a linear head for the three classes. The backbones are EchoJEPA-B (ViT-B/16, V-JEPA~2.1 pretraining on MIMIC-IV-Echo, tubelets of two frames, $16\times14\times14$ tokens) and EchoFM (ViT-L/16, video masked autoencoder, tubelets of four frames, $8\times14\times14$ tokens). Each is trained in two ways: the whole model (ft) or only the view embedding and the probe on the frozen backbone (probe). On the test samples (four clips per video, as in the evaluation of the teachers) the accuracy, balanced accuracy and AUROC are $82.8$, $81.8$ and $0.937$ (EchoJEPA-B, ft), $82.3$, $76.1$, $0.905$ (EchoJEPA-B, probe), $81.3$, $77.0$, $0.931$ (EchoFM, ft) and $78.9$, $78.7$, $0.903$ (EchoFM, probe). We explain two decisions of each teacher, ``the top-1 class is no defect'' and ``the top-1 class is a defect in all three views'', on the first evaluation clip of every video; its top-1 agrees with that of the four clips on $93$--$99.5\%$ of the test samples.

\paragraph{Descriptors and training.} The descriptor of a token is what the probe reads: the layer-normalized key and value input, i.e.\ the backbone token plus the view embedding (EchoFM's class token is dropped). A sample has $3\times16$ (EchoJEPA-B) or $3\times8$ (EchoFM) slices, one per view and time step, and every slice is treated as an image: for the ft teachers four shifted views of its frames (shifts $0$ and $8$ pixels) are interleaved into a $28\times28$ lattice, as for the chest X-rays (\cref{app:cxr}) but with reflection padding instead of real windows, and for the probe teachers the native $14\times14$ grid is used. One prototype softmax is shared by all slices (ten used and ten discard channels), the coefficients are signed, the monotone readout scores every slice, and the logit of the sample is the mean of the slice scores; the mean is monotone, so \cref{eq:sign} holds for the sample. TV is the spatial TV of every slice. The surrogates are fitted to the teacher decisions on $80\%$ of the training persons ($938$ samples), the epoch and the threshold are chosen on the other $20\%$ ($220$ samples), and we report the test and the validation samples; $40$ epochs, TV weight $0$ or $0.05$, channel dropout $0.25$, three seeds.

\begin{table*}[tp]
\centering
\small
\caption{Echocardiography: surrogates with TV $0.05$ ($28\times28$ lattice for the ft teachers, $14\times14$ for the probe teachers; mean over three seeds, $\pm$ standard deviation). Fidelity: balanced accuracy against the teacher decision at the selection threshold (test: $209$ samples, validation: $99$). Lin.\ probe: a balanced logistic regression on the mean descriptor of every view, test / validation. The last four columns use the test slices: the number of Doppler groups of every seed (lift at least $3$) and their sign, their mean lift (share of colored pixels in the group divided by the share in all slices, $4.7\%$), the share of all colored pixels that they cover, and the share of the area of the used groups that lies in groups mostly outside the scan sector (the sector leaves $30\%$ of the frame black).}
\label{tab:echo}
\setlength{\tabcolsep}{2.8pt}
\begin{tabular}{llcccccccc}
\toprule
teacher & top-1 is & test fid.\ (\%) & AUROC & val.\ fid.\ (\%) & lin.\ probe & Doppler groups & lift & Dopp.\ covered & outside\\
\midrule
EchoJEPA-B, ft & defect, 3 views & $86.9\pm1.5$ & $.98$ & $71.8$ & $90.9$ / $74.7$ & $0$, $0$, $0$ & -- & $0\%$ & $41\%$\\
 & no defect & $87.3\pm0.5$ & $.96$ & $83.8$ & $85.6$ / $83.9$ & $1$, $1$, $1$ ($-$) & $13.8$ & $34\%$ & $44\%$\\
EchoJEPA-B, probe & defect, 3 views & $92.7\pm1.7$ & $.97$ & $89.9$ & $94.3$ / $89.4$ & $1$, $1$, $1$ ($+$) & $15.2$ & $31\%$ & $20\%$\\
 & no defect & $93.7\pm0.6$ & $.97$ & $91.4$ & $82.5$ / $88.9$ & $1$, $1$, $1$ ($-$) & $10.9$ & $26\%$ & $27\%$\\
EchoFM, ft & defect, 3 views & $92.5\pm0.7$ & $.99$ & $81.3$ & $90.2$ / $81.2$ & $3$, $3$, $2$ ($+$) & $7.0$ & $14\%$ & $50\%$\\
 & no defect & $91.0\pm0.6$ & $.96$ & $87.4$ & $92.1$ / $92.4$ & $3$, $3$, $3$ ($-$) & $6.2$ & $28\%$ & $21\%$\\
EchoFM, probe & defect, 3 views & $87.4\pm2.4$ & $.97$ & $76.1$ & $89.3$ / $76.0$ & $4$, $3$, $3$ ($+$) & $7.5$ & $15\%$ & $32\%$\\
 & no defect & $91.2\pm0.9$ & $.98$ & $73.2$ & $93.3$ / $80.5$ & $3$, $3$, $3$ ($-$) & $7.5$ & $22\%$ & $42\%$\\
\bottomrule
\end{tabular}
\end{table*}

\paragraph{Fidelity.} \Cref{tab:echo} gives the numbers. On the test samples the surrogates agree with the teacher on $87$--$94\%$, at an AUROC of $0.96$--$0.99$, and are within $-4$ to $+11$ points of a linear probe on the mean descriptor of every view, which has no spatial structure. On the validation samples, from the persons and the newer source that the surrogates never saw, the fidelity falls to $72$--$91\%$, most for the decisions of the EchoFM probe and of ``a defect in all three views''. TV $0.05$ lowers the area of the used groups from $13$--$28\%$ to $10$--$20\%$ of the frame and changes the fidelity by $-0.9$ to $+2.1$ points, except for the EchoFM probe on defects, which gains from it ($87.4$ against $81.2\pm5.6\%$). The $28\times28$ lattice changes the fidelity of the ft teachers by $-1.2$ to $+1.5$ points against $14\times14$. An unconstrained readout (one seed, $14\times14$, as in \cref{app:sweep}) is within $-1$ to $+4$ points of the signed one on six of the eight decisions and lower on the two of the EchoFM probe ($80.7$ and $59.5\%$).

\begin{figure*}[tp]
\centering
\includegraphics[height=.8\textheight]{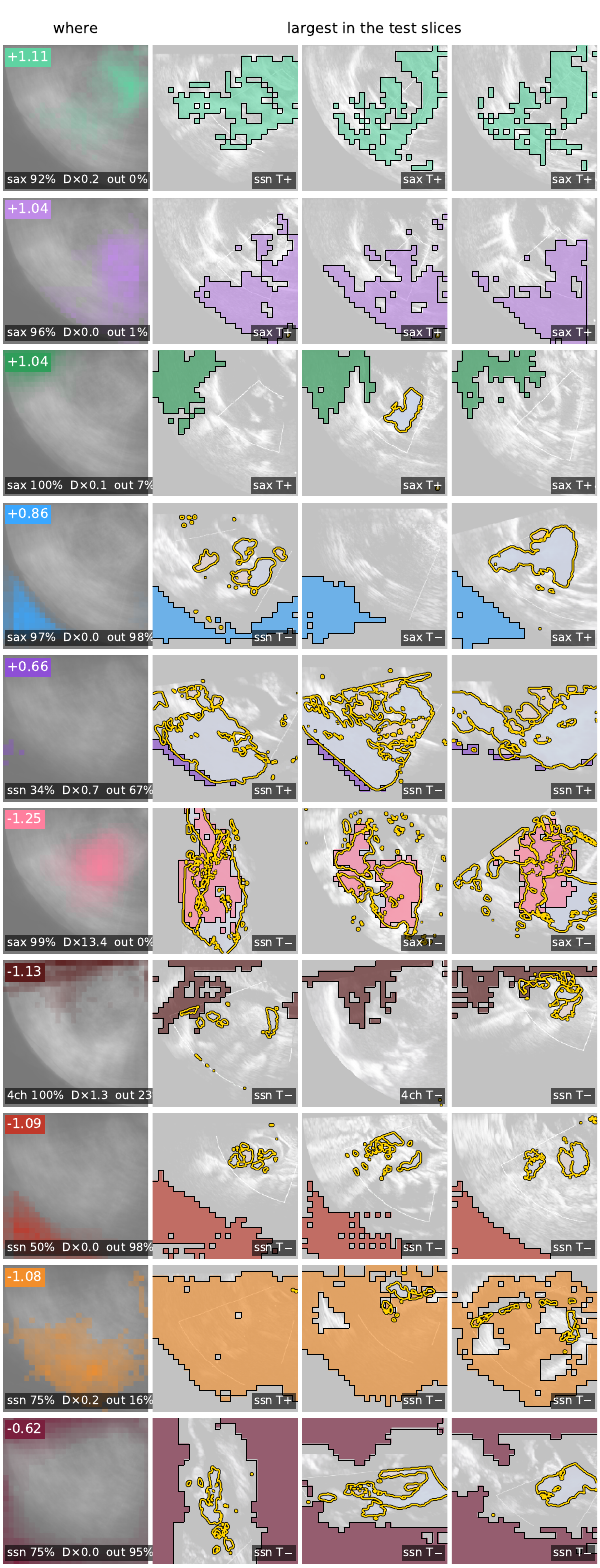}\hspace{2em}
\includegraphics[height=.8\textheight]{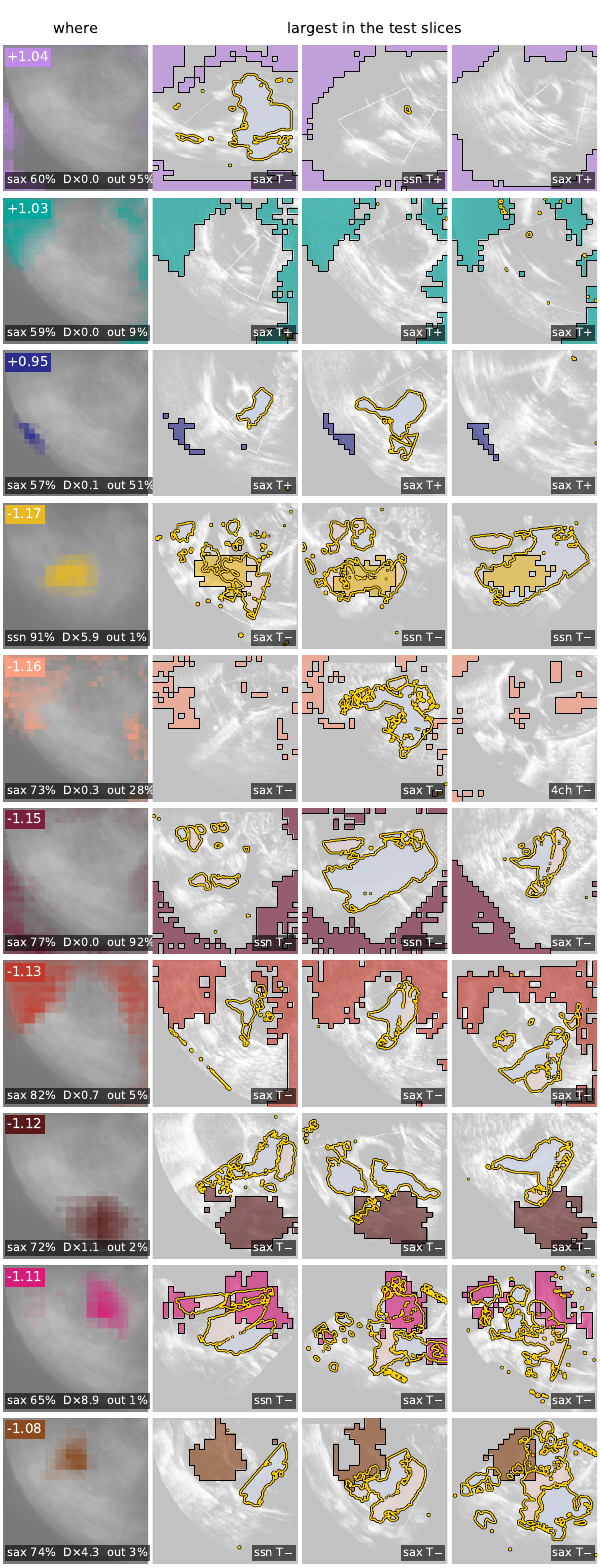}
\caption{The used groups of the ``no defect'' surrogates of EchoJEPA-B (left) and EchoFM (right), both ft (seed $101$, TV $0.05$, $28\times28$ lattice), one group per row, positive groups first, each in its own color as in \cref{fig:teaser} (cool: positive coefficient, warm: negative; the label gives the coefficient). First column: in the view where the group is most frequent, the share of the test slices in which a frame pixel belongs to the group (opacity), over the mean frame of that view; the view, the share of samples in which the group appears in that view, the Doppler lift of the group (D$\times$) and the share of its area outside the scan sector. Next three: the test slices (one per person) in which the group is largest, with the group (black outline), the colored Doppler pixels (yellow) and the teacher decision (T$+$/T$-$).}
\label{fig:echo-groups}
\end{figure*}

\paragraph{What the groups are.} \Cref{fig:echo-groups} shows the used groups of the ``no defect'' surrogates of both ft teachers on the test slices. Two properties of a slice tell the kinds apart. A pixel is \emph{color Doppler} if its RGB channels differ by more than $40$ (the color-flow overlay is the only colored part of an echo frame), and \emph{outside the sector} if it is never brighter than $12$ in the clip. A group is a Doppler group if its share of colored pixels is at least three times that of all slices. In seven of the eight settings, and in all three seeds of each, one to four groups are color Doppler (lift $6$--$15$; they cover $14$--$34\%$ of all colored pixels at $2$--$5\%$ of the frame), and their sign is that of a defect: positive for ``a defect in all three views'', negative for ``no defect''. EchoFM spreads the Doppler over two to four groups, EchoJEPA-B puts it into one; for EchoJEPA-B, ft, on defects the Doppler lies in the discard channels. Groups mostly outside the sector take $20$--$50\%$ of the area of the used groups, with both signs, about as much as the $30\%$ of the frame that lies there; their tokens carry no local image content, so what they contribute is context, the geometry of the sector, or information mixed into them by attention. The remaining groups lie on tissue inside the sector.

\paragraph{Scorecards.} The integer cards are read from the frozen groups as for the chest X-rays (\cref{app:cxr}): the area $a_k$ of a group is its share of all sites of the sample, over all slices of the three views ($3\times16$ slices of $28\times28$ sites for EchoJEPA-B, ft, down to $3\times8$ slices of $14\times14$ sites for EchoFM, probe), the cards are fitted to the teacher's decisions on the fit samples with their thresholds on the selection samples (\cref{app:card}), and \cref{tab:echo-cards} reports them on the test and the validation samples. The cards of \cref{eq:card} have three to six groups with $1$ to $6$ points and need $2$ to $8$ points; none accepts a sample on which none of its groups counts. Which one-sided card is the more faithful depends on the decision. For a defect in all three views the card of the evidence for it is within $3.1$ points of the surrogate wherever the surrogate has Doppler groups ($88.3$--$91.5\%$) and above the card of the groups against it; for EchoJEPA-B, ft, whose Doppler lies in the discard channels, it falls to $75.1\%$. For ``no defect'' the card of the groups against it, ``no defect unless their points reach a threshold'', is the more faithful for all four teachers ($79.4$--$93.2\%$ against $78.5$--$86.9\%$): as for edema on the chest X-rays, the decision is recognized better by the absence of the evidence of the other side, here of a defect, than by groups of its own. The color Doppler is on the cards in every case: each of the $45$ Doppler groups of the $21$ surrogates that have them is on the one-sided card of its sign, with $3$ to $5$ points for the one Doppler group of EchoJEPA-B and $1$ to $3$ points for each of the two to four of EchoFM ($5$ and $8$ in one seed). \Cref{fig:echo-card} shows the card of the EchoJEPA-B probe for a defect in all three views, the setting whose Doppler sign the teacher confirms (below). The two-sided card and the combination of the two one-sided cards stay within $4.8$ points of the surrogate on the test samples ($88.8$ and $88.4\%$ on average, against $90.3\%$; the two-sided card keeps seven or eight groups). On the validation samples, from the newer source, they are as faithful as the surrogate on average ($82.0$ and $81.4\%$ against $81.9\%$; per setting from $-5.6$ to $+10.3$ points) and above it where it falls most ($80.3$ against $71.8\%$ for the defect decision of EchoJEPA-B, ft, and $83.3$ against $73.2\%$ for ``no defect'' of the EchoFM probe, two-sided). Without TV the cards are about as faithful (two-sided $88.8\%$, combined $89.4\%$, surrogate $89.2\%$; test samples). Read per view instead of over the whole sample, a condition names the view as well (``group $k$ covers more than $m_k$ of the ssn view''). Such one-sided cards use seven or eight conditions, the limit, and are slightly more faithful on the test samples ($84.3$ and $86.4\%$ for and against the decision, against $83.7$ and $84.9\%$) but less so on the validation samples ($76.3$ and $79.5\%$ against $78.3$ and $80.3\%$); the per-view card of the EchoJEPA-B probe for a defect in all three views is nearly as faithful as its surrogate ($92.6$ against $92.7\%$). Their conditions on the Doppler groups name sax and ssn, the views with color, for EchoJEPA-B, and for EchoFM also 4ch ($15$ of its $55$ such conditions), whose videos have no color: as on whole videos (below), the Doppler groups of EchoFM mark a place in the frame as well as the flow.

\begin{table*}[tp]
\centering
\small
\caption{Echocardiography: balanced fidelity to the teacher (\%, mean over three seeds) of the surrogates of \cref{tab:echo} (TV $0.05$) and of the integer cards read from their groups: the card of the evidence for the decision (for, \cref{eq:card}: the groups with $\gamma_k>0$), the card of the groups with $\gamma_k<0$ (against: the decision unless their points reach a threshold), the two-sided card (\cref{app:sides}) and the two one-sided cards combined by $aS_+-bS_->c$ (\cref{app:cxr}). The area of a group is its share of all sites of the sample, all slices of the three views; the cards are fitted on the fit samples and their thresholds chosen on the selection samples (\texttt{echo\_card\_areas.py}, then the commands of the chest X-rays). Groups: the number of groups on the card for and on the card against (range over the seeds).}
\label{tab:echo-cards}
\setlength{\tabcolsep}{3pt}
\resizebox{\textwidth}{!}{%
\begin{tabular}{llccccccccccc}
\toprule
 & & \multicolumn{5}{c}{test ($209$ samples)} & \multicolumn{5}{c}{validation ($99$ samples)} & \\
\cmidrule(lr){3-7}\cmidrule(lr){8-12}
teacher & top-1 is & surrogate & for & against & two-sided & combined & surrogate & for & against & two-sided & combined & groups\\
\midrule
EchoJEPA-B, ft & defect, 3 views & $86.9$ & $75.1$ & $85.0$ & $88.3$ & $86.4$ & $71.8$ & $72.1$ & $75.6$ & $80.3$ & $73.2$ & $3$--$4$ / $6$--$7$\\
 & no defect & $87.3$ & $78.5$ & $79.4$ & $83.1$ & $82.5$ & $83.8$ & $74.9$ & $81.9$ & $81.0$ & $84.4$ & $5$ / $5$\\
EchoJEPA-B, probe & defect, 3 views & $92.7$ & $89.6$ & $76.7$ & $90.5$ & $90.7$ & $89.9$ & $85.9$ & $78.3$ & $84.3$ & $86.4$ & $5$--$6$ / $4$--$5$\\
 & no defect & $93.7$ & $81.0$ & $93.2$ & $91.1$ & $92.3$ & $91.4$ & $91.3$ & $90.6$ & $90.7$ & $91.5$ & $4$--$5$ / $5$--$6$\\
EchoFM, ft & defect, 3 views & $92.5$ & $91.5$ & $88.4$ & $92.1$ & $90.3$ & $81.3$ & $78.9$ & $72.8$ & $77.4$ & $76.0$ & $6$ / $4$\\
 & no defect & $91.0$ & $86.9$ & $88.6$ & $89.3$ & $87.9$ & $87.4$ & $79.7$ & $87.2$ & $84.5$ & $82.9$ & $3$--$4$ / $6$--$7$\\
EchoFM, probe & defect, 3 views & $87.4$ & $88.3$ & $79.4$ & $87.1$ & $87.4$ & $76.1$ & $74.9$ & $73.6$ & $74.6$ & $73.2$ & $5$--$6$ / $4$--$5$\\
 & no defect & $91.2$ & $79.0$ & $88.3$ & $88.9$ & $89.4$ & $73.2$ & $68.7$ & $82.5$ & $83.3$ & $83.5$ & $3$--$6$ / $4$--$6$\\
\midrule
mean & & $90.3$ & $83.7$ & $84.9$ & $88.8$ & $88.4$ & $81.9$ & $78.3$ & $80.3$ & $82.0$ & $81.4$ & \\
\bottomrule
\end{tabular}}
\end{table*}

\begin{figure*}[tp]
\centering
\includegraphics[width=\textwidth]{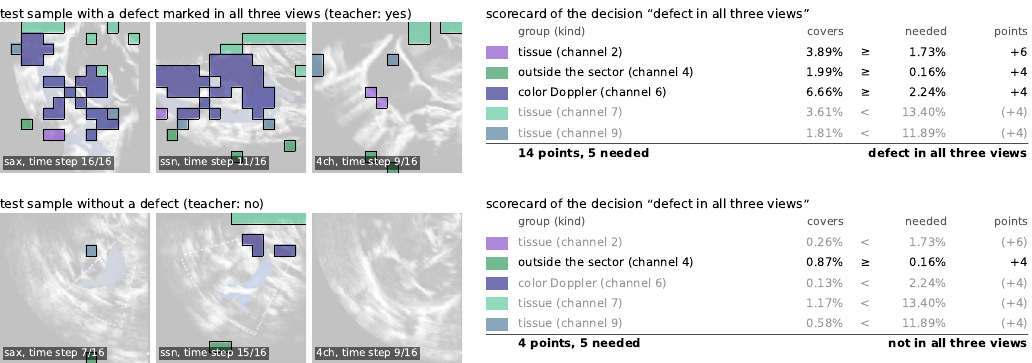}
\caption{The card of \cref{eq:card} of the EchoJEPA-B probe surrogate for ``a defect in all three views'' (seed $101$, TV $0.05$; test fidelity $88.5\%$, surrogate $92.6\%$), drawn as in \cref{fig:card-beagle}, on two test samples fixed in advance: the first of a newborn with a defect marked in all three views and the first of a healthy one (teacher: whether its top-1 class is a defect in all three views). Left: one frame of every view, the time step with the most colored pixels (the 4ch videos have none: their middle step), with the groups of the card on the native $14\times14$ lattice, the lattice the card reads. Right: the card with the area of every group over all $48$ slices of the sample ($9{,}408$ sites), the share it needs and its points; the kinds are those of \cref{fig:echo-groups}. Channels $7$ and $9$ need more than any test or validation sample reaches, so in effect the card accepts a sample when the tissue group of channel $2$ counts ($6$ points) or when the Doppler group and the group outside the sector (channel $4$) both count ($4+4$). Channel $4$ counts on $85\%$ of the test samples that the teacher does not call so; the Doppler group counts on $16\%$ of them and on $84\%$ of those that it calls so.}
\label{fig:echo-card}
\end{figure*}

\begin{figure*}[tp]
\centering
\includegraphics[width=.94\textwidth]{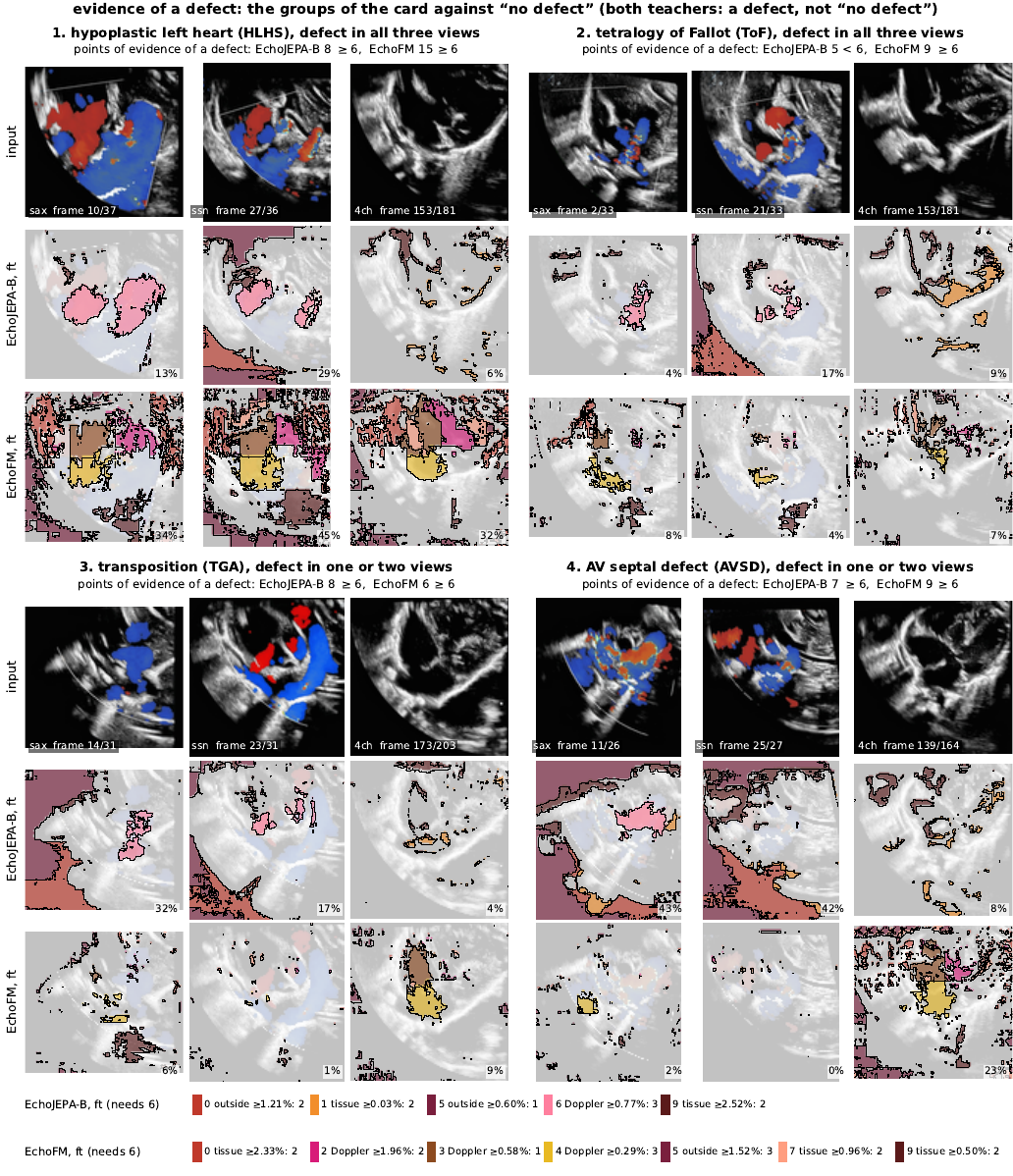}
\caption{Whole videos: four training samples chosen by us, one frame of every view, the input and, for both ft teachers (seed $101$, TV $0.05$, best-fidelity checkpoint, $28\times28$ lattice), the evidence of a defect: the groups of the card against ``no defect'' (\cref{tab:echo-cards}) as dense maps ($256$ shifted views, as in \cref{fig:gallery}), each in its own color, the other groups not drawn; key: channel, kind, the share of all slices of the sample it needs, and its points. Above each block: the points each sample collects (``a defect'' from $6$). Teachers and surrogates answer ``a defect'' (not ``no defect'') on all four; the card of EchoJEPA-B misses ToF. The views marked as showing the defect: all three for HLHS and ToF, sax and ssn for TGA, 4ch for AVSD. Frame: the one of the explained clip with the most colored pixels (its middle frame for the 4ch videos, which have no color), shown at the aspect ratio of the recording (the teacher reads it resized to $224\times224$); percentage: area of the drawn groups. The frames are those of movies in which every frame of the twelve videos is explained.}
\label{fig:echo-frames}
\end{figure*}

\begin{figure*}[tp]
\centering
\includegraphics[width=\textwidth]{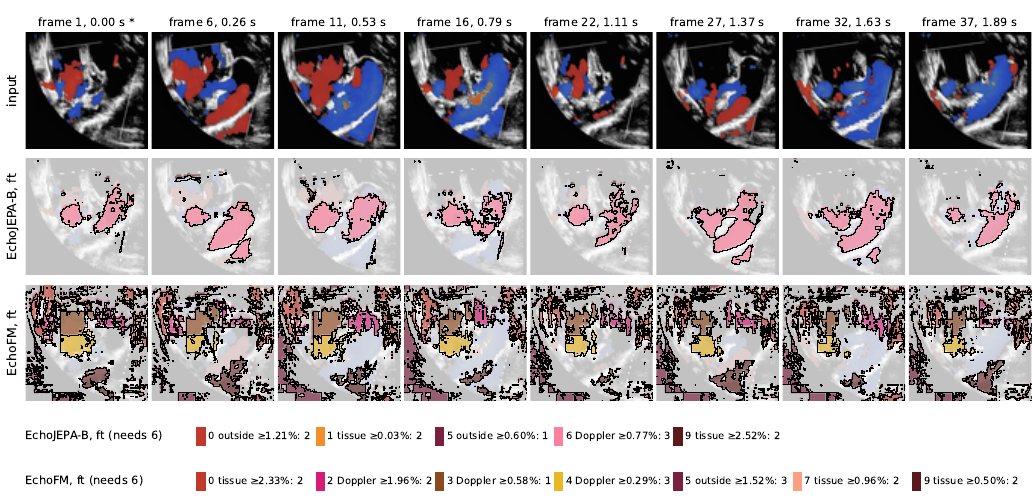}
\caption{Eight evenly spaced frames of the sax video of the HLHS sample of \cref{fig:echo-frames} ($37$ frames at $19$ frames per second; frame $1$, marked $*$, lies outside the explained clip, the last $32$ frames), with the evidence of a defect of \cref{fig:echo-frames}. The Doppler group of EchoJEPA-B (pink, channel $6$) moves with the jets; the Doppler groups of EchoFM in the upper middle and the middle of the frame (brown and gold, channels $3$ and $4$) stay in place.}
\label{fig:echo-film}
\end{figure*}

\paragraph{Whole videos.} The tables use one clip per video, the last $32$ frames of the frame grid resampled to the frame rate of the view (window $0$). To explain every frame, we cut further clips of $32$ frames from the same grid and, where the resampling skips frames (about one in six for ssn, up to one in two for 4ch), also from the same grid moved back by one or two frames; a frame takes the groups of its time step (two frames for EchoJEPA-B, four for EchoFM) in the first clip that contains it, window $0$ first, so every map comes from a clip that the teacher reads. We did this for the twelve videos of four samples chosen for their diagnoses: hypoplastic left heart (HLHS) and tetralogy of Fallot (ToF), marked in all three views, transposition of the great arteries (TGA), marked in sax and ssn, and an atrioventricular septal defect (AVSD), marked in 4ch. All four are training samples of the teachers, three were used to fit the surrogates and the fourth to select them, so the figures illustrate the explanations rather than test them. Both ft teachers give all four the right class, and teachers and ``no defect'' surrogates answer ``no''. \Cref{fig:echo-frames} shows one frame of every view and \cref{fig:echo-film} eight frames of one video. They draw ``no defect'' the way its card reads it (\cref{tab:echo-cards}), as the evidence against it: only the groups of the card against ``no defect'', each with its kind and points. On these samples that card collects $6$--$15$ points for EchoFM and $7$ or $8$ for EchoJEPA-B, which needs $6$, except for ToF ($5$); for HLHS and TGA the Doppler group of EchoJEPA-B counts, for AVSD a tissue group (channel $9$). Not all of the evidence is the heart: the tissue group of channel $1$ of EchoJEPA-B counts on $83\%$ of the test samples of both kinds, and its two groups outside the sector count on $63$--$69\%$ of the samples the teacher does not call ``no defect'' against $23$--$25\%$ of those it does, so the acquisition is evidence too. The Doppler group of EchoJEPA-B follows the color-flow jets from frame to frame: on the sax and ssn videos of HLHS, which have the most color ($19\%$ and $11\%$ of the frame), it covers $45\%$ and $40\%$ of the colored pixels of a frame on average ($26$--$72\%$ and $14$--$56\%$ over the frames), and $75\%$ and $83\%$ of its area is colored; on the other three samples, with $1$--$6\%$ color, it covers $0$--$48\%$, and $67$--$82\%$ of the used area of the sax and ssn frames of TGA and AVSD lies in the black frame outside the sector. The three Doppler groups of EchoFM cover $29\%$ and $65\%$ of the color of HLHS, but two of them (channels $3$ and $4$) sit at nearly the same place in every view of every sample, with or without color (over the frames of a video, the standard deviation of their center is at most $4\%$ of the frame), and the three groups take $3$--$13\%$ of the 4ch frames, which have no color. For this teacher they mark a place in the frame as much as the flow. The logit of the surrogate is the mean of the slice scores, so the three views add up, and the two surrogates draw their answer from different views: for AVSD, marked in 4ch, the 4ch slices of EchoFM score $-18.8$ on average against $+0.5$ (sax) and $+3.6$ (ssn), while those of EchoJEPA-B score $-3.4$ against $-9.9$ and $-12.5$.

\begin{figure*}[tp]
\centering
\includegraphics[width=.92\textwidth]{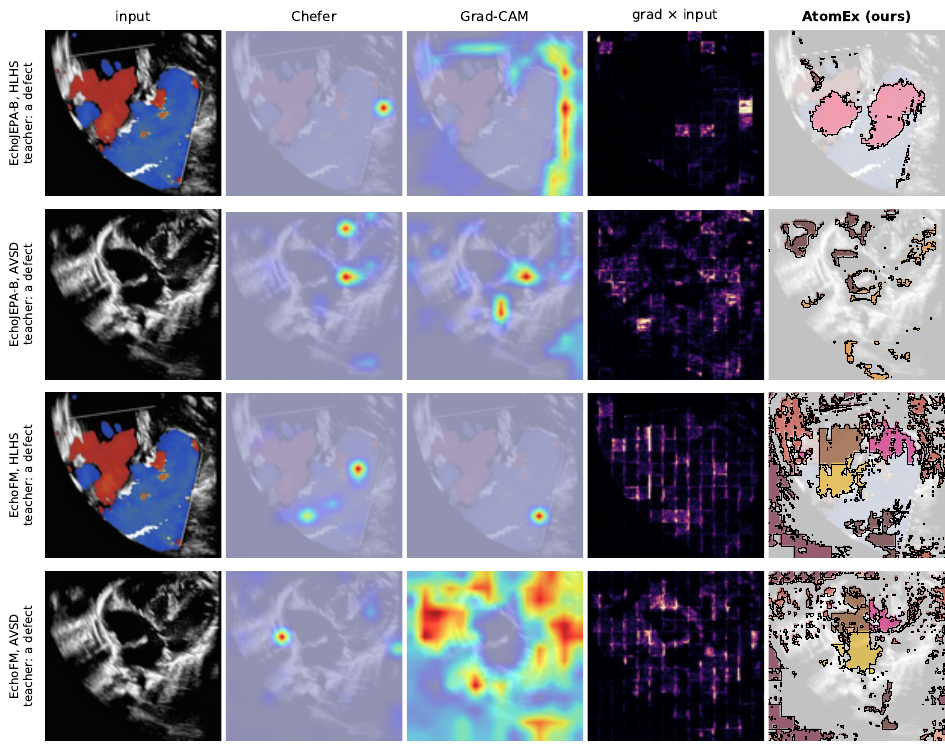}
\caption{Attribution methods next to AtomEx on the decision ``no defect'' of both ft teachers, as in \cref{fig:teaser}: the sax frame of HLHS and the 4ch frame of AVSD of \cref{fig:echo-frames} (both teachers: a defect), Chefer's rule on the probe and Grad-CAM (ReLU) on the tokens the probe reads (jet, each map scaled to its maximum), gradient $\times$ input on the pixels (absolute, its own scale), and the evidence of a defect of AtomEx as in \cref{fig:echo-frames}. On AVSD EchoJEPA-B draws its answer from the sax and ssn views (mean slice scores $-9.9$ and $-12.5$ against $-3.4$ for 4ch), EchoFM from the 4ch view ($-18.8$).}
\label{fig:echo-compare}
\end{figure*}

\paragraph{A cardiologist's reading.} We asked a pediatric cardiologist to judge the groups on the samples of this section. (i)~For the flow defect, HLHS, the Doppler group of EchoJEPA-B covers the regions of abnormal flow well (\cref{fig:echo-groups,fig:echo-frames,fig:echo-film}) and partly in the first row of \cref{fig:echo-compare}; the groups of EchoFM overlap them partly in \cref{fig:echo-groups} and miss them in \cref{fig:echo-film}. (ii)~On the card of \cref{fig:echo-card} most of the region of the defect is covered; on the healthy sample the groups are too few to confirm a normal heart. (iii)~Where the defect is visible in B-mode, as AVSD in the 4ch view, which most clinicians would recognize at once, none of the four methods of \cref{fig:echo-compare}, AtomEx included, marks it, and on ToF and TGA the groups lie away from the defect. Our measurements say why. The teachers read the Doppler (\cref{tab:echo-int}), and their evidence against ``no defect'' is the Doppler, the acquisition and fixed places in the frame. On AVSD the two teachers draw their answers from different views: EchoJEPA-B from the Doppler views sax and ssn, so its 4ch frame carries little of the decision, and EchoFM from the 4ch view, where its groups do not lie on the septal defect. And ``no defect'' is read by its card against, as the absence of evidence of a defect rather than as evidence of a normal heart (\cref{tab:echo-cards}), which is what the cardiologist sees on the healthy sample. The groups thus show a clinician what the teachers use: the abnormal flow where there is color, and otherwise the acquisition, not the anatomy of a defect seen in B-mode.

\begin{table*}[tp]
\centering
\small
\caption{Echocardiography: removing the color Doppler from the input of the teacher ($308$ test and validation samples). Groups: sign of the AtomEx Doppler groups (seed $101$, TV $0.05$). Dark $-$ shift: mean change of the margin of the decision when the colored pixels are set to black, minus the change when the same mask, moved to uncolored pixels, is set to black (bootstrap interval over samples; several samples can come from one person, so these intervals are optimistic); pred.\ dir.: share of samples where it has the direction that the sign of the groups predicts; swap: mean change when red and blue are exchanged in the colored pixels. Spearman $\rho$ over samples between the change and the prediction of each method: AtomEx, its score with the Doppler groups removed minus its score; col.\ area, the share of colored pixels with the direction of the groups; grad$\times$in., gradient $\times$ input summed over the colored pixels; 1st order, gradient $\cdot$ (gray input $-$ input), a first-order prediction of graying the colored pixels rather than of the compound edit dark $-$ shift; Grad-CAM, signed, on the tokens of the probe. Sign agrees: share of samples where the prediction and the change have the same sign.}
\label{tab:echo-int}
\setlength{\tabcolsep}{3pt}
\resizebox{\textwidth}{!}{%
\begin{tabular}{llcccccccccc}
\toprule
 & & \multicolumn{4}{c}{teacher: change of the margin} & \multicolumn{5}{c}{Spearman $\rho$ with the change (dark $-$ shift)} & sign agrees\\
\cmidrule(lr){3-6}\cmidrule(lr){7-11}\cmidrule(lr){12-12}
teacher & top-1 is & groups & dark $-$ shift [95\%] & pred.\ dir. & swap & AtomEx & col.\ area & grad$\times$in. & 1st order & Grad-CAM & AtomEx / grad$\times$in.\\
\midrule
EchoJEPA-B, ft & defect, 3 views & none & $-0.65$ [$-1.01$, $-0.30$] & -- & $+2.52$ & -- & $.54$ & $.06$ & $.22$ & $-.72$ & -- / $53\%$\\
 & no defect & $-$ & $-0.37$ [$-0.59$, $-0.16$] & $37\%$ & $-2.40$ & $.53$ & $.31$ & $.09$ & $.28$ & $-.51$ & $37$ / $60\%$\\
EchoJEPA-B, probe & defect, 3 views & $+$ & $-1.53$ [$-1.71$, $-1.36$] & $94\%$ & $+1.05$ & $.78$ & $.63$ & $-.01$ & $.11$ & $-.67$ & $93$ / $41\%$\\
 & no defect & $-$ & $+1.01$ [$+0.89$, $+1.13$] & $88\%$ & $-0.93$ & $.54$ & $.44$ & $.04$ & $.19$ & $-.57$ & $88$ / $39\%$\\
EchoFM, ft & defect, 3 views & $+$ & $+0.92$ [$+0.70$, $+1.14$] & $26\%$ & $+2.47$ & $.46$ & $.38$ & $.22$ & $.20$ & $-.39$ & $25$ / $73\%$\\
 & no defect & $-$ & $-0.63$ [$-0.81$, $-0.43$] & $34\%$ & $-1.75$ & $.48$ & $.36$ & $.29$ & $.28$ & $-.40$ & $33$ / $68\%$\\
EchoFM, probe & defect, 3 views & $+$ & $+0.53$ [$+0.36$, $+0.70$] & $33\%$ & $+1.85$ & $.56$ & $.44$ & $.12$ & $.21$ & $-.21$ & $33$ / $67\%$\\
 & no defect & $-$ & $-0.35$ [$-0.50$, $-0.19$] & $35\%$ & $-1.40$ & $.70$ & $.51$ & $.31$ & $.25$ & $-.24$ & $35$ / $63\%$\\
\bottomrule
\end{tabular}}
\end{table*}

\paragraph{Doppler and the acquisition.} Color Doppler is diagnostic for many defects, since it shows abnormal flow, but how often it is recorded is a choice of the protocol. In the training data $72\%$ of the combinations of sick children and $16\%$ of those of healthy ones contain a short clip ($\le47$ frames at $\le24$ frames per second), mostly a Doppler clip. The share of colored pixels alone predicts the decision ``a defect in all three views'' with an AUROC of $0.83$--$0.90$ on the test samples ($0.63$--$0.72$ on the validation samples), and the healthy validation newborns whom a teacher does not call ``no defect'' have $4.2$--$4.3\%$ colored pixels against $2.8$--$3.4\%$ for those it does. The groups do not say whether a teacher reads the flow of a defect or the presence of Doppler as such; they say that it reads the Doppler, with the sign of a defect, and where. The next paragraph tests this on the teacher.

\paragraph{Intervention on the teacher.} We change only the colored pixels (RGB spread above $20$, so that the faint rim of the overlay goes too) of the three clips of a sample and pass them through the teacher: \emph{dark} sets them to black (under the overlay lies mostly blood, which is dark in B-mode), \emph{gray} replaces them by their luminance, \emph{swap} exchanges the red and the blue channel (the color stays, the direction of the flow is reversed), and the control \emph{shift} sets to black the same mask moved to uncolored pixels inside the sector ($82\%$ of its area on the median). Nearly every sax and ssn video is a color-Doppler recording (median $6\%$ colored pixels; the 4ch videos have almost none), so there are no samples without color to serve as a null. \Cref{tab:echo-int} gives the change of the smooth margin $z_c-\log\sum_{j\neq c}e^{z_j}$ of the decision on the $308$ test and validation samples. The control moves the margins by $0.5$ at most, and removing the Doppler (dark minus shift) changes every decision beyond it, but in the direction of the Doppler groups only for the EchoJEPA-B probe: there the margin of ``a defect in all three views'' falls by $1.53$ on average and in $94\%$ of the samples, half of these decisions are lost ($130$ of the samples before, $67$ after), and the margin of ``no defect'' rises in $88\%$. For both EchoFM teachers the direction is the opposite ($25$--$35\%$ of the samples in the predicted direction); for EchoJEPA-B, ft, whose Doppler lies in the discard channels for the defect decision, the margin of that decision falls ($-0.65$), and its ``no defect'' decision moves against the sign of its Doppler group ($37\%$ in the predicted direction). Reversing the colors moves all four teachers toward a defect ($+1.1$ to $+2.5$ for the defect decision), about as much as or more than removing them: the teachers read the direction of the flow, not only its presence. The Doppler groups thus name evidence that the teachers use, but their sign, one per group, is right for one teacher; for EchoFM the same colored region counts for or against a defect depending on its colors, which one signed group cannot express, nor a condition of a card, which counts for one side only. Re-run on the descriptors of the changed clips, the surrogates keep their fidelity for the EchoJEPA-B probe ($93$--$95\%$ on the test samples) and lose it for the other teachers ($53$--$82\%$).

\paragraph{Comparison with attribution methods.} Which explanation predicts how much a sample changes? For every method we sum its attribution over the colored pixels of the original input and compare it over the samples with the change (dark minus shift), by Spearman's $\rho$ and by the share of samples where the two have the same sign. AtomEx predicts the change by its readout alone: the score with the Doppler groups removed from the maps minus the score (seed $101$). The baselines are written for this teacher: Grad-CAM on the tokens the probe reads (signed), gradient $\times$ input on the pixels, and the first-order prediction of the intervention, gradient $\cdot$ (gray input $-$ input), which knows the intervention; the attention of the probe's query and the rule of Chefer et al.~\cite{chefer2021transformer} applied to the probe are unsigned and predict no direction. We add a naive predictor, the share of colored pixels with the direction of the Doppler groups. The groups rank the samples best in all seven settings that have them ($\rho=0.46$--$0.78$), above the colored area alone ($0.31$--$0.63$), gradient $\times$ input ($-0.01$ to $0.31$) and the first-order prediction ($0.11$--$0.28$); Grad-CAM is anti-correlated ($-0.21$ to $-0.72$). \Cref{fig:echo-compare} shows the maps on two of the samples of \cref{fig:echo-frames}: Chefer's rule on the probe puts nearly all relevance on one to three tokens, Grad-CAM marks a few spots, the edge of the sector or the whole frame, gradient $\times$ input shows the patch grid, and none of them separates the color-flow jets from the tissue around them; the Doppler group of the EchoJEPA-B surrogate covers the jets as one negative group, while for EchoFM the jets lie among other negative groups that also cover tissue and the frame. The direction is another matter: the groups have the sign of the change in $88$--$93\%$ of the samples for the EchoJEPA-B probe and in $25$--$37\%$ for the other teachers, where the gradient methods, which have no fixed sign per region, do better ($60$--$74\%$), while for the EchoJEPA-B probe they are at chance ($39$--$57\%$).

\section{User study: predicting the model from its scorecard}
\label{app:userstudy}

We asked whether people can learn the decision rule of an image classifier from its integer scorecard.
Participants studied a model's scorecard with six solved examples and then predicted the model's YES/NO
answer on new photos. We never named the category the model detects.

\subsection{Models and scorecards}
All three models are ViT-B/16 image classifiers trained on ImageNet, and each decides whether a photo shows one target category. A scorecard lists coloured groups (parts of the image found by the model's surrogate). Each group adds points when it covers enough of the image, and the model answers YES when the total reaches
the threshold. Table~\ref{tab:us_design} gives the groups, points and threshold of the three scorecards.

\begin{table}[t]
\centering\footnotesize\setlength{\tabcolsep}{4pt}
\caption{The three models used in the user study. A group scores its points when it covers its required share of the
image; the model answers YES when the total reaches the threshold.}
\label{tab:us_design}
\begin{tabularx}{\columnwidth}{@{}cl>{\raggedright\arraybackslash}Xc@{}}
\toprule
Set & Target & Points per group & Needed \\
\midrule
1 & prairie chicken & blue 4, orange 3, green 4, red 1, purple 4 & 4 \\
4 & beagle & blue 2, orange 2, green 2, red 5, purple 2 & 6 \\
5 & school bus & blue 4, orange 4, green 4, red 2, purple 2, brown 2 & 5 \\
\bottomrule
\end{tabularx}
\end{table}

Thirty participants took part in an online form (Google Forms). Participation was voluntary and anonymous. The form
asked for no name, e-mail address, or other personal information, and no background questions were asked. It recorded
only the answers and a timestamp.
\begin{figure}[t]
\centering
\includegraphics[width=\linewidth]{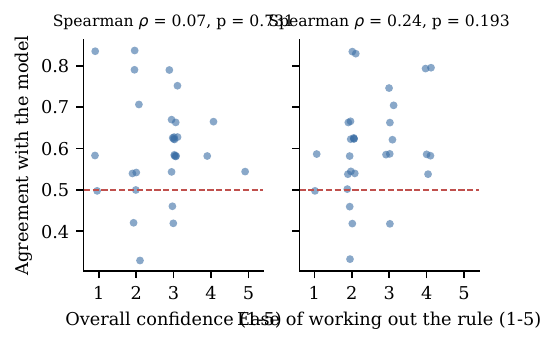}
\caption{Agreement with the model against self-reported confidence (left) and ease of working out the rule (right). Each dot is a participant
(jittered); dashed line: chance.}
\label{fig:us_confidence}
\end{figure}
\subsection{Materials and procedure}
The form had nine pages. The first described the study and what participants would do. The second explained how the
model answers (``add up the points of the coloured parts that cover enough of the photo; if the total reaches the
points needed, the model answers YES, otherwise NO'') and gave two solved examples from set~1. The next six pages
covered the three models in a fixed order. Each model got a study page with its scorecard and six solved examples
(the groups the model found, the points that count, and the model's answer). From the second set on, the page also
said that this was a new model detecting a different, again unnamed, object. A test page followed with eight photos,
and for each photo participants chose YES or NO for the model's answer. They could go back to the study page with the
browser's Back button. The ninth page held two closing questions (Section~\ref{sec:us_measures}). All participants saw
the models and photos in the same order, and there was no time limit.

\subsection{Photos}
Each model was tested on eight ImageNet validation photos, picked by a fixed rule from a larger stratified pool:
three photos of the target category; one look-alike that the model calls the target; one look-alike that the scorecard
accepts but the model rejects (a scorecard error); two look-alikes that both reject; and one other object of the same
kind (e.g.\ another animal) that both reject. Each set thus has four photos with the model answer YES and four with NO,
and the 24 photos contain 12 of each, so plain accuracy equals balanced accuracy. A participant who applies the
scorecard perfectly agrees with the model on 21 of the 24 photos (87.5\%), since each model has one scorecard error in
the sample. The photos are not a random sample: the rule over-represents look-alikes.

\subsection{Measures and analysis}
\label{sec:us_measures}
The ground truth is the model's own answer for each photo, and the main measure is how often a participant's answers
match it across the 24 photos. Each participant also rated (1--5) how sure they were of their answers overall
(1 = mostly guessing, 5 = very sure) and how easy it was to work out how the model decides (1 = very hard,
5 = very easy).

We tested mean agreement against chance (50\%) with a one-sample $t$-test and a Wilcoxon signed-rank test, and report
Cohen's $d=(\bar a-0.5)/s$. Confidence intervals are 95\% percentile bootstrap intervals over participants
(10\,000 resamples). Because answers from the same participant are correlated, we also fitted an intercept-only logistic
GEE with an exchangeable working correlation and participants as clusters. The three sets were compared with a Friedman
test and Holm-corrected pairwise Wilcoxon tests. From signal detection theory we computed the sensitivity $d'$
(answering YES to a model-YES photo is a hit, answering YES to a model-NO photo is a false alarm; counts were corrected
by adding 0.5 to each count and 1 to each total) and tested it against zero. Relations with the closing ratings were
tested with Spearman's $\rho$. Every analysis was run with and without the two participants who gave the same answer
to at least 22 of the 24 photos. All $p$-values are two-sided and uncorrected unless stated.

\subsection{Results}

\begin{figure*}[t]
\centering
\includegraphics[width=.95\linewidth]{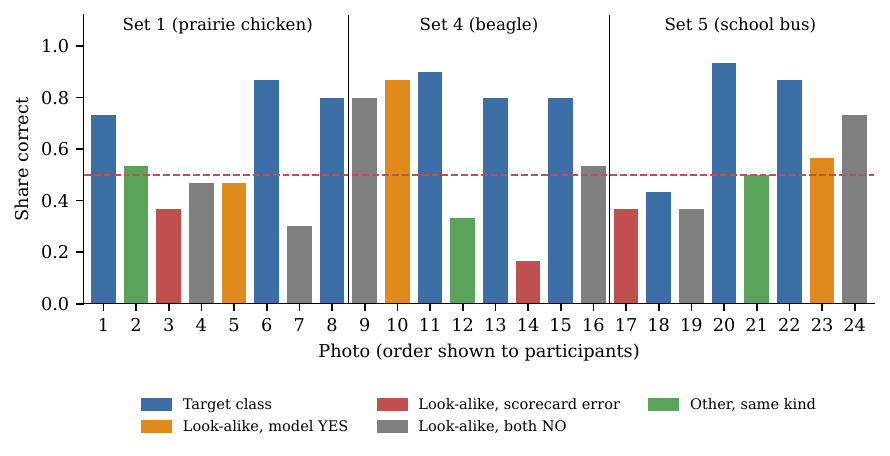}
\caption{Share of participants who agreed with the model, photo by photo, in the order shown to participants. Colour: photo type.}
\label{fig:us_photos}
\end{figure*}
Participants agreed with the model on $60.4\%$ of the photos on average (SD $12.1$, 95\% CI $[56.1,64.7]$), which is
above chance ($t(29)=4.71$, $p<0.001$; Wilcoxon $p<0.001$; $d=0.86$). The GEE gives the same estimate ($60.4\%$,
95\% CI $[56.1,64.6]$, $p<0.001$). Without the two near-constant responders the mean is $60.9\%$ ($N=28$, $p<0.001$).
Table~\ref{tab:us_accuracy} and Figure~\ref{fig:us_participants} show the results per model. Agreement was $65.0\%$ for
the beagle model, $59.6\%$ for the school-bus model and $56.7\%$ for the prairie-chicken model; the last is not
significantly above chance ($p=0.065$). Accuracy did not differ significantly between the three sets (Friedman
$\chi^2(2)=5.29$, $p=0.071$, Kendall's $W=0.09$; Holm-corrected Wilcoxon: set~1 vs.\ 4 $p=0.088$, set~1 vs.\ 5
$p=0.572$, set~4 vs.\ 5 $p=0.320$).

\begin{table}[t]
\centering\footnotesize\setlength{\tabcolsep}{3pt}
\caption{Agreement with the model's YES/NO answer (24 photos per participant, 12 YES and 12 NO). Mean and SD over participants;
CI = 95\% participant-cluster bootstrap; $t$-test and Wilcoxon signed-rank test against chance (50\%).}
\label{tab:us_accuracy}
\resizebox{\linewidth}{!}{%
\begin{tabular}{lrrcrrr}
\toprule
Photos & $N$ & Mean (SD), \% & 95\% CI & $t(29)$ & $p_t$ & $p_W$ \\
\midrule
All 24 & 30 & 60.4 (12.1) & [56.1, 64.7] & 4.71 & $<$0.001 & $<$0.001 \\
Set 1 (prairie chicken) & 30 & 56.7 (19.1) & [50.0, 63.3] & 1.92 & 0.065 & 0.058 \\
Set 4 (beagle) & 30 & 65.0 (15.2) & [59.6, 70.4] & 5.41 & $<$0.001 & $<$0.001 \\
Set 5 (school bus) & 30 & 59.6 (19.1) & [52.9, 66.2] & 2.76 & 0.010 & 0.013 \\
\midrule
All 24, without near-constant responders$^\dagger$ & 28 & 60.9 (12.4) & [56.4, 65.5] & 4.64 & $<$0.001 & $<$0.001 \\
\bottomrule
\end{tabular}}
\vspace{2pt}\par\footnotesize $^\dagger$ Participants who gave the same answer to at least 22 of the 24 photos.
\end{table}
\begin{figure}[h!]
\centering
\includegraphics[width=\linewidth]{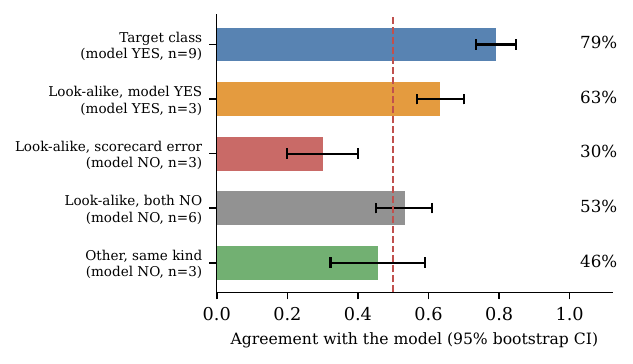}
\caption{Agreement with the model by photo type (mean over participants, 95\% participant-cluster bootstrap CI). Dashed line: chance.}
\label{fig:us_strata}
\end{figure}
\begin{figure}[t]
\centering
\includegraphics[width=\linewidth]{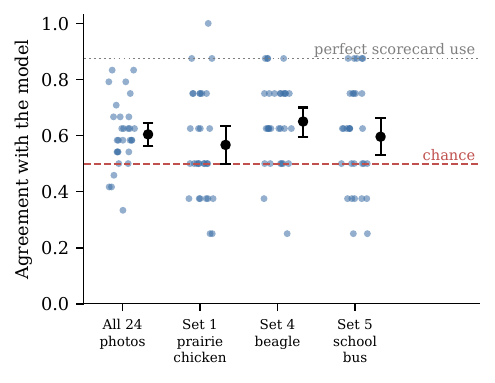}
\caption{Agreement of each participant (dots) with the model's answer, overall and per set. Black marker: mean with 95\% bootstrap CI.
Dashed line: chance; dotted line: a participant who applies the scorecard perfectly (87.5\%).}
\label{fig:us_participants}
\end{figure}

\paragraph{Agreement by photo type.}
Table~\ref{tab:us_strata} and Figures~\ref{fig:us_strata} and~\ref{fig:us_photos} give the results for each type of
photo. Agreement was highest on photos of the target category ($79.3\%$) and on look-alikes that the model calls the
target ($63.3\%$). The three photos where the scorecard says YES but the model says NO were the hardest: participants
agreed with the model on only $30.0\%$ of these answers (per photo, $37\%$, $17\%$ and $37\%$). They answered YES in
$70.0\%$ of these cases, so they followed the scorecard and not the model. Look-alikes that both reject ($53.3\%$) and
other objects of the same kind ($45.6\%$) stayed near 50\%.

\begin{table}[t]
\centering\footnotesize\setlength{\tabcolsep}{3pt}
\caption{Agreement with the model by type of photo (pooled over the three sets). The photos are not a random sample: they are drawn
by a fixed rule that over-represents look-alike classes. CI = 95\% participant-cluster bootstrap.}
\label{tab:us_strata}
\resizebox{\linewidth}{!}{%
\begin{tabular}{lcrcr}
\toprule
Photo type & Model & $n$ & Agreement, \% [95\% CI] & YES, \% \\
\midrule
Target class & YES & 9 & 79.3 [73.3, 84.8] & 79.3 \\
Look-alike, model YES & YES & 3 & 63.3 [56.7, 70.0] & 63.3 \\
Look-alike, scorecard error & NO & 3 & 30.0 [20.0, 40.0] & 70.0 \\
Look-alike, both NO & NO & 6 & 53.3 [45.0, 61.1] & 46.7 \\
Other, same kind & NO & 3 & 45.6 [32.2, 58.9] & 54.4 \\
\bottomrule
\end{tabular}}
\end{table}

\paragraph{Answer bias and sensitivity.}
Participants answered YES on $64.9\%$ of the photos, while the model answered YES on $50\%$. The hit rate was $75.3\%$
and the false-alarm rate $54.4\%$; a one-sample $t$-test of the YES rate against 50\% gives $p<0.001$. The mean
sensitivity is $d'=0.55$ (SD $0.64$; $t(29)=4.75$, $p<0.001$), which is above zero.

\paragraph{Confidence and perceived difficulty.}
On a 1--5 scale, mean overall confidence was $2.67$ (SD $0.88$) and mean ease of working out the rule was $2.50$
(SD $0.86$). Neither correlated significantly with a participant's agreement with the model (confidence: Spearman
$\rho=0.07$, $p=0.731$; ease: $\rho=0.24$, $p=0.193$; Figure~\ref{fig:us_confidence}).

\FloatBarrier  
{\small
\putbib[refs]
}
\end{bibunit}

\end{document}

%% file: appendix_refs.tex
\newcommand{\appxtag}{App.}          
\newcommand{\appxtags}{Apps.}        
\newcommand{\Appxtag}{Appendix}      
\newcommand{\Appxtags}{Appendices}   
\newcommand{\maintag}{Main}          
\newcommand{\Maintag}{Main}          
\newcommand{\appxsecstyle}{\arabic}  
\newif\ifappxrestart
\appxrestarttrue
\newif\ifappxSnumbers
\appxSnumbersfalse

\newif\ifinappendix
\protected\def\toappx{\ifinappendix\else\appxtag~\fi}
\protected\def\Toappx{\ifinappendix\else\Appxtag~\fi}
\protected\def\tomain{\ifinappendix\maintag~\fi}
\protected\def\Tomain{\ifinappendix\Maintag~\fi}
\protected\def\appxsec{\ifinappendix Sec.\else\appxtag\fi}
\protected\def\appxsecs{\ifinappendix Secs.\else\appxtags\fi}
\protected\def\Appxsec{\ifinappendix Section\else\Appxtag\fi}
\protected\def\Appxsecs{\ifinappendix Sections\else\Appxtags\fi}

\makeatletter
\newcommand{\appendixrefs}{%
  \global\inappendixtrue
  \renewcommand{\thesection}{\appxsecstyle{section}}
  \renewcommand{\thepage}{\texorpdfstring{\arabic{page}}{S\arabic{page}}}%
  \ifappxSnumbers\gdef\appx@numprefix{S}\else\global\let\appx@numprefix\@empty\fi
  \appx@renumber{figure}\appx@renumber{table}\appx@renumber{equation}\appx@renumber{algocf}%
  \appx@renumber{theorem}\appx@renumber{proposition}\appx@renumber{lemma}\appx@renumber{corollary}%
  \crefalias{section}{appendix}%
  \crefalias{subsection}{subappendix}%
  \crefalias{subsubsection}{subsubappendix}%
  \crefalias{figure}{appfigure}%
  \crefalias{table}{apptable}%
  \crefalias{equation}{appequation}%
  \crefalias{algocf}{appalgorithm}
  \crefalias{algorithm}{appalgorithm}
  \crefalias{theorem}{apptheorem}%
  \crefalias{proposition}{appproposition}%
  \crefalias{lemma}{applemma}%
  \crefalias{corollary}{appcorollary}}
\newcommand{\appx@renumber}[1]{%
  \@ifundefined{c@#1}{}{%
    \ifappxrestart
      \setcounter{#1}{0}%
      \@namedef{the#1}{\appx@numprefix\arabic{#1}}%
      \@namedef{theH#1}{S\arabic{#1}}%
    \fi}}

\newcommand{\appx@names}[6]{%
  \crefname{#1}{\tomain #3}{\tomain #4}\Crefname{#1}{\Tomain #5}{\Tomain #6}%
  \crefname{#2}{\toappx #3}{\toappx #4}\Crefname{#2}{\Toappx #5}{\Toappx #6}}
\AtEndPreamble{%
  \appx@names{section}{appendix}{Sec.}{Secs.}{Section}{Sections}%
  \appx@names{figure}{appfigure}{Fig.}{Figs.}{Figure}{Figures}%
  \appx@names{table}{apptable}{Tab.}{Tabs.}{Table}{Tables}%
  \appx@names{equation}{appequation}{Eq.}{Eqs.}{Equation}{Equations}%
  \appx@names{algorithm}{appalgorithm}{Alg.}{Algs.}{Algorithm}{Algorithms}
  \appx@names{theorem}{apptheorem}{Thm.}{Thms.}{Theorem}{Theorems}%
  \appx@names{proposition}{appproposition}{Prop.}{Props.}{Proposition}{Propositions}%
  \appx@names{lemma}{applemma}{Lem.}{Lems.}{Lemma}{Lemmas}%
  \appx@names{corollary}{appcorollary}{Cor.}{Cors.}{Corollary}{Corollaries}%
  \crefname{appendix}{\appxsec}{\appxsecs}\Crefname{appendix}{\Appxsec}{\Appxsecs}%
  \creflabelformat{appequation}{\textup{(#2#1#3)}}%
}
\makeatother